\documentclass{article}

\usepackage{graphicx}
\usepackage{subcaption}

 \usepackage[main, final]{neurips_2026}

\usepackage[utf8]{inputenc} 
\usepackage[T1]{fontenc}    
\usepackage{hyperref}       
\usepackage{url}            
\usepackage{booktabs}       
\usepackage{amsfonts}       
\usepackage{nicefrac}       
\usepackage{microtype}      
\usepackage{xcolor}         
\usepackage{amsmath}
\usepackage{array}
\usepackage{multirow}
\usepackage{makecell}
\usepackage{wrapfig}
\title{VETime: Vision Enhanced Zero-Shot Time Series Anomaly Detection}

\hypersetup{
  pdftitle={VETime: Vision Enhanced Zero-Shot Time Series Anomaly Detection},
  pdfauthor={Yingyuan Yang, Tian Lan, Yifei Gao, Yimeng Lu, Liming An, Wenjun He, Meng Wang, Chenghao Liu, Chen Zhang}
}

\author{%
  Yingyuan Yang$^1$ \quad
  Tian Lan$^1$ \quad
  Yifei Gao$^1$ \quad
  Yimeng Lu$^1$ \quad
  Xuming An$^1$ \\
  \textbf{Wenjun He$^2$} \quad
  \textbf{Meng Wang$^2$} \quad
  \textbf{Chenghao Liu$^{3,\dagger,*}$} \quad
\textbf{Chen Zhang$^{1,\dagger}$} \\
  \\
  $^1$Department of Industrial Engineering, Tsinghua University, Beijing, China \\
  $^2$2012 Lab, Huawei Technologies Ltd, Shenzhen, China \\
  $^3$Datadog AI Research \\
  yangyingyuan22@mails.ucas.ac.cn \quad lant23@mails.tsinghua.edu.cn \\
  gao-yf@mail.tsinghua.edu.cn \quad \{luym25,axm24\}@mails.tsinghua.edu.cn\\
   \{hewenjun8, wangmeng71\}@huawei.com\\
  twinsken@gmail.com \quad  zhangchen01@tsinghua.edu.cn
}

\begin{document}

\maketitle
\begingroup
\renewcommand{\thefootnote}{\fnsymbol{footnote}}
\footnotetext[2]{Corresponding authors.}
\footnotetext[1]{This work was completed prior to joining Datadog.}
\endgroup

\begin{abstract}
 Time-series anomaly detection (TSAD) requires identifying both immediate point anomalies and long-range context anomalies. However, existing zero-shot foundation models face a fundamental trade-off: 1D temporal models provide fine-grained pointwise localization but lack a global contextual perspective, while 2D vision-based models capture global patterns but suffer from weak temporal alignment and coarse-grained localization. To resolve this dilemma, we propose VETime, the first zero-shot TSAD framework that unifies temporal and visual modalities through fine-grained visual-temporal alignment and dynamic fusion. VETime introduces a Traceable Image Conversion and a Patch-level Temporal Alignment module to establish a shared visual-temporal timeline, preserving discriminative details while maintaining temporal sensitivity. Furthermore, we design an Anomaly Window Contrastive Learning mechanism and a Task-adaptive Multi-modal Fusion to adaptively integrate the complementary perceptual strengths of both modalities. Extensive experiments demonstrate that VETime significantly outperforms state-of-the-art models in zero-shot scenarios, achieving superior localization precision with lower computational overhead than current vision-based approaches.
\end{abstract}


\section{Introduction}

Time-series anomaly detection (TSAD) is a fundamental yet challenging problem, which requires identifying rare deviations while providing precise temporal localization.
In practice, domain heterogeneity and resource-constrained deployment render dataset-specific training impractical, driving the development of Time-Series Foundation Models (TSFMs) for zero-shot anomaly detection. In this setting, models \citep{shentu2024towards,lan2025foundationmodelszeroshottime} are evaluated directly on unseen target domains without any target-specific training or fine-tuning.

\begin{figure}[t]
  \centering
  \includegraphics[width=0.9\linewidth]{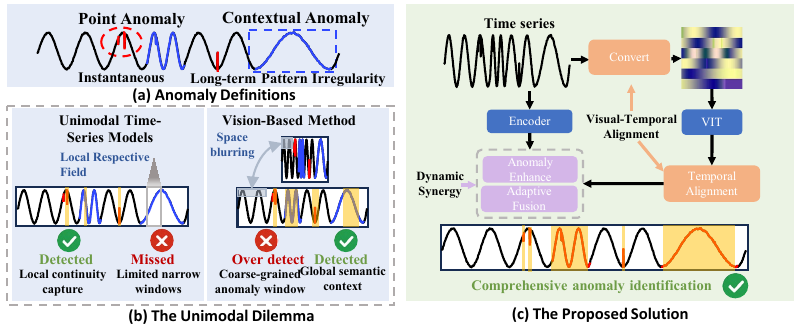}
  \caption{Comparison of the previous TSAD methods and the proposed VETime. While existing methods often rely on single-modality or coarse visual mapping, VETime introduces fine-grained temporal alignment and dynamic fusion to bridge the gap between visual and temporal perceptions.}
  \label{fig:img1}
\end{figure}

A general zero-shot anomaly detector must handle at least two complementary anomaly patterns, as shown in Figure~\ref{fig:img1}(a). \textbf{Point anomalies} appear as abrupt, short-lived numerical deviations \citep{shentu2024towards}, requiring fine-grained local sensitivity. \textbf{Context anomalies} correspond to longer-range violations or structural context \citep{he2025harnessing}, requiring a macroscopic view of the sequence. Models that overfit to either local continuity or global structure may generalize poorly across unseen datasets where anomaly duration and morphology vary drastically.

Existing zero-shot TSAD methods mainly follow two directions.
First, TSFMs \citep{ansari2024chronos,shi2024time,ekambaram2025tspulse} operate on 1D numerical sequences and preserve local temporal resolution, making them suitable for pointwise localization.
However, their windowed numerical modeling often provides a local receptive field and lacks macroscopic context for long-range anomalies.
Conversely, vision- or VLM-based methods \citep{zhou2024llmsunderstandtimeseries,xu2025can,he2025harnessing} render time series as images and rely on prompt-based reasoning, exploiting pretrained visual perception to capture global patterns.
Furthermore, while existing methods \citep{he2025harnessing,zhou2024llmsunderstandtimeseries} demonstrate the utility of visual information for TSAD, they suffer from coarse windows and a lack of temporal alignment, leading to poor pointwise anomaly localization \citep{zhuang2024see}.
These single-perspective foundation models face a critical competency gap in effectively handling both anomaly types simultaneously, as shown in Figure~\ref{fig:img1}(b).

A natural progression to resolve this dilemma is integrating temporal and visual modalities to leverage their complementary strengths. Although pioneers like Time-VLM \citep{zhong2025time} and Aurora \citep{wu2026Aurora} explore multi-modal time-series modeling, they are tailored for forecasting. This causes critical task misalignment: forecasting prioritizes macro sequence trends over dense timestamp-level localization, while prediction-residual paradigms struggle to generalize across diverse anomaly patterns under zero-shot settings. Consequently, developing a native zero-shot multimodal TSAD framework requires tackling two key obstacles: \textbf{Visual-Temporal Ordered Alignment:} restoring explicit temporal indexing onto a shared timeline required for robust anomaly discrimination during 1D-to-2D visual conversion and multi-modal alignment; and \textbf{Dynamic Synergy for Diverse Anomaly Patterns:} adaptively fusing 1D local correlations and 2D macroscopic patterns to provide comprehensive context information for tackling complex unseen anomaly scenarios.

To bridge these gaps, we propose VETime, the first framework that unifies temporal and visual features via effective alignment and fusion, enabling robust anomaly detection and precise localization, as shown in Figure~\ref{fig:img1}(c).
To overcome the \textbf{alignment barrier}, we first develop a novel \textit{Traceable Image Conversion} method which embeds the 1D sequence into a 2D topological manifold, preserving information-dense anomaly details without losing explicit temporal indices.
Complementing this, we propose the \textit{Patch-level Temporal Alignment} module, which enhances 2D visual representations from the pre-trained ViT with a 1D temporal ordering, establishing the structural basis for fine-grained cross-modal interaction.
Subsequently, to facilitate \textbf{dynamic synergy}, we introduce two key mechanisms. Considering the distinct perceptual characteristics of visual and temporal modalities regarding anomalies, we propose a specialized \textit{Anomaly Window Contrastive Learning}. This incorporates intra- and inter-window contrast between multi-modal features to achieve comprehensive anomaly identification.
Finally, a \textit{Task-adaptive Multi-modal Fusion} module equipped with a reconstruction head is applied to facilitate robust multi-modal fusion. This mechanism performs dynamic, adaptive weighted fusion of the aligned features, ensuring precise performance in downstream anomaly detection. Our main contributions are listed as follows:
\begin{itemize}
    \item We propose VETime, the first zero-shot TSAD framework that integrates visual and temporal features, effectively exploiting the distinct perceptual advantages of each modality.
    \item We introduce the Traceable Image Conversion and Patch-level Temporal Alignment, which jointly capture informative visual contexts while preserving critical temporal sensitivities.
    \item We design Anomaly Window Contrastive Learning and Task-adaptive Multi-modal Fusion, harnessing multi-modal perceptual strengths across diverse anomaly types.
    \item Extensive experiments demonstrate that VETime significantly outperforms state-of-the-art zero-shot TSFMs. Compared to VLM-based baselines, VETime achieves superior precision with much lower computational overhead.
\end{itemize}

\section{Related work}
\label{gen_related}
\paragraph{TSAD and time-series foundation models.}
Traditional TSAD relies on domain-specific reconstruction or prediction models \citep{hundman2018detecting,su2019robust,audibert2020usad,tuli2022tranad,xu2022anomaly,yang2023dcdetector}. Recently, time-series foundation models (TSFMs) have advanced cross-domain zero-shot generalization via large-scale pretraining \citep{goswami2024moment,das2024decoder,ansari2024chronos,shi2024time,ekambaram2025tspulse,gao2024units}. Anomaly-oriented models such as DADA and TimeRCD \citep{shentu2024towards,lan2025foundationmodelszeroshottime} further introduce anomaly-specific objectives or synthetic anomaly supervision for zero-shot TSAD. Despite preserving fine-grained local temporal precision, these 1D numerical approaches inherently lack the macroscopic perspective required for identifying long-range contextual anomalies. VETime addresses this gap by complementing zero-shot temporal modeling with traceable 2D visual context and visual-temporal fusion.
\begin{figure*}[t]
  \centering
  \includegraphics[width=1\linewidth]{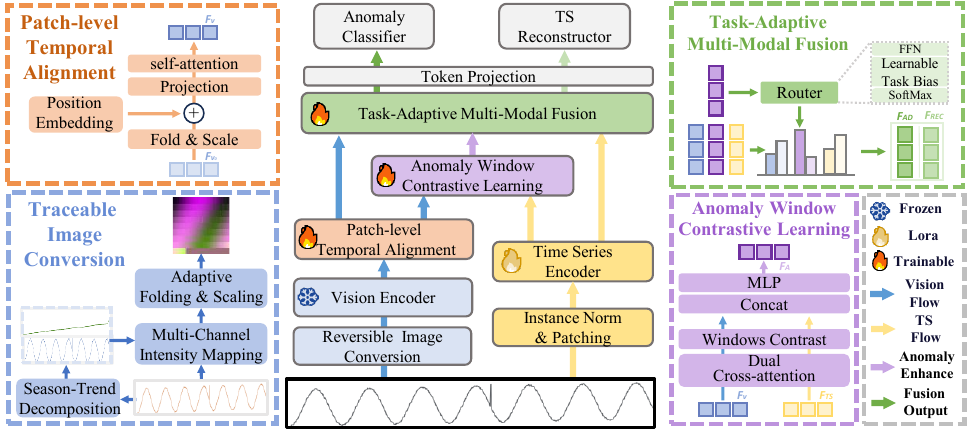}
  \caption{Overview of the proposed framework. The time series is first processed by a time-series encoder to extract temporal features $F_{TS}$ while simultaneously undergoing Traceable Image Conversion to generate visual input. Then visual features $F_{V_0}$ extracted from a frozen pre-trained image encoder are subsequently transformed into $F_V$ through Patch-level Temporal Alignment to reinforce their temporal positional association. Then, $F_{TS}$ and $F_V$ are input into the Anomaly Window Contrastive Learning to derive an anomaly-enhanced representation $F_A$. Finally, a Task-adaptive Multi-modal Fusion module integrates all features ($F_A$, $F_{TS}$ , $F_V$). Final outputs $F_{AD}$ and $F_{Rec}$ are mapped to the original sequence length via token projection for the respective anomaly classification and reconstruction heads.}
  \label{fig:img2}
\end{figure*}

\paragraph{Time-series-to-image representations and vision-enhanced time-series modeling.}
Transforming 1D time series into 2D visual representations allows models to leverage mature Vision Transformers (ViTs) and capture multi-scale periodic patterns~ \citep{he2025harnessing}. Traditional methods often rely on fixed spatial-temporal transforms like Gramian Angular Fields (GAF) or Markov Transition Fields (MTF)~\citep{wang2015imaging}. However, these transformations are non-invertible, destroying temporal phase details essential for point-wise localization. Recent vision-enhanced methods \citep{chen2024visionts, wang2024timemixer, dong2025teaching} dilute the mapping between image patches and original timestamps, inevitably introducing severe information bottlenecks. In contrast, our work introduces an adaptive folding mechanism with exact spatial-temporal mapping, maintaining full reversibility and fine-grained temporal fidelity.

\paragraph{Vision- and VLM-based zero-shot TSAD.}
Recent studies \citep{zhuang2024see,he2025harnessing,xu2025can} also leverage pretrained vision or multimodal large language models (MLLMs) for anomaly detection.
VLM4TS \citep{he2025harnessing} localizes candidate anomaly intervals via the ViT4TS module and applies a VLM to verify and refine them using global visual context.
While validating the efficacy of visual perception for global patterns, their reliance on rendered line plots or prompt-based reasoning incurs high inference costs and weak timestamp-to-patch mapping. In contrast, VETime is an efficient detector that realigns visual patches to 1D temporal tokens, enabling dense pointwise anomaly scoring without computationally expensive VLM prompting.

\section{Methodology}

To harness the differentiated yet complementary advantages of temporal and visual modalities, we propose a unified framework capable of fine-grained temporal alignment and dynamic interaction for anomaly detection. As illustrated in Figure~\ref{fig:img2}, the framework comprises four core components:

\begin{itemize}
    \item \textbf{Traceable Image Conversion (TIC):} Transforms 1D time series into traceable visual representations, encapsulating global and local correlations into semantically rich visual formats.
\item \textbf{Patch-level Temporal Alignment (PTA):} Realigns visual features to the temporal axis to ensure fine-grained semantic correspondence.
\item \textbf{Anomaly Window Contrastive Learning (AWCL):} Leverages the complementary nature of visual and temporal modalities to enforce multi-scale discriminability.
\item \textbf{Task-adaptive Multi-modal Fusion (TMF):} Dynamically integrates multimodal features using sequence reconstruction as an auxiliary constraint.
\end{itemize}

Ultimately, the fused representations capitalize on the synergy between temporal and visual domains, facilitating accurate localization across diverse anomaly types. It is noted that our framework supports univariate and multivariate time series settings. The main text focuses on univariate cases, with the multivariate extension detailed in Appendix \ref{appendix:Multivariate}.

\subsection{Traceable image conversion}
\begin{figure*}
    \centering
    \includegraphics[width=1\linewidth]{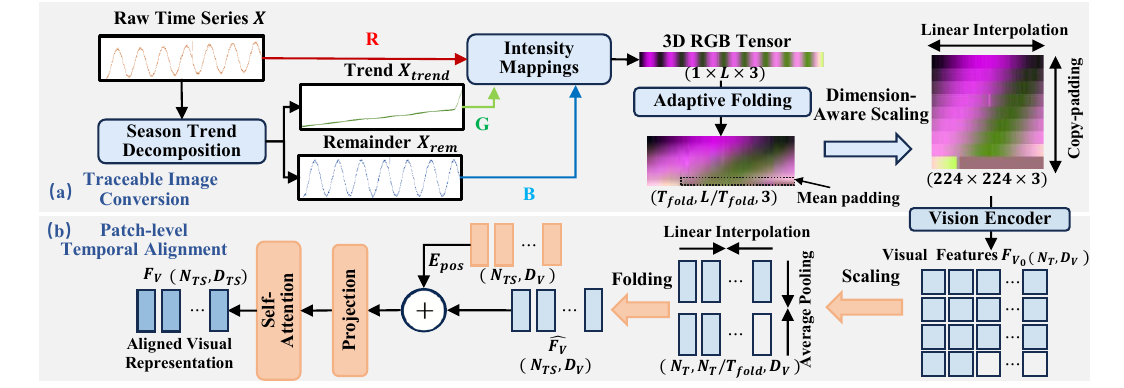}
    \caption{The framework of Traceable Image Conversion and Patch-level Temporal Alignment.}
    \label{fig:img3}
\end{figure*}

To accentuate anomalous signals, we transform univariate time series into high-density visual representations via a three-stage pipeline as shown in Figure~\ref{fig:img3}(a).
\paragraph{Multi-channel intensity mapping:}
Unlike prior single-channel mapping approaches \citep{chen2024visionts}, we employ a multi-channel encoding strategy, which capitalizes on the standard RGB architecture to incorporate denser information and better expose latent anomalous patterns. Following DLinear \citep{Zeng2022AreTE}, the raw series $X \in \mathbb{R}^{L}$ is decomposed into a trend component $X_{trend}$ and a remainder component $X_{rem}$, where $L$ is the length of the series. The triplet $\{X, X_{trend}, X_{rem}\}$ is independently Min-Max normalized to $[0, 255]$ and mapped to the R, G, and B channels, yielding a $1 \times L \times 3$ tensor that explicitly encompasses both global trends and high-frequency residuals.

\paragraph{Adaptive folding:}
To accommodate diverse temporal scales, we adopt a periodicity-based folding strategy \citep{wu2023timesnet} to transform a 1D sequence into a 2D grid with dimensions $(T_{fold}, \lceil L/T_{fold} \rceil, 3)$. The folding period $T_{fold}$ is dynamically estimated via the autocorrelation function (defaulting to $\sqrt{L}$ for non-periodic samples), and adjusted to be a multiple of the ViT patch size to prevent temporal information discontinuity. For indivisible lengths, we use the mean of the final period to pad trailing segments, minimizing distribution shifts compared to zero-padding.

\paragraph{Dimension-aware scaling:}
To standardize input resolution from $(T_{fold}, \lceil L/T_{fold} \rceil, 3)$ to $(224, 224, 3)$ without fidelity loss, we apply a decoupled scaling strategy. We use linear interpolation along the time axis (horizontal) to preserve waveform continuity, while employing a copy-padding strategy along the period axis (vertical) to prevent semantic distortion or pseudo-patterns that could arise from interpolation across distinct periods.

\subsection{Patch-level temporal alignment}
The generated visual representations are fed into a frozen visual encoder to extract initial visual features $F_{V_0} \in \mathbb{R}^{N_V \times D_V}$. In parallel, the time series undergoes instance normalization and patching, followed by a temporal encoder to yield $F_{TS} \in \mathbb{R}^{N_{TS} \times D_{TS}}$. Here, $N$ and $D$ denote the patch count and feature dimension for the respective modalities. To bridge the structural discrepancy between modalities, we introduce a Patch-level Temporal Alignment module for $F_{V_0}$ (Figure~\ref{fig:img3}(b)).

Specifically, visual features are mapped back to the 1D temporal domain by inverting the folding logic. $F_{V_0}$ is reshaped to the initial 2D grid, linearly interpolated along the temporal axis to match the temporal patch count, and average-pooled along the periodicity axis to aggregate redundant repetitions. The resulting features are flattened to yield the concise aligned image features $\hat{F}_{V} \in \mathbb{R}^{N_{TS} \times D_V}$.

Finally, to recover temporal context lost during visual encoding, we add learnable positional encoding $E_{POS} \in \mathbb{R}^{N_{TS} \times D_V} $ followed by projection and self-attention layers to model intra- and inter-patch dependencies. Ultimately, visual representations $F_{V}$ maintain temporal correspondence with $F_{TS}$.
\subsection{Anomaly window contrastive learning}

\begin{wrapfigure}[12]{r}{0.43\textwidth}
    \centering
    \vspace{-\baselineskip}
    \includegraphics[width=0.38\textwidth]{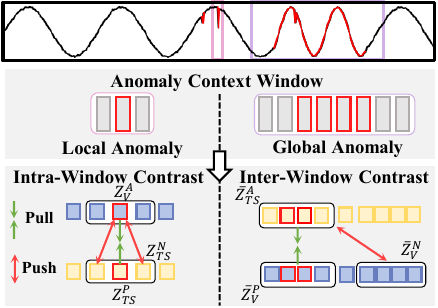}
    \caption{An illustration of Anomaly window contrastive learning}
    \label{fig:img4}
\end{wrapfigure}

Time-series ($F_{TS}$) and visual $F_{V}$ features are first input into a dual cross-attention. Here, features from each modality serve as queries to integrate complementary contexts, followed by a residual FFN for refinement, yielding updated representations $Z_{TS}$ and $Z_{V}$ (detailed in Appendix~\ref{appx.cross-attention}).
Given the inherent disparities between time-series and image features, specifically concerning their receptive fields and sensitivity to anomalies, we employ a hybrid strategy comprising Intra-Window and Inter-Window contrastive learning to model discriminative features across scales, as shown in Figure~\ref{fig:img4}.

\paragraph{Anomaly context windows:}
During source-domain pre-training, we dynamically construct adaptive windows around anomalies. Point-level labels are converted to patch-level labels (set to 1 if any timestamp within the patch is anomalous). For a continuous anomaly segment spanning $L_a$ patches, we define an Anomaly Context Window by symmetrically extending the anomaly segment bidirectionally into the normal context, with a maximum length of $L_w = 2 L_a$. Each window contains one continuous anomaly instance surrounded by its local context.

\paragraph{Intra-window contrastive learning:}
Targeting short-duration point anomalies ($L_a \le 1$ patches, e.g., outliers), this component enforces fine-grained anomaly alignment for the visual feature $Z_{V}$. Within a given window, the anomalous visual feature serves as the anchor $Z_{V}^{A}$, paired with the corresponding temporal feature as the positive sample $Z_{TS}^{A}$. Normal temporal features within the same window act as negatives $Z_{TS}^{N}$. The intra-window contrastive loss $\mathcal{L}_{intra}$ is defined as:
\begin{equation}
\mathcal{L}_{intra} = - \log \frac{\exp(Z_{V}^{A} \cdot Z_{TS}^{A}/\tau)}{\exp(Z_{V}^{A} \cdot Z_{TS}^{A}/\tau) + \sum_{k \in \mathcal{N}_{neg}} \exp(Z_{V}^{A}\cdot Z_{TS}^{k}/\tau)},
\end{equation}
where $\cdot$ denotes dot product similarity, $\tau$ is the temperature parameter, and $\mathcal{N}_{neg}$ represents the set of normal features within the window.

\paragraph{Inter-window contrastive learning:}
For long-duration context anomalies ($L_a \ge 2$  patches, e.g., seasonal anomaly), we enhance global discriminability for temporal feature $Z_{TS}$. Features are average-pooled along the temporal dimension within each window. The pooled time-series feature of an anomaly window $\bar{Z}_{TS}^{A}$ serves as the anchor, paired with its visual counterpart $\bar{Z}_{V}^{A}$, while pooled features from normal windows $\bar{Z}_{V}^{N}$ serve as negatives. The inter-window contrastive loss $\mathcal{L}_{inter}$ is:

\begin{equation}
\mathcal{L}_{inter} = - \log \frac{\exp(\bar{Z}_{TS}^{A}\cdot \bar{Z}_{V}^{A}/\tau)}{\exp(\bar{Z}_{TS}^{A}\cdot \bar{Z}_{V}^{A}/\tau) + \sum_{j \in \mathcal{W}_{neg}} \exp(\bar{Z}_{TS}^{A}\cdot \bar{Z}_{V}^{j}/\tau)},
\end{equation}
where $\mathcal{W}_{neg}$ denotes the set of normal window features. The total contrastive loss $\mathcal{L}_{aw}$ is computed independently for each sample to prevent interference from the diverse normal and anomalous patterns across instances:
\begin{equation}
\mathcal{L}_{aw} = \frac{1}{N} \sum_{i=1}^{N} ( \frac{1}{W_s}\sum_{s=1}^{W_s} \mathcal{L}_{intra}^{(i,s)} +\frac{1}{W_l}\sum_{l=1}^{W_l} \mathcal{L}_{inter}^{(i,l)}),
\end{equation}
where $N$ is the number of samples, $W_s$ and $W_l$ are the numbers of short and long anomalies in each instance. The optimized representations are concatenated and refined via an FFN to produce the anomaly-enhanced representation $F_A= {\text{FFN}}([Z_{TS};Z_{V}])$, where $[\cdot;\cdot]$ denotes feature concatenation.

\subsection{Task-adaptive multi-modal fusion}
To realize dynamic semantic integration and efficient end-to-end training, we propose a Task-adaptive Multi-modal Fusion Module. This module acts as a dynamic routing mechanism, treating time-series ($F_{TS}$), vision ($F_{V}$), and anomaly-enhanced ($F_{A}$) features as experts in a candidate space.

\paragraph{Dynamic weight routing:}
To assign importance to the three experts $\mathcal{M} = \{TS, V, A\}$ for anomaly detection ($AD$) and reconstruction ($Rec$), the routing network generates patch-level weights $W^t \in \mathbb{R}^{N_{TS} \times \mathcal{M}}$ from the anomaly-enhanced feature $F_A \in \mathbb{R}^{N_{TS} \times D}$:
\begin{equation}
W^t = \text{Softmax}(\text{MLP}(F_A) + \mathbf{b}^t),
\end{equation}
where $\mathbf{b}^t \in \mathbb{R}^{\mathcal{M}}$ is a learnable task-specific bias, and the Softmax function normalizes the probabilities across the expert dimension.

To prevent premature routing collapse and encourage diverse expert utilization, we introduce an auxiliary load-balancing loss. We compute the temporally averaged expert utilization $\hat{w}_{m}^t = \frac{1}{N_{TS}} \sum_{i=1}^{N_{TS}} w_{i,m}^t$, where $w_{i,m}^t$ is the assigned probability of expert $m$ at patch $i$ for task $t$, and minimize its negative entropy to maximize global routing diversity:
$
\mathcal{L}_{e} = \sum_{t \in \{AD, Rec\}} \sum_{m \in \mathcal{M}} \hat{w}_{m}^t \log \hat{w}_{m}^t.
$
This balances global modality allocation while permitting local dynamic gating based on contextual demands.

\paragraph{Task-adaptive fusion:}
Based on the routing weights, the fused feature $F_{t}$ for each task is computed as a weighted sum of the experts $F_m \in \mathbb{R}^{N_{TS} \times D}$:
\begin{equation}
F_{t} = \sum_{m \in \mathcal{M}} W_{m}^t \odot F_{m}, \quad t \in \{AD, Rec\},
\end{equation}
where $W_{m}^t \in \mathbb{R}^{N_{TS} \times 1}$ is the weight vector for expert $m$, and $\odot$ denotes element-wise multiplication broadcasted along the feature dimension.

The resulting $F_{AD}$ and $F_{Rec}$ are subsequently fed into their respective heads. Crucially, sequence reconstruction serves as an auxiliary task; by forcing the network to reconstruct the raw input, it fosters deep cross-modal interactions and mitigates overfitting to sparse anomaly labels.

\begin{table*}[b!]
    \centering
    \renewcommand{\arraystretch}{0.75}
    \caption{
        Performance of VETime against zero-shot baselines on 26 public datasets (11 univariate and 15 multivariate). Asterisked (*) results are excluded from ranking due to data leakage of models marked with (\(\dagger\)). \textcolor{red}{Red}: best, \textcolor{blue}{Blue}: second best.
    }
    \label{tab:tab1}
    \resizebox{\textwidth}{!}{%
    \begin{tabular}
    {
    >{\centering\arraybackslash}p{1.8cm}
    >{\centering\arraybackslash}p{2.0cm}
    *{11}{>{\centering\arraybackslash}p{1.1cm}}
    *{3}{>{\centering\arraybackslash}p{0.8cm}}
    }

        \hline
        \multirow{2}{*}{\textbf{Metric}} & \multirow{2}{*}{\textbf{Model}} & \multicolumn{11}{c}{\textbf{Univariate Datasets}} & \textbf{Total} & \textbf{Total} & \textbf{Avg} \\
        \cline{3-13}
        & & \textbf{IOPS} & \textbf{MGAB} & \textbf{NAB} & \textbf{NEK} & \textbf{Power} & \textbf{SED} & \textbf{Stock} & \textbf{TODS} & \textbf{UCR} & \textbf{WSD} & \textbf{YAHOO} & \textbf{1st} & \textbf{2nd} & \textbf{Rank} \\
        \hline
        \multirow{8}{*}{Affiliation-F1} & VETime      & \textcolor{red}{\textbf{90.53}} & \textcolor{blue}{\textbf{68.03}} & \textcolor{blue}{\textbf{88.57}} & 79.56 & \textcolor{blue}{\textbf{78.78}} & \textcolor{red}{\textbf{97.31}} & 69.60 & 85.85 & \textcolor{red}{\textbf{85.06}} & \textcolor{blue}{\textbf{94.31}} & \textcolor{red}{\textbf{97.15}} & \textcolor{red}{\textbf{04}}& \textcolor{red}{\textbf{04}} &\textcolor{red}{\textbf{2.55}}\\
        & TimeRCD     & 83.28 & \textcolor{red}{\textbf{70.69}} & 82.48 & 79.73 & \textcolor{red}{\textbf{85.51}} & \textcolor{blue}{\textbf{96.87}} & 71.84 & 86.37 & \textcolor{blue}{\textbf{84.63}} & 90.33 & \textcolor{blue}{\textbf{96.65}} & \textcolor{blue}{\textbf{02}} & \textcolor{blue}{\textbf{03}}&\textcolor{blue}{\textbf{3.27}}\\
        & DADA $\dagger$   & 89.37\rlap{*} & 67.66\rlap{*} & 86.56 & \textcolor{red}{\textbf{95.40}} & 69.79 & 65.18 & \textcolor{red}{\textbf{98.77}} & 76.89 & 72.21 & 93.92 & 92.20\rlap{*} & \textcolor{blue}{\textbf{02}} & 00&4.00 \\
        & TS-Pulse    & 68.76 & 67.33 & 70.80 & 73.05 & 69.94 & 67.44 & 67.93 & 67.90 & 67.70 & 68.22 & 70.05 & 00 & 00 &6.64\\
        & MOMENT$\dagger$     & 87.54\rlap{*} & 66.76\rlap{*} & \textcolor{red}{\textbf{90.45}}\rlap{*} & 92.26 & 75.97 & 59.13 & 45.26 & 59.76 & 75.77 & \textcolor{red}{\textbf{95.39}} & 79.99\rlap{*} & \textcolor{blue}{\textbf{02}} & 00 &4.73\\
        & TimesFM     & 81.88 & 66.95 & 79.73 & 90.49 & 69.88 & 67.14 & \textcolor{blue}{\textbf{97.53}} & \textcolor{blue}{\textbf{89.08}} & 70.03 & 78.97 & 91.28 & 00 & 02&5.27 \\
        & Chronos     & \textcolor{blue}{\textbf{90.12}} & 67.89 & 86.66 & \textcolor{blue}{\textbf{93.63}} & 69.72 & 67.89 & 96.85 & \textcolor{red}{\textbf{91.96}} & 74.35 & 90.98 & 96.34 & 01 & 02 &\textcolor{blue}{\textbf{3.27}}\\
        & Time MOE    & 76.34 & 67.23 & 80.51 & 80.50 & 71.19 & 60.98 & 63.28 & 54.68 & 73.56 & 80.25 & 69.70 & 00 & 00&6.27 \\
        \hline
        \multirow{8}{*}{F1\_T} & VETime & \textcolor{blue}{\textbf{46.15}} & \textcolor{red}{\textbf{2.00}} & \textcolor{red}{\textbf{44.31}} & \textcolor{blue}{\textbf{60.86}} & \textcolor{blue}{\textbf{20.16}} & \textcolor{red}{\textbf{74.35}} & 22.41 & \textcolor{red}{\textbf{68.33}} & \textcolor{blue}{\textbf{36.25}} & \textcolor{red}{\textbf{50.42}} & \textcolor{red}{\textbf{91.54}} & \textcolor{red}{\textbf{06}} & \textcolor{blue}{\textbf{04}} &\textcolor{red}{\textbf{1.91}}\\
        & TimeRCD & 28.44 & \textcolor{blue}{\textbf{1.81}} & \textcolor{blue}{\textbf{38.85}} & 35.87 & \textcolor{red}{\textbf{28.47}} & \textcolor{blue}{\textbf{69.43}} & 31.73 & \textcolor{blue}{\textbf{65.89}} & 34.30 & 35.04 & \textcolor{blue}{\textbf{85.86}} & \textcolor{blue}{\textbf{01}} & \textcolor{red}{\textbf{05}} &\textcolor{blue}{\textbf{3.27}}\\
        & DADA $\dagger$ & 42.50\rlap{*} & 00.91\rlap{*} & 37.24 & 47.98 & 19.80 & 09.56 & \textcolor{red}{\textbf{95.49}} & 35.18 & 07.22 & \textcolor{blue}{\textbf{48.46}} & 79.52\rlap{*} & \textcolor{blue}{\textbf{01}} & 01 &4.45\\
        & TS-Pulse & 04.10 & 00.81 & 34.61 & 27.07 & 19.90 & 09.71 & 15.98 & 13.45 & 05.12 & 4.57 & 05.50 & 00 & 00 &7.00\\
        & MOMENT $\dagger$ & 33.15\rlap{*} & 00.80\rlap{*} & 52.27\rlap{*} & \textcolor{red}{\textbf{63.66}} & 19.91 & 09.54 & 18.04 & 17.47 & 13.02 & 41.98 & 11.69\rlap{*} & \textcolor{blue}{\textbf{01}} & 00&5.00 \\
        & TimesFM & \textcolor{red}{\textbf{48.95}} & 00.93 & 36.74 & 36.63 & 19.80 & 09.58 & 88.94 & 51.13 & \textcolor{red}{\textbf{41.38}} & 10.78 & 83.46 & \textcolor{blue}{\textbf{02}} & 00&4.18 \\
        & Chronos & 45.45 & 01.10 & 36.10 & 33.16 & 19.90 & 13.18 & \textcolor{blue}{\textbf{89.30}} & 53.90 & 10.88 & 39.82 & 79.00 & 00 & \textcolor{blue}{\textbf{01}} &4.27\\
        & Time MOE & 25.95 & 00.63 & 38.70 & 15.78 & 19.85 & 17.73 & 34.13 & 20.91 & 08.29 & 22.6 & 37.11 & 00 & 00&5.91 \\
        \hline
        \multirow{8}{*}{Standard-F1} & VETime & \textcolor{red}{\textbf{34.56}} & \textcolor{red}{\textbf{1.89}} & \textcolor{red}{\textbf{37.87}} & \textcolor{blue}{\textbf{53.95}} & \textcolor{blue}{\textbf{20.12}} & \textcolor{red}{\textbf{75.02}} & 23.83 & \textcolor{red}{\textbf{69.13}} & \textcolor{red}{\textbf{30.61}} & \textcolor{red}{\textbf{46.55}} & \textcolor{red}{\textbf{88.93}} & \textcolor{red}{\textbf{08}} & \textcolor{blue}{\textbf{02}}&\textcolor{red}{\textbf{1.73}} \\
        & TimeRCD & 24.22 & \textcolor{blue}{\textbf{01.62}} & \textcolor{blue}{\textbf{27.70}} & 33.05 & \textcolor{red}{\textbf{28.59}} & \textcolor{blue}{\textbf{69.88}} & 32.61 & \textcolor{blue}{\textbf{67.02}} & \textcolor{blue}{\textbf{28.13}} & 31.96 & \textcolor{blue}{\textbf{87.02}} & \textcolor{blue}{\textbf{01}} & \textcolor{red}{\textbf{06}} &\textcolor{blue}{\textbf{3.27}}\\
        & DADA $\dagger$ & 32.76\rlap{*} & 00.80\rlap{*} & 26.91 & 48.24 & 15.99 & 02.69 & \textcolor{red}{\textbf{95.59}} & 28.18 & 03.36 & \textcolor{blue}{\textbf{45.06}} & 79.30\rlap{*} & \textcolor{blue}{\textbf{01}} & 01&4.36 \\
        & TS-Pulse & 03.54 & 00.73 & 21.61 & 23.96 & 18.27 & 08.84 & 15.46 & 12.45 & 02.05 & 2.17 & 04.00 & 00 & 00 &6.82\\
        & MOMENT $\dagger$ & 30.69\rlap{*} & 00.67\rlap{*} & 44.75\rlap{*} & \textcolor{red}{\textbf{63.85}} & 16.39 & 03.36 & 19.38 & 14.64 & 09.00 & 41.42 & 10.54\rlap{*} & \textcolor{blue}{\textbf{01}} & 00&4.73 \\
        & TimesFM & \textcolor{blue}{\textbf{34.28}} & 00.83 & 26.46 & 38.15 & 16.73 & 02.96 & 89.13 & 40.08 & 07.86 & 38.5 & 84.44 & 00 & \textcolor{blue}{\textbf{01}} &4.27\\
        & Chronos & 32.69 & 00.99 & 26.22 & 33.54 & 17.47 & 08.74 & \textcolor{blue}{\textbf{89.41}} & 40.52 & 08.21 & 34.58 & 78.89 & 00 & \textcolor{blue}{\textbf{01}} &4.55\\
        & Time MOE & 26.52 & 00.45 & 26.20 & 11.47 & 12.16 & 17.73 & 34.32 & 16.38 & 04.09 & 20.09 & 27.50 & 00 & 00&6.55\\
        \hline
        \multirow{8}{*}{VUS-PR} & VETime & \textcolor{red}{\textbf{30.79}} & \textcolor{blue}{\textbf{0.70}}& \textcolor{red}{\textbf{32.98}} & \textcolor{blue}{\textbf{57.54}} & \textcolor{blue}{\textbf{11.78}} & \textcolor{red}{\textbf{89.05}} & 76.08 & \textcolor{red}{\textbf{94.20}} & \textcolor{red}{\textbf{24.99}} & \textcolor{blue}{\textbf{33.85}} & \textcolor{red}{\textbf{88.61}} & \textcolor{red}{\textbf{06}} & \textcolor{red}{\textbf{04}}&\textcolor{red}{\textbf{2.00}} \\
        & TimeRCD & \textcolor{blue}{\textbf{20.23}} & \textcolor{red}{\textbf{01.05}} & 24.32 & 27.88 & \textcolor{red}{\textbf{21.25}} & \textcolor{blue}{\textbf{80.75}} & 77.28 & \textcolor{blue}{\textbf{93.46}} & \textcolor{blue}{\textbf{23.09}} & 21.77 & 84.41 & \textcolor{blue}{\textbf{02}} & \textcolor{red}{\textbf{04}}&\textcolor{blue}{\textbf{3.09}} \\
        & DADA $\dagger$ & 24.97\rlap{*} & 00.57\rlap{*} & \textcolor{blue}{\textbf{24.73}} & 46.85 & 10.61 & 06.42 & \textcolor{red}{\textbf{99.51}} & 64.83 & 02.94 & 33.42 & 70.74\rlap{*} & 01 & 01 &4.00\\
        & TS-Pulse & 04.64 & 00.56 & 16.40 & 19.39 & 11.72 & 09.11 & 70.95 & 45.86 & 01.20 & 1.83 & 09.93 & 00 & 00&6.82 \\
        & MOMENT $\dagger$ & 37.35\rlap{*} & 00.56\rlap{*} & 45.38\rlap{*} & \textcolor{red}{\textbf{67.74}} & 10.50 & 04.31 & 76.97 & 56.45 & 06.17 & \textcolor{red}{\textbf{55.26}} & 30.81\rlap{*} & \textcolor{blue}{\textbf{02}} & 00&3.73 \\
        & TimesFM & 19.56 & 00.58 & 24.01 & 35.02 & 10.44 & 06.13 & \textcolor{blue}{\textbf{98.39}} & 72.89 & 06.03 & 21.57 & \textcolor{blue}{\textbf{86.78}} & 00 & \textcolor{blue}{\textbf{02}} &4.45\\
        & Chronos & 19.00 & 00.60 & 23.76 & 31.80 & 10.95 & 08.65 & 97.49 & 70.66 & 06.56 & 18.81 & 83.54 & 00 & 00&4.91 \\
        & Time MOE & 16.63 & 00.52 & 22.62 & 19.76 & 09.34 & 10.87 & 74.78 & 48.78 & 02.10 & 10.93 & 20.90 & 00 & 00&6.82 \\
        \hline
        \multicolumn{13}{l}{\textbf{VETime Grand Total (Univariate)}} & \textbf{24} & \textbf{14} & \textbf{2.05}\\
        \hline
    \end{tabular}
    }
    \renewcommand{\arraystretch}{0.85}
    \resizebox{\textwidth}{!}{%
    \begin{tabular}{cccccccccccccccccccc}
        \hline
        \multirow{2}{*}{\textbf{Metric}} & \multirow{2}{*}{\textbf{Model}} & \multicolumn{15}{c}{\textbf{Multivariate Datasets}} & \textbf{Total} & \textbf{Total} & \textbf{Avg} \\
        \cline{3-17}
        & & \textbf{MSL} & \textbf{PSM} & \textbf{SMAP} & \textbf{SMD} & \textbf{SWaT} & \textbf{GHL} & \textbf{Daph.} & \textbf{Exath.} & \textbf{Gen.} & \textbf{GECCO} & \textbf{MIT} & \textbf{SVDB} & \textbf{LTDB} & \textbf{CATS} & \textbf{TAO} & \textbf{1st} & \textbf{2nd} & \textbf{Rank} \\
        \hline
        \multirow{5}{*}{Affiliation-F1}
        & VETime & \textcolor{red}{\textbf{87.02}} & 77.27 & \textcolor{blue}{\textbf{87.06}} & \textcolor{blue}{\textbf{88.94}} & \textcolor{blue}{\textbf{75.45}} & \textcolor{blue}{\textbf{74.93}} & \textcolor{blue}{\textbf{84.61}} & \textcolor{red}{\textbf{99.90}} & \textcolor{blue}{\textbf{75.41}} & \textcolor{red}{\textbf{87.63}} & \textcolor{blue}{\textbf{75.16}} & \textcolor{red}{\textbf{73.03}} & \textcolor{red}{\textbf{78.28}} & 72.08 & \textcolor{blue}{\textbf{75.41}} & 05 & 08 & \textcolor{red}{\textbf{1.80}} \\
        & TimeRCD & 81.16 & \textcolor{red}{\textbf{81.61}} & \textcolor{red}{\textbf{87.73}} & \textcolor{red}{\textbf{92.58}} & 71.55 & \textcolor{red}{\textbf{74.96}} & 81.90 & 68.02 & 73.19 & 70.18 & \textcolor{red}{\textbf{75.18}} & 70.05 & 75.28 & 68.02 & 73.74 & 05 & 00 & 2.60 \\
        & DADA $\dagger$ & 76.57 & \textcolor{blue}{\textbf{81.27}} & 76.92 & 83.74 & \textcolor{red}{\textbf{76.18}} & 71.84 & \textcolor{red}{\textbf{92.40}} & 93.72 & \textcolor{red}{\textbf{87.84}} & \textcolor{blue}{\textbf{79.40}} & 71.73 & \textcolor{blue}{\textbf{72.20}} & \textcolor{blue}{\textbf{76.99}} & 68.59 & \textcolor{red}{\textbf{99.66}} & 04 & 04 & \textcolor{blue}{\textbf{2.40}} \\
        & MOMENT $\dagger$ & 74.55 & 65.79 & 77.42 & 74.00 & 70.17 & 40.90 & 67.74 & \textcolor{blue}{\textbf{93.89}} & 68.63 & 18.28 & 70.95 & 71.96 & 64.86 & \textcolor{red}{\textbf{74.32}} & 20.31 & 01 & 01 & 3.73 \\
        & LOF & \textcolor{blue}{\textbf{84.35}} & 61.98 & 63.32 & 64.13 & 56.34 & 61.00 & 64.00 & 61.00 & 51.00 & 56.00 & 66.00 & 53.00 & 40.00 & \textcolor{blue}{\textbf{73.00}} & 10.00 & 00 & 02 & 4.47 \\
        \hline
        \multirow{5}{*}{F1\_T}
        & VETime & \textcolor{red}{\textbf{43.52}} & \textcolor{red}{\textbf{39.06}} & \textcolor{red}{\textbf{45.52}} & \textcolor{blue}{\textbf{53.36}} & \textcolor{red}{\textbf{39.94}} & \textcolor{red}{\textbf{24.12}} & \textcolor{red}{\textbf{55.42}} & \textcolor{red}{\textbf{97.45}} & 3.06 & 20.25 & \textcolor{red}{\textbf{17.80}} & \textcolor{red}{\textbf{16.17}} & \textcolor{red}{\textbf{32.10}} & \textcolor{blue}{\textbf{21.12}} & 28.32 & 10 & 02 & \textcolor{red}{\textbf{1.63}} \\
        & TimeRCD & \textcolor{blue}{\textbf{42.47}} & \textcolor{blue}{\textbf{37.98}} & \textcolor{blue}{\textbf{33.74}} & \textcolor{red}{\textbf{53.91}} & 30.28 & \textcolor{blue}{\textbf{24.10}} & \textcolor{blue}{\textbf{50.36}} & 28.17 & 3.06 & 4.90 & 14.17 & 13.29 & \textcolor{blue}{\textbf{29.91}} & 12.50 & 19.86 & 01 & 05 & \textcolor{blue}{\textbf{3.03}} \\
        & DADA $\dagger$ & 34.58 & 31.84 & 30.42 & 40.80 & \textcolor{blue}{\textbf{35.13}} & 7.24 & 48.81 & 74.65 & \textcolor{red}{\textbf{20.19}} & 18.06 & 9.14 & 14.16 & 29.85 & 7.92 & \textcolor{red}{\textbf{99.04}} & 02 & 01 & 3.07 \\
        & MOMENT $\dagger$ & 25.97 & 27.77 & 17.93 & 28.68 & 28.76 & 3.92 & 22.74 & \textcolor{blue}{\textbf{96.00}} & 6.19 & \textcolor{red}{\textbf{41.59}} & \textcolor{blue}{\textbf{15.74}} & \textcolor{blue}{\textbf{15.83}} & 30.77 & \textcolor{red}{\textbf{24.06}} & \textcolor{blue}{\textbf{37.58}} & 02 & 05 & 3.07 \\
        & LOF & 38.97 & 25.58 & 21.81 & 10.13 & 30.62 & 3.00 & 20.00 & 19.00 & \textcolor{blue}{\textbf{13.00}} & \textcolor{blue}{\textbf{26.00}} & 2.00 & 5.00 & 22.00 & 10.00 & 19.00 & 00 & 02 & 4.20 \\
        \hline
        \multirow{5}{*}{Standard-F1}
        & VETime & \textcolor{red}{\textbf{35.24}} & \textcolor{red}{\textbf{27.05}} & \textcolor{red}{\textbf{32.16}} & \textcolor{red}{\textbf{51.41}} & \textcolor{red}{\textbf{55.87}} & \textcolor{blue}{\textbf{11.72}} & \textcolor{blue}{\textbf{41.06}} & \textcolor{red}{\textbf{97.43}} & 2.98 & \textcolor{blue}{\textbf{29.31}} & \textcolor{red}{\textbf{16.74}} & \textcolor{red}{\textbf{15.63}} & \textcolor{red}{\textbf{32.02}} & \textcolor{red}{\textbf{19.88}} & \textcolor{blue}{\textbf{30.43}} & 10 & 04 & \textcolor{red}{\textbf{1.47}} \\
        & TimeRCD & \textcolor{blue}{\textbf{30.66}} & \textcolor{blue}{\textbf{26.00}} & \textcolor{blue}{\textbf{30.48}} & \textcolor{blue}{\textbf{44.89}} & 28.73 & \textcolor{red}{\textbf{11.78}} & \textcolor{blue}{\textbf{41.06}} & 27.56 & 2.98 & 10.00 & \textcolor{blue}{\textbf{8.19}} & 12.82 & \textcolor{blue}{\textbf{29.57}} & 11.59 & 21.57 & 01 & 06 & 2.73 \\
        & DADA $\dagger$ & 22.13 & 24.07 & 26.75 & 34.94 & \textcolor{blue}{\textbf{34.78}} & 4.68 & \textcolor{red}{\textbf{43.67}} & \textcolor{blue}{\textbf{73.11}} & \textcolor{red}{\textbf{27.12}} & 21.95 & 7.13 & \textcolor{blue}{\textbf{13.81}} & 27.69 & 7.18 & \textcolor{red}{\textbf{99.03}} & 03 & 03 & \textcolor{blue}{\textbf{2.60}} \\
        & MOMENT $\dagger$ & 14.43 & 23.83 & 12.92 & 29.78 & 21.30 & 0.00 & 1.77 & 27.65 & \textcolor{blue}{\textbf{4.19}} & \textcolor{red}{\textbf{35.48}} & 7.25 & 8.86 & 6.23 & \textcolor{blue}{\textbf{19.76}} & 13.84 & 01 & 02 & 3.63 \\
        & LOF & 30.65 & 18.80 & 18.70 & 8.41 & 29.08 & 0.00 & 1.00 & 1.00 & 2.00 & 19.00 & 0.00 & 0.00 & 0.00 & 7.00 & 9.00 & 00 & 00 & 4.57 \\
        \hline
        \multirow{5}{*}{VUS-PR}
        & VETime & \textcolor{red}{\textbf{30.51}} & \textcolor{red}{\textbf{23.86}} & 23.70 & \textcolor{red}{\textbf{48.21}} & 42.93 & \textcolor{blue}{\textbf{4.82}} & \textcolor{blue}{\textbf{35.09}} & \textcolor{red}{\textbf{99.77}} & \textcolor{red}{\textbf{3.08}} & 15.37 & \textcolor{red}{\textbf{14.19}} & \textcolor{blue}{\textbf{11.87}} & \textcolor{red}{\textbf{27.11}} & 12.00 & \textcolor{blue}{\textbf{92.15}} & 09 & 06 & \textcolor{red}{\textbf{1.40}} \\
        & TimeRCD & 20.45 & \textcolor{blue}{\textbf{18.69}} & \textcolor{blue}{\textbf{22.68}} & \textcolor{blue}{\textbf{37.03}} & 17.58 & \textcolor{red}{\textbf{4.85}} & 35.01 & 9.61 & 3.07 & 2.64 & 5.53 & 8.61 & \textcolor{blue}{\textbf{23.75}} & 5.56 & 86.53 & 01 & 04 & \textcolor{blue}{\textbf{2.93}} \\
        & DADA $\dagger$ & 12.74 & 17.17 & 20.02 & 25.98 & \textcolor{blue}{\textbf{21.13}} & 1.75 & \textcolor{red}{\textbf{35.69}} & 73.94 & 2.84 & 9.81 & 3.79 & 7.94 & 19.91 & 3.86 & \textcolor{red}{\textbf{99.42}} & 02 & 01 & 3.20 \\
        & MOMENT $\dagger$ & 09.32 & 16.48 & 08.97 & 15.96 & 14.90 & 1.31 & 14.27 & \textcolor{blue}{\textbf{96.19}} & 1.63 & \textcolor{red}{\textbf{17.43}} & \textcolor{blue}{\textbf{6.12}} & \textcolor{blue}{\textbf{10.60}} & 22.28 & \textcolor{red}{\textbf{16.97}} & 26.04 & 02 & 03 & 3.33 \\
        & LOF & \textcolor{blue}{\textbf{24.67}} & 13.58 & 10.59 & 4.40 & 14.50 & 1.00 & 11.00 & 16.00 & \textcolor{red}{\textbf{6.00}} & 10.00 & 3.00 & 5.00 & 16.00 & 6.00 & 15.00 & 01 & 01 & 4.13 \\
        \hline
        \multicolumn{17}{l}{\textbf{VETime Grand Total (Multivariate)}} & \textbf{34} & \textbf{20} & \textbf{1.58} \\
        \hline
    \end{tabular}
    }
\end{table*}

\subsection{Optimization}
The features $F_{AD}$ and $F_{Rec}$ are projected to the original sequence length and fed into separate specific heads. The model is pre-trained on source domains via a multi-task objective:
\begin{equation}
\mathcal{L}_{total} = \mathcal{L}_{BCE} + \mathcal{L}_{MSE} + \lambda_{aw} \mathcal{L}_{aw} + \lambda_{e} \mathcal{L}_{e}
\end{equation}
where $\mathcal{L}_{BCE}$ and $\mathcal{L}_{MSE}$ target anomaly classification and sequence reconstruction; the detailed computation is provided in Appendix \ref{appx.loss}. $\lambda$ denotes the weights balancing the different losses.

\section{Experiments}

\paragraph{Datasets and metrics.}
Our evaluation is conducted on a comprehensive suite of 26 public datasets from the TSB-AD benchmark \citep{liu2024elephant}, comprising 11 univariate (excluding versions derived from multivariate sources to prevent evaluation bias \citep{wu2022cur}) and 15 multivariate datasets. We evaluate performance using four standard metrics: Affiliation-F1, F1-T, Standard-F1, and VUS-PR.
Comprehensive details regarding dataset statistics, formal metric definitions, and baseline configurations are provided in Appendix \ref{appendix:B}.

\subsection{Comparative experiment}

To guarantee a strict zero-shot evaluation and prevent data leakage, VETime is trained on a 0.5-billion-point synthetic corpus \citep{lan2025foundationmodelszeroshottime} generated from mathematical equations, without using target-domain data or statistics. Model selection, early stopping, and hyperparameter tuning are performed exclusively on a held-out synthetic validation set. It is then evaluated directly on unseen target domains without any target-specific fine-tuning (detailed in Appendix~\ref{appendix:setting}). Further evaluations against full-shot models, extended analyses regarding scalability, and diverse anomaly types are detailed in Appendix \ref{appendix:C}.

\begin{wraptable}[21]{r}{0.5\textwidth}
    \centering
    \vspace{-\intextsep}
    \scriptsize
    \setlength{\tabcolsep}{2pt}
    \renewcommand{\arraystretch}{0.9}
    \caption{Performance of VETime against vision-based baselines (SMAP and MSL are univariate settings following VLM4TS\citep{he2025harnessing}).}
    \label{tab:tab2}
    \resizebox{\linewidth}{!}{%
    \begin{tabular}{l l c c c c}
    \toprule
    \textbf{Metric} & \textbf{Method} & \textbf{NAB} & \textbf{YAHOO} & \textbf{SMAP} & \textbf{MSL} \\
    \midrule
    \multirow{5}{1.4cm}{\centering Affiliation\\-F1}
    & VETime & \textcolor{red}{\textbf{88.57}} & \textcolor{red}{\textbf{97.15}} & \textcolor{red}{\textbf{98.83}} & \textcolor{red}{\textbf{92.78}} \\
    & VIT4TS & 61.12 & 60.66  & 91.55 & 65.80 \\
    & VLM4TS & \textcolor{blue}{\textbf{70.37}} & \textcolor{blue}{\textbf{64.25}}&\textcolor{blue}{\textbf{94.76}}  & 66.15 \\
    & VisualTimeAnomaly & 62.46 & 61.03 & 91.65 & \textcolor{blue}{\textbf{86.20}} \\
    & AnomLLM & 58.53 & 36.43 & 42.52 & 46.08 \\
    \hline
    \multirow{5}{1.4cm}{\centering F1\_T}
    & VETime & \textcolor{red}{\textbf{44.31}} & \textcolor{red}{\textbf{91.54}} & \textcolor{red}{\textbf{70.21}} & \textcolor{red}{\textbf{50.38}} \\
    & VIT4TS & 28.29 & 7.61 & 50.97 & 23.48 \\
    & VLM4TS & 30.75 & \textcolor{blue}{\textbf{9.10}} & \textcolor{blue}{\textbf{57.13}} & \textcolor{blue}{\textbf{37.50}} \\
    & VisualTimeAnomaly & \textcolor{blue}{\textbf{41.64}} & 3.02 & 14.80 & 13.84 \\
    & AnomLLM & 17.76 & 1.67 & 4.17 & 3.32 \\
    \hline
    \multirow{5}{1.4cm}{\centering Standard\\-F1}
    & VETime & \textcolor{red}{\textbf{37.87}} & \textcolor{red}{\textbf{88.93}} & \textcolor{red}{\textbf{64.18}} & \textcolor{red}{\textbf{52.18}} \\
    & VIT4TS & 33.81 &7.74  & 51.65 & 27.05 \\
    & VLM4TS & \textcolor{blue}{\textbf{34.78}} & \textcolor{blue}{\textbf{9.16}}& \textcolor{blue}{\textbf{57.56}} & \textcolor{blue}{\textbf{42.21}} \\
    & VisualTimeAnomaly & 31.72 & 3.63 & 16.19 & 16.96 \\
    & AnomLLM & 23.70 & 2.36 & 6.65 & 10.70 \\
    \hline
    \multirow{5}{1.4cm}{\centering VUS-PR}
    & VETime & \textcolor{red}{\textbf{32.98}} & \textcolor{red}{\textbf{88.61}} & \textcolor{red}{\textbf{60.67}} & \textcolor{red}{\textbf{42.81}} \\
    & VIT4TS & 25.78 &31.63  &46.44  & 26.52 \\
    & VLM4TS & \textcolor{blue}{\textbf{27.89}} & \textcolor{blue}{\textbf{33.09}} & \textcolor{blue}{\textbf{52.28}} & \textcolor{blue}{\textbf{33.57}} \\
    & VisualTimeAnomaly & 26.87 & 13.13 & 11.94 & 17.91 \\
    & AnomLLM & 19.50 & 10.44 & 4.92 & 9.77 \\
    \hline
    \multirow{5}{1.4cm}{\centering Per-series \\ runtime(s)}
    &VETime& \textcolor{red}{\textbf{0.04}} & \textcolor{red}{\textbf{0.02}} & \textcolor{red}{\textbf{0.04}} & \textcolor{red}{\textbf{0.02}} \\
    &VIT4TS   & \textcolor{blue}{\textbf{1.72}} & 5.86 & \textcolor{blue}{\textbf{2.23}} & \textcolor{blue}{\textbf{2.75}} \\
    &VLM4TS     & 6.78 & 9.27 & 7.00 & 7.39 \\
    &VisualTimeAnomaly   & 2.52 & \textcolor{blue}{\textbf{2.35}} & 2.34 & 2.85 \\
    &AnomLLM  & 3.59 & 3.91 & 3.40 & 3.95 \\
    \bottomrule
    \end{tabular}%
    }
\end{wraptable}

\paragraph{Comparison with time-series models:}
As shown in Table \ref{tab:tab1}, VETime establishes a new state-of-the-art in zero-shot TSAD, demonstrating superior generalization across both univariate and multivariate settings.
On univariate datasets, VETime secures 24 out of 44 first-place rankings, consistently achieving the lowest average rank of 2.05.
On multivariate datasets, it dominates by securing the top rank in 34 comparison instances. Notably, VETime achieves the best overall rank and the largest number of first-place results in critical metrics like VUS-PR, highlighting its stability independent of threshold selection. Remarkably, it even outperforms full-shot models (detailed in Appendix \ref{appendix:fullshot}), validating its ability to capture universal discriminative anomaly patterns without domain-specific training.

\paragraph{Comparison with vision-based models:}
Due to the high computational cost and token consumption of vision-based models, we evaluate VETime against them on four univariate public datasets, following the protocol of VLM4TS \citep{he2025harnessing}.
As presented in Table \ref{tab:tab2}, VETime demonstrates a decisive advantage in both detection accuracy and computational efficiency. It consistently achieves superior performance, surpassing competing methods across diverse datasets. Furthermore, it achieves an inference speed approximately 100 times faster than its vision-based counterparts, establishing it as a highly practical framework for real-time TSAD.

\paragraph{Comparison under different anomaly types.}
To demonstrate the effectiveness of our method on both local and contextual anomalies, we categorized and reorganized the NAB and YAHOO datasets into Point, Context, and Mixed anomaly types. Our approach is compared with VLM4TS and TimeRCD, and the results are presented in Table \ref{tab:tab10}. By synergizing the fine-grained sensitivity of temporal features with the global semantic context of visual representations, VETime effectively mitigates the limitations of unimodal baselines and achieves consistent superiority across all anomaly categories.

\begin{table*}[b]
\centering
\begin{minipage}[t]{0.40\textwidth}
\centering
\small
\caption{Performance comparison across anomaly types.}
\label{tab:tab10}
\resizebox{\linewidth}{!}{%
\begin{tabular}{llccc}
\toprule
\textbf{Metric} & \textbf{Model} & \textbf{Point} & \textbf{Context} & \makecell{\textbf{Mixed}\\\textbf{anomaly}} \\
\midrule
\multirow{3}{*}{Affiliation-F1}
& VETime    & \textcolor{red}{\textbf{93.20}} & \textcolor{red}{\textbf{91.79}} & \textcolor{red}{\textbf{87.78}} \\
& TimeRCD    & \textcolor{blue}{\textbf{74.84}}&\textcolor{blue}{\textbf{73.84}}&\textcolor{blue}{\textbf{73.03}}\\
& VLM4TS & 67.11&61.94&60.79
 \\

\midrule
\multirow{3}{*}{F1\_T}
& VETime    & \textcolor{red}{\textbf{66.17}} & \textcolor{red}{\textbf{63.32}} & \textcolor{red}{\textbf{52.61}} \\
& TimeRCD    & \textcolor{blue}{\textbf{31.51}}&\textcolor{blue}{\textbf{54.06}}&	\textcolor{blue}{\textbf{42.12}} \\
& VLM4TS & 21.60&35.90&29.63 \\

\midrule
\multirow{3}{*}{Standard-F1}
& VETime    & \textcolor{red}{\textbf{62.33}} & \textcolor{red}{\textbf{59.74}} & \textcolor{red}{\textbf{44.28}} \\
& TimeRCD    & \textcolor{blue}{\textbf{28.34}}&32.18&26.20\\
& VLM4TS & 23.15&\textcolor{blue}{\textbf{35.90}}&\textcolor{blue}{\textbf{34.52}} \\
\midrule
\multirow{3}{*}{VUS-PR}
& VETime    & \textcolor{blue}{\textbf{59.33}} & \textcolor{red}{\textbf{57.59}} & \textcolor{red}{\textbf{44.61}} \\
& TimeRCD    & \textcolor{red}{\textbf{61.85}} & \textcolor{blue}{\textbf{44.93}}&\textcolor{blue}{\textbf{42.90}}\\
& VLM4TS &30.57&32.10&25.61
 \\

\bottomrule
\end{tabular}
}
\end{minipage}%
\hfill
\begin{minipage}[t]{0.56\textwidth}
\centering
\scriptsize
\setlength{\tabcolsep}{2pt}
\caption{Ablation study on NAB, WSD, and YAHOO (Affiliation-F1 / VUS-PR (\%)). Base: TS Encoder + Classifier.
}
\label{tab:A}
\resizebox{\linewidth}{!}{%
\begin{tabular}{%
    >{\raggedright\arraybackslash}p{0.8cm}
    *{4}{>{\centering\arraybackslash}p{0.8cm}}
    *{3}{>{\centering\arraybackslash}p{1.5cm}}
}
\toprule
\textbf{Variant} & \textbf{TIC} & \textbf{PTA} & \textbf{AWCL} & \textbf{TMF} & \textbf{NAB} & \textbf{WSD} & \textbf{YAHOO} \\
\midrule
Base & & & & & 82.48 / 24.32 & 90.33 / 21.77 & 96.65 / 84.41 \\
V1 & $\checkmark$ & & & & 83.31 / 26.99 & 89.53 / 22.76 & 94.97 / 80.33 \\
V2 & $\checkmark$ & $\checkmark$ & & & 82.93 / 27.75 & 90.56 / 24.70 & 95.19 / 83.82 \\
V3 & $\checkmark$ & $\checkmark$ & $\checkmark$ & & 83.99 / 29.52 & 90.11 / 26.30 & 96.86 / 85.31 \\
V4 & $\checkmark$ & $\checkmark$ & & $\checkmark$ & \textcolor{blue}{\textbf{86.84}} / \textcolor{blue}{\textbf{31.23}} & \textcolor{blue}{\textbf{91.63}} / \textcolor{blue}{\textbf{29.00}} & \textcolor{blue}{\textbf{97.11}} / \textcolor{blue}{\textbf{87.26}} \\
\midrule
V5 & $\checkmark$ & & & $\checkmark$ & 83.23 / 24.81 & 90.54 / 24.91 & 96.06 / 82.62 \\
V6 & & $\checkmark$ & & $\checkmark$ & 82.89 / 25.64 & 88.63 / 22.32 & 94.84 / 81.95 \\
V7 & & & $\checkmark$ & $\checkmark$ & 81.84 / 24.86 & 88.34 / 20.20 & 94.12 / 77.68 \\\midrule
\makecell[l]{Full-\\Model} & $\checkmark$ & $\checkmark$ & $\checkmark$ & $\checkmark$ & \textcolor{red}{\textbf{88.57}} / \textcolor{red}{\textbf{32.98}} & \textcolor{red}{\textbf{94.31}} / \textcolor{red}{\textbf{33.85}} & \textcolor{red}{\textbf{97.15}} / \textcolor{red}{\textbf{88.61}} \\
\bottomrule
\end{tabular}
}
\end{minipage}
\vspace{-6pt}
\end{table*}

\subsection{Ablation study}

Table~\ref{tab:A} ablates individual VETime components (see Appendix~\ref{appendix:D} for extended analyses on architectures, mapping strategies, and training objectives), revealing two key design imperatives:

\paragraph{Visual-temporal alignment (TIC + PTA).}
Effective multimodal modeling requires explicit feature alignment rather than naive visual conversion. TIC generates discriminative visual anomaly representations, while PTA remaps these features onto a shared 1D timeline. Removing this foundation (V6 and V7) causes multimodal interaction to fail, often degrading performance below the unimodal baseline.

\paragraph{Dynamic aggregation (AWCL + TMF).}
Merely generating and aligning visual representations (V1 and V2) yields marginal gains. Substantial improvements emerge when AWCL enforces multi-scale discriminability, successfully isolating short-term (local) and long-term (global) anomalies. Concurrently, TMF’s task-adaptive routing dynamically allocates modality importance, significantly outperforming simple concatenation. This synergy (V3 and V4) enables the full model to resolve complex anomalies across challenging datasets like NAB and WSD.

\begin{figure}[t]
    \centering
    \includegraphics[width=0.98\linewidth]{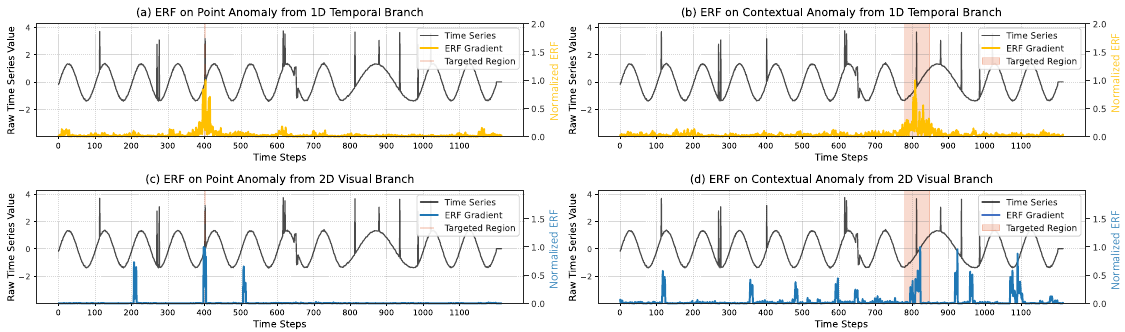}
    \caption{Effective Receptive Field (ERF) analysis comparing 1D temporal and 2D visual branches across different anomaly types. ERFs are computed via tracking the gradient flow from the final anomaly output back to the raw 1D or 2D inputs, with (b) and (d) averaged across the entire targeted region. 1D temporal branch (a, b) is strictly confined to local contexts, whereas 2D visual branch (c, d) exhibits period-aligned attention peaks that capture cross-period global awareness.}
    \label{fig:F}
\end{figure}

\subsection{Model analysis}

This section explores the model's internal mechanisms through interpretability and sensitivity evaluations. Complementary analyses covering feature separability, computational overhead, and expanded visualizations are detailed in Appendices \ref{appendix:C} and \ref{appendix:E}.

\paragraph{Effective receptive field (ERF) analysis:}
Figure~\ref{fig:F} visualizes the ERF of both branches, demonstrating that the visual branch attends to cross-periodic features, thereby overcoming the intrinsic locality of 1D receptive fields. For a marked region, the temporal branch is confined to adjacent segments, while the visual branch attends to long-range features across periods. This stems from aggregating multi-component interactions (via RGB mapping) and structurally converting long-range temporal distances into spatial adjacencies (via folding). This restructuring enhances information routing efficiency and yields semantically dense tokens, providing a global perspective absent in pure 1D models. Further structural analysis is detailed in Appendix \ref{appendix:C4}.

\begin{figure}[t!]
    \centering
    \begin{minipage}[t]{0.42\linewidth}
        \centering
        \includegraphics[width=\linewidth]{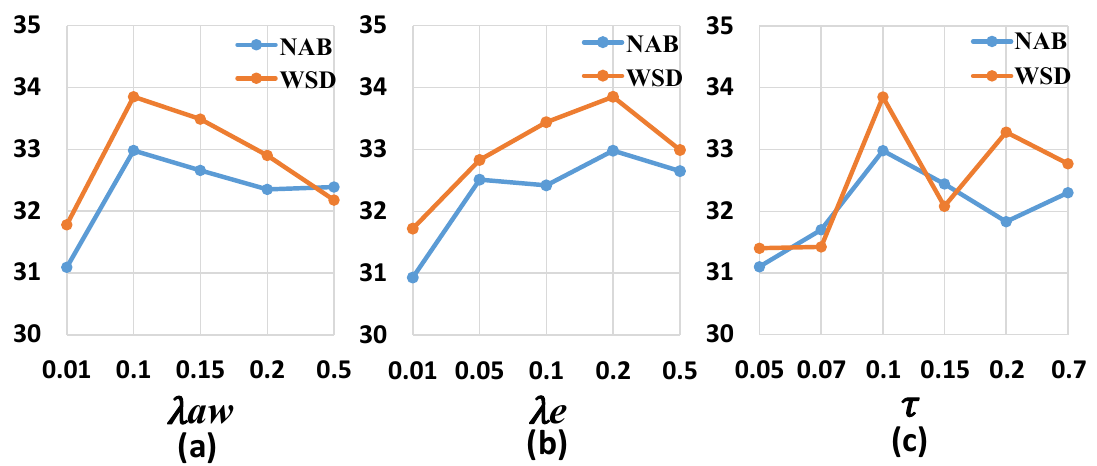}
        \caption{Hyperparameter analysis of $\lambda_{aw}$, $\lambda_{e}$, and $\tau$ under VUS-PR (\%).}
        \label{fig:img5}
    \end{minipage}
    \hfill
    \begin{minipage}[t]{0.5\linewidth}
        \centering
        \includegraphics[width=\linewidth]{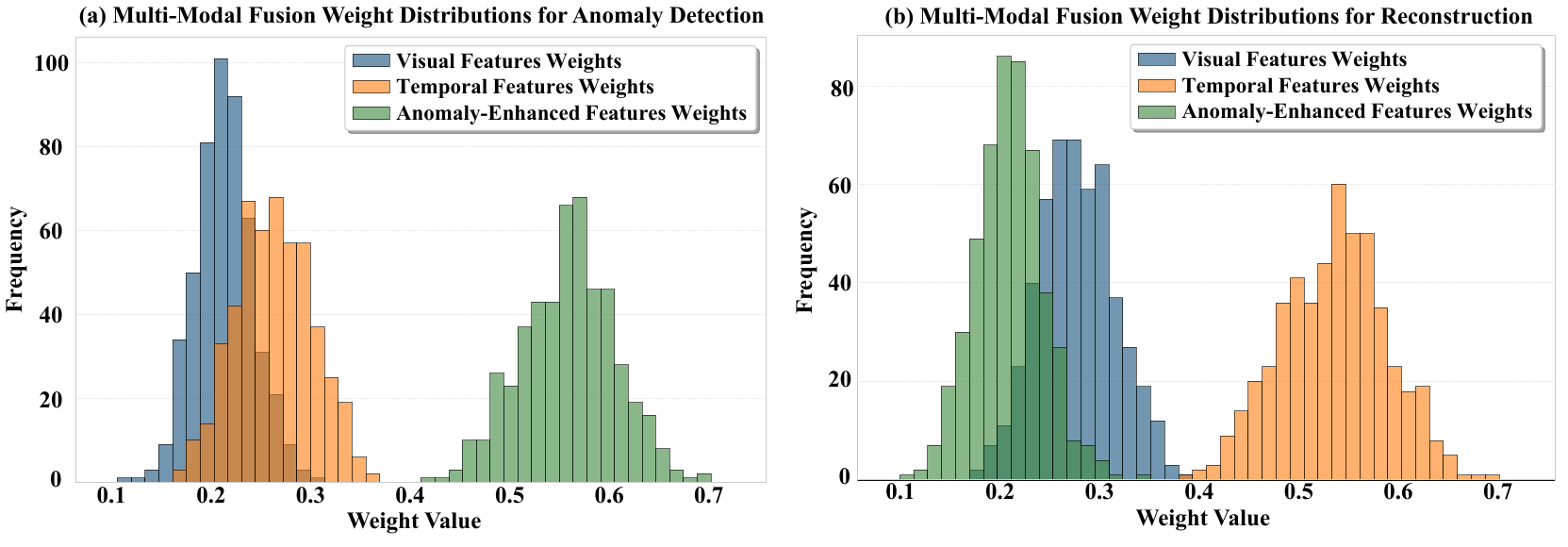}
        \caption{Visualization of fusion weights on the univariate datasets.}
        \label{fig:img6}
    \end{minipage}
\end{figure}

\paragraph{Hyperparameter sensitivity:}
We evaluate the impact of $\lambda_{aw}$, $\lambda_{e}$, and $\tau$ as shown in Figure~\ref{fig:img5}. For $\lambda_{aw}$ and $\tau$ in contrastive learning, deviations from empirical optima degrade performance, as excessive values over-smooth feature distributions. Additionally, $\lambda_{e}$ is crucial for preventing expert collapse; insufficient regularization ($<0.2$) impairs modality balancing and weakens discriminability.

\paragraph{Visualization of fusion weights.} Figure~\ref{fig:img6} reveals a pronounced task divergence in routing weights: Anomaly-Enhanced features dominate detection to maximize discriminability, while Temporal features govern reconstruction to retain numerical continuity. This strategy mitigates expert collapse by effectively separating high-level semantics from low-level precision.

\begin{figure*}[t!]
    \centering
    \includegraphics[width=0.9\linewidth]{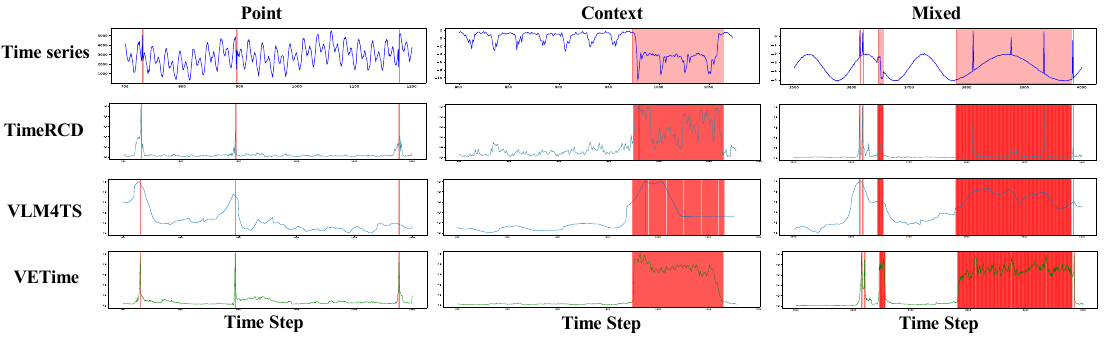}
     \caption{Qualitative anomaly detection results. The top row displays the time series with ground-truth anomalies (red shading), followed by anomaly scores from TimeRCD, VLM4TS, and VETime.}
    \label{fig:img7}
\end{figure*}

\paragraph{Qualitative results:} Figure~\ref{fig:img7} visualizes anomaly scores across Point, Contextual, and Mixed scenarios. VETime consistently outperforms unimodal baselines: it maintains precise point anomaly localization while leveraging global visual semantics to identify long-duration contextual deviations. In complex mixed scenarios, VETime significantly suppresses background noise compared to TimeRCD and rectifies VLM4TS's missed detections, demonstrating robust generalization.

\section{Conclusion and limitations}

We propose VETime, which is the first framework to fuse visual and temporal modalities for zero-shot time-series anomaly detection (TSAD). To bridge the gap between heterogeneous modalities, we introduce a Traceable Image Conversion method combined with Patch-level Temporal Alignment. Furthermore, by leveraging Anomaly Window Contrastive Learning and Task-adaptive Multi-modal Fusion, the model addresses the limitations of unimodal approaches in capturing diverse anomaly patterns. Extensive experiments validate its state-of-the-art zero-shot performance. Future work will incorporate textual modalities to provide semantic explanations for detected anomalies.

\textbf{Limitations.} \textbf{(1) Sensitivity to high-frequency noise:} Reliance on seasonal-trend decomposition makes VETime susceptible to irregular fluctuations where true structure and noise are entangled. \textbf{(2) High-dimensional constraints:} The channel-independent multivariate extension may miss complex inter-variable dependencies, and mapping ultra-high-dimensional data to a fixed visual canvas risks obscuring fine-grained signatures. \textbf{(3) Overhead and interpretability:} Incorporating a vision backbone inevitably increases computational complexity. Moreover, while our empirical evaluations are exhaustive and align with prior works, establishing rigorous theoretical justifications for multi-modal time-series representations remains an inherent, field-wide challenge.

\subsection*{Acknowledgements}
\textbf{Funding Statement}
This work was supported by the NSFC Grant 72271138, Tsinghua University Initiative Scientific Research Program 20243080039, and Tsinghua-Huawei Collaboration Project 20252001894. This work was partially supported by the computing resources from the Tsinghua High-Performance Computing Center and the KaiTuo 1000 supercomputer system.

\textbf{Competing Interests}
The authors declare no competing interests relevant to this work during the 36 months preceding submission.

\bibliographystyle{unsrtnat}
\bibliography{ref}

\newpage
\appendix
\onecolumn

\section{Multivariate model adaptations}\label{appendix:Multivariate}
In the multivariate setting, we adapt the model as follows:

For multivariate time series input, inspired by the proven efficacy of Channel-Independent (CI) architectures in recent time-series foundation models, we extend the aforementioned univariate \textbf{Traceable Image Conversion} to construct a unified three-channel heatmap representation. By treating each variable independently, we avoid the optimization bottlenecks in modeling complex, noisy cross-channel correlations in zero-shot scenarios. Specifically, each variable is first decomposed via Seasonal-Trend decomposition, and subsequently transformed through a variable-specific Gamma correction to apply differentiated non-linear mappings. To strictly preserve the zero-shot constraint, this Gamma parameter is dynamically computed during inference using only the statistics of the current input instance, preventing any leakage of target-domain global priors. This strategy enhances inter-variable discriminability in the visual domain by adapting to their distinct statistical distributions. To prevent distorted periodicities and signal discontinuities at variable boundaries, the folding period is calculated prior to concatenation. Specifically, we estimate the period for each variable independently and select their greatest common divisor as the unified folding period. The resulting individual heatmaps are then concatenated along the temporal dimension to form an integrated multivariate heatmap of size $1 \times (N_v L) \times 3$, where $N_v$ denotes the number of variables and $L$ the temporal length per variable. This composite representation then undergoes the folding and scaling operations. Crucially, to address the high dimensionality where the vertical axis after folding may far exceed the target resolution, an adaptive pooling operation is applied to dynamically downsample the vertical dimension to exactly 224 if it is greater than 224. This resolves the limitations of copy-padding in multivariate scenarios and yields a consistent and unified visual input.

Regarding \textbf{Patch-level Temporal Alignment}, the folding and scaling operations applied to the image features remain unchanged in the multivariate setting; the output is $\hat{F}_{V} \in \mathbb{R}^{N_{TS} \times N_v \times D_V}$. However, we inject an identical positional encoding vector $E_{POS}$ into the token sequence of each variable. Furthermore, we adapt the self-attention mechanism by leveraging the any-variate attention \citep{woo2024moirai}, which processes each variable’s temporal dynamics correction independently. Formally, the output is formulated as $F_{V} \in \mathbb{R}^{(N_{TS} N_v) \times D_{TS}}$. This design enables the model to separately reorganize the temporal order of features for each variable.

The \textbf{Anomaly Window Contrastive Learning} objective is applied independently to the feature sequence of each variable. Specifically, for every variable, we construct its own anomaly context window and perform positive/negative sample selection within its own temporal trajectory.

Finally, the \textbf{Task-adaptive Multi-modal Fusion} module remains unmodified in the multivariate setting, preserving its original architecture and functionality.

\section{Experimental details}\label{appendix:B}
\subsection{Benchmark datasets}\label{appendix:data_details}

Our evaluation uses a selection of datasets from one primary source, the TSB-AD benchmark by \citet{liu2024elephant}, including Univariate Datasets and Multivariate Datasets. The specific characteristics of these datasets, including their domain, number of time series (TS), average length, and anomaly ratio, are summarized in Table~\ref{tab:univariate_datasets} and Table~\ref{tab:multivariate_datasets}.

\begin{table}[h!]
\centering
\caption{Univariate Datasets}\label{tab:univariate_datasets}
\begin{tabular}{lllll}
\toprule
\textbf{Name} & \textbf{Domain} & \textbf{\#TS} & \textbf{Avg Length} & \textbf{AR (\%)} \\
\midrule
UCR & Misc. & 228 & 67818.7 & 0.6 \\
NAB & Web & 28 & 5099.7 & 10.6 \\
YAHOO & Web & 259 & 1560.2 & 0.6 \\
IOPS & Operations & 17 & 72792.3 & 1.3 \\
MGAB & Sensor & 9 & 97777.8 & 0.2 \\
SED & Energy & 3 & 23332.3 & 4.1 \\
TODS & Traffic & 15 & 5000.0 & 6.3 \\
NEK & Weather & 9 & 1073.0 & 8.0 \\
Power & Power Grid & 1 & 35040.0 & 8.5 \\
Stock&Stock&20&15000.0&9.4\\
WSD&Web&111&17444.5&0.6\\
\bottomrule
\end{tabular}
\end{table}

\begin{table}[h!]
\centering
\caption{Multivariate Datasets}\label{tab:multivariate_datasets}
\begin{tabular}{lllll}
\toprule
\textbf{Name} & \textbf{Domain} & \textbf{\#TS} & \textbf{Avg Length} & \textbf{AR (\%)} \\
\midrule
MSL & Space & 16 & 3119.4 & 5.1 \\
PSM & Sensor & 1 & 217624.0 & 11.2 \\
SMAP & Space & 27 & 7855.9 & 2.9 \\
SMD & Server & 22 & 25466.4 & 3.8 \\
SWaT & ICS & 2 & 207457.5 & 12.7 \\
GHL & Reservoir & 25 & 199001.0 & 1.1 \\
Daphnet & Medical & 1 & 38774.0 & 5.9 \\
Exathlon & Cluster & 27 & 60878.4 & 9.8 \\
Genesis & Robotics & 1 & 16220.0 & 0.3 \\
GECCO & Water & 1 & 138521.0 & 1.2 \\
MITDB & Medical & 13 & 336153.8 & 2.7 \\
SVDB & Medical & 31 & 207122.6 & 4.8 \\
LTDB & Medical & 5 & 100000.0 & 15.5 \\
CATSv2 & Simulated & 6 & 240000.0 & 3.7 \\
TAO & Ocean & 13 & 10000.0 & 8.7 \\
\bottomrule
\end{tabular}
\end{table}

\subsection{Baselines}\label{appendix:baselines}
To ensure a fair and consistent evaluation, the window sizes and hyperparameters for all baseline models are strictly configured according to the default settings recommended in their respective original papers.

\textbf{Zero-Shot Models:} These models are pre-trained on large-scale datasets and can be applied directly to new time series without fine-tuning.

\begin{itemize}
    \item \textbf{Time-RCD}: This model proposed by \citep{lan2025foundationmodelszeroshottime} is a pre-trained general anomaly detector. We set the window size to 5000 and patch size is 16.
    \item \textbf{DADA}: This model proposed by \citep{shentu2024towards} is a pre-trained general anomaly detector. We set the window size to 100.
    \item \textbf{TS-Pulse}: An IBM foundation model that uses a TSMixer architecture \citep{ekambaram2025tspulse}. It is pre-trained on a massive corpus for multi-task applications, including anomaly detection. Following the suggestion from the paper, we set the window size with a length of 96.
    \item \textbf{MOMENT}: A foundation model that utilizes a patch-based pre-training strategy to learn rich representations from diverse time series data \citep{goswami2024moment}. We use a window size of 64.
    \item \textbf{TimesFM}: A decoder-only transformer model from Google trained on a large time series corpus using a patching strategy, enabling strong zero-shot generalization \citep{das2024decoder}. We set the window size to 96.
    \item \textbf{Chronos}: A generative model that frames time series analysis as a language modeling task, using a transformer-based architecture to learn and predict time series values \citep{ansari2024chronos}. Its window size is 100.
    \item \textbf{Time MOE}: A decoder-only transformer model with a sparse Mixture-of-Experts (MoE) architecture. It is pre-trained on a large time series corpus for forecasting and multi-task learning \citep{shi2024time}. We set the window size to 96.
\end{itemize}

\textbf{Full-Shot Models:} These models require training on the target dataset.

\begin{itemize}
    \item \textbf{TranAD}: A transformer-based model that uses a reconstructive approach to detect anomalies by comparing original and reconstructed time series \citep{tuli2022tranad}. It is configured with a window size of 10.
    \item \textbf{USAD}: An autoencoder-based model that employs an adversarial training mechanism to enhance its reconstruction capability and anomaly detection \citep{audibert2020usad}. We set the window size to 100.
    \item \textbf{OmniAnomaly}: A deep learning model that uses a Variational Autoencoder (VAE) with a Gated Recurrent Unit (GRU) to learn normal patterns and detect deviations \citep{su2019robust}. The model is configured with a window size of 100.
    \item \textbf{LOF}: A traditional statistical method that measures the local deviation of a data point from its neighbors, identifying anomalies with lower local density \citep{breunig2000lof}. For univariate datasets, we set \texttt{n\_neighbors} to 50, and for multivariate, we set \texttt{n\_neighbors} to 50 with metric as euclidean.
    \item \textbf{IForest}: An ensemble of Isolation Trees that isolates anomalies based on the number of random partitions required to separate them from the rest of the data \citep{liu2008isolation}. For univariate datasets, we set \texttt{n\_estimators} to 200, and for multivariate, we set \texttt{n\_estimators} to 25 and \texttt{max\_features} to 0.8.
\end{itemize}

\textbf{Vision-based Models:} These models use powerful pre-trained models for temporal anomaly detection without the need for additional training. It is important to note that prior vision-based methods do not provide native support for multivariate processing; therefore, our comparative evaluation against these baselines is restricted solely to univariate scenarios.

\begin{itemize}
    \item \textbf{VIT4TS}: A lightweight anomaly screening module that uses a pretrained Vision Transformer to encode time series rendered as line plots from overlapping sliding windows (window length 224, stride 56) and computes cross-window patch-level dissimilarity against a median normal reference to generate high-resolution anomaly heatmaps without fine-tuning \citep{he2025harnessing}.
    \item \textbf{VLM4TS}: A zero-shot time series anomaly detection framework that combines a pretrained vision-language model (GPT-4o) with a two-stage pipeline. It first uses ViT4TS for candidate localization and then prompting the full VLM with a global line plot and textual candidate list to perform semantic validation and refinement of anomalies under a unified multimodal reasoning process \citep{he2025harnessing}.
    \item \textbf{VisualTimeAnomaly}: It converts univariate and multivariate time series into line-plot images (TSIs) and uses off-the-shelf multimodal large language models (GPT-4o) on point-wise, range-wise, and variate-wise anomaly detection tasks. \citep{xu2025can}.
    \item \textbf{AnomLLM}: It involves plotting time series data as line charts and leverages GPT-4o-mini's visual multimodal capabilities to detect anomaly patterns by directly "viewing" the plot, rather than analyzing numerical text. \citep{zhou2024llmsunderstandtimeseries}.
\end{itemize}

\subsection{Evaluation setting and zero-shot protocol}\label{appendix:setting}
To ensure a rigorous and transparent evaluation, we explicitly clarify our zero-shot setting as follows:
\begin{enumerate}
    \item \textbf{Definition of Zero-Shot Setting:} In this work, ``zero-shot'' strictly denotes \textbf{zero target-domain training or fine-tuning}. Following TSFM pre-training protocols \citep{shentu2024towards,lan2025foundationmodelszeroshottime}, VETime is trained solely with anomaly supervision on a synthetically generated corpus (0.5B points) constructed from mathematical equations. This synthetic pre-training incorporates zero statistical priors or labels from target benchmark datasets.
    \item \textbf{Strict Target-Domain Isolation:} All model selection, early stopping, and hyperparameter configurations ($\lambda_{aw}, \lambda_{e}, \tau$) are determined exclusively on a held-out validation set split from our synthetic corpus. Any dataset-specific hyperparameter tuning on the target evaluation domains is strictly prohibited.
    \item \textbf{Unsupervised Anomaly Thresholding:} For metric calculation (e.g., F1 scores), we employ the unsupervised Peaks-Over-Threshold (POT) dynamic thresholding algorithm \citep{sa2017POT}. Anomaly thresholds $\delta$ are adaptively calculated for each individual time series based solely on test-time predicted score distributions, completely without access to target benchmark ground-truth labels.
    \item \textbf{Baseline Evaluation Protocols:} Zero-shot baselines are evaluated using their official pre-trained weights, as retraining unsupervised reconstruction models on anomaly-inclusive synthetic data would invalidate their detection mechanisms.
\end{enumerate}

\subsection{Evaluation metric calculations}\label{appendix:metric}

Our performance evaluation is conducted using four metrics: Affiliation-F1, Temporal-F1 ($F1_T$), Standard-F1 ($F1$), and Volume Under Surface - Precision/Recall (VUS-PR).

\begin{itemize}

\item \textbf{Standard-F1} is a widely used metric that provides a harmonic mean of precision and recall. It is calculated using point-wise True Positives (TP), False Positives (FP), and False Negatives (FN).
\item \textbf{F1-T} as described by \citet{sarfraz2024position}, is a range-based metric that evaluates anomaly detection performance by considering the temporal context of anomalies. It is a variant of the F1-score that addresses common issues in time series evaluation, such as overlapping predictions and temporal proximity.
\item \textbf{Affiliation-F1} is a distance-based metric that measures the "affiliation" or proximity between the ground truth and predicted anomaly points. It is designed to be less sensitive to minor temporal shifts in the predicted anomalies. The score is calculated by finding the optimal one-to-one mapping between the ground truth and detected anomalies and then computing the F1-score based on these affiliations \citep{huet2022local}.
\item \textbf{VUS-PR} is a threshold-independent, parameter-free metric for time series anomaly detection. Unlike point-wise metrics, VUS-PR is robust to time lags and measures the area under a 3D surface plot of precision, recall, and a buffer parameter \citep{paparrizos2022volume}. It addresses the limitations of standard F1 scores by creating a continuous buffer region around each anomaly, thus providing a more reliable and nuanced evaluation of model performance.
\end{itemize}

\subsection{Implementation details} \label{appendix:impl_details}

\paragraph{Implementation:}
The proposed model was implemented using the PyTorch framework on Linux with one H20 24GB GPU. The input time series is divided into patches of size 16. The visual and temporal backbones employ the encoders of the frozen MAE \citep{chen2024visionts} and pre-trained transformer architecture \citep{lan2025foundationmodelszeroshottime}, respectively, with feature dimensions of 768 and 512. These are projected to a shared dimension of 512 during fusion. Hyper-parameters $\lambda_{aw}, \lambda_{e}$, and $\tau$ are fixed at $0.1$, $0.2$, and $0.1$. The model is trained on synthetic data \citep{lan2025foundationmodelszeroshottime} with a batch size of 32, using AdamW optimizer with a learning rate of 5e-4 and a weight decay of 1e-5, for up to 25 epochs, with early stopping if no improvement is observed for 4 consecutive epochs.
For extremely long sequences, VETime follows the sliding-window design of the temporal backbone and processes the input sequentially with a fixed window size of 5000 rather than compressing the full series into one image, which improves scalability with respect to sequence length.

\paragraph{Training data:} To ensure robust zero-shot generalization, we constructed a large-scale synthetic dataset comprising a total of 0.5 billion data points following \citet{lan2025foundationmodelszeroshottime}. The synthetic generation process relies strictly on predefined mathematical equations and does not incorporate any target-domain statistical priors, ensuring absolutely zero exposure to any target-domain data. The dataset features sequences with variable lengths ranging from 500 to 10,000 time steps. It is designed to cover a diverse spectrum of anomaly patterns, encompassing both long-term context deviations (e.g., trend shifts) and fine-grained local anomalies (e.g., point outliers).

\paragraph{Encoders:} Our framework leverages a dual-encoder architecture to process heterogeneous modalities.
\begin{itemize}
\item \textbf{Vision Encoder:} We adopt the encoder from a pre-trained Masked Autoencoder (MAE) as our visual backbone. Specifically, we utilize the lightweight ViT-Base architecture to extract robust semantic representations from the converted time-series images.
\item \textbf{Time-Series Encoder:} We employ a classical Transformer architecture consisting of learnable positional encodings, a stack of Encoder layers with Multi-Head Self-Attention (MHSA), and Feed-Forward Networks (FFN), followed by Layer Normalization. This encoder features 8 transformer layers, each with 8 attention heads. We initialize our encoder using the pre-trained weights from Time-RCD \citep{lan2025foundationmodelszeroshottime}.
\end{itemize}

To achieve parameter-efficient adaptation, we apply Low-Rank Adaptation (LoRA) to fine-tune the linear projection matrices within both the Attention mechanism and the FFN modules in the time-series encoder. Specifically, we set the LoRA rank $r = 8$ and the scaling hyperparameter $\alpha = 16$, resulting in a scaling factor of $\alpha/r = 2$. LoRA adapters are inserted into all trainable linear layers of the encoder, while the original pre-trained weights remain frozen during fine-tuning.

\paragraph{AWCL window size selection:}
The AWCL window mechanism is exclusively active to compute contrastive losses during the synthetic pre-training phase. While its base size is informed by source-domain labels ($L_w = 2L_a$), the synthetic corpus inherently contains highly diverse anomaly lengths ($L_a$ ranging from a few steps to hundreds). Consequently, the absolute physical receptive field of $L_w$ is constantly shifting. This implicitly acts as scale-jittering, forcing AWCL to learn scale-invariant, domain-agnostic representations of local (intra) and global (inter) variations rather than overfitting to a specific curriculum ratio. During inference, the fused feature is simply concatenated from the bimodal features ($F_A = \text{FFN}([Z_{TS}; Z_{V}])$) without constructing windows. Therefore, when deploying to new datasets, there is no need to adjust any window size, ensuring ease of use and full zero-shot reproducibility.

\paragraph{Dual cross-attention:}\label{appx.cross-attention}
To facilitate deep interaction and integrate complementary contexts from the heterogeneous modalities, the aligned time-series features $F_{TS}$ and visual features $F_V$ are fed into a dual cross-attention module. Within this module, features from each modality serve as mutual queries to extract relevant information from the other. Formally, the process is defined as:
\begin{equation}
\begin{aligned}
    Z_{TS}^{0} &= F_{TS} + \text{CrossAttn}(F_{TS}, F_V), & \qquad Z_V^{0} &= F_V + \text{CrossAttn}(F_V, F_{TS}), \\
    Z_{TS} &= Z_{TS}^{0} + \text{FFN}\big(Z_{TS}^{0}\big), & \qquad Z_V &= Z_V^{0} + \text{FFN}\big(Z_V^{0}\big),
\end{aligned}
\end{equation}
where
\begin{equation}
    \text{CrossAttn}(X, Y) = \text{Softmax}\left( \frac{X W_Q (Y W_K)^\top}{\sqrt{d}} \right) (Y W_V),
\end{equation}
denotes the cross-attention mechanism with $X$ serving as queries and $Y$ providing keys and values;
$W_Q$, $W_K$, and $W_V \in \mathbb{R}^{d \times d}$ are learnable projection matrices;
$d$ is the  dimension of the feature embeddings;
and $\text{FFN}(\cdot)$ represents a feed-forward network applied with residual connection.

\paragraph{Loss:} \label{appx.loss}
The task-specific features $F_{AD}$ and $F_{Rec}$, derived from the fusion module, are passed through dedicated prediction heads to generate the predicted anomaly probabilities $\hat{y}_t = \text{Sigmoid}(\text{MLP}_{AD}(F_{AD})) \in (0,1)$ and reconstructed sequence $\hat{x}_t = \text{MLP}_{Rec}(F_{Rec})$, where $\text{MLP}_{AD}$ and $\text{MLP}_{Rec}$ act as the classifier and projector, respectively.

The model is trained end-to-end using a composite loss function balancing classification accuracy, reconstruction quality, and feature alignment:
\begin{equation}
\mathcal{L}_{total} = \mathcal{L}_{BCE} + \mathcal{L}_{MSE} + \lambda_{aw} \mathcal{L}_{aw} + \lambda_{e} \mathcal{L}_{e}
\end{equation}
where $\mathcal{L}_{BCE} = - \frac{1}{L} \sum_{t=1}^{L} \left[ y_t \log(\hat{y}_t) + (1 - y_t) \log(1 - \hat{y}_t) \right]$ is the Binary Cross-Entropy loss over sequence length $L$ with ground-truth binary labels $y_t \in \{0,1\}$, and $\mathcal{L}_{MSE} = \frac{1}{L} \sum_{t=1}^{L} \| x_t - \hat{x}_t \|_2^2$ is the Mean Squared Error loss supervising sequence reconstruction against the normalized input $x_t$.

\paragraph{Inference pipeline:}
\begin{enumerate}
    \item \textbf{Anomaly Score Computation:} Patch-level features $F_{AD}$ are projected to point resolution and mapped via the MLP head to per-timestamp anomaly scores $S_t \in (0, 1)$, relying purely on classification without reconstruction error.
    \item \textbf{Thresholding and Binary Prediction:} Following standard protocols \citep{shentu2024towards,tuli2022tranad}, we employ the unsupervised POT algorithm \citep{sa2017POT} to dynamically calculate thresholds $\delta$ independently for each time series. Specifically, the risk quantile hyperparameter $q$ defining the extreme value distribution is fixed at the standard default of $10^{-4}$ across all evaluations, consistent with the setup in DADA \citep{shentu2024towards}. This adaptive thresholding requires no ground-truth labels or dataset-specific tuning, strictly preventing target-domain label leakage under a fair zero-shot setting.
\end{enumerate}

\section{Additional experimental results}\label{appendix:C}

\subsection{Comparison with Full-Shot Models}\label{appendix:fullshot}
To comprehensively evaluate VETime, we compare its zero-shot performance against state-of-the-art full-shot baselines. Full-shot models are trained directly on the training splits of the target datasets, giving them an inherent advantage in capturing dataset-specific distributions. Remarkably, VETime, operating strictly in a zero-shot manner, achieves the best aggregate rank while remaining competitive across both univariate (Table \ref{tab:tab1_full}) and multivariate (Table \ref{tab:mainm_full}) scenarios.

\begin{table*}[h!]
    \centering
    \renewcommand{\arraystretch}{0.75}
    \caption{Performance of VETime (zero-shot) against full-shot baselines on 11 public univariate datasets. \textcolor{red}{Red}: best, \textcolor{blue}{Blue}: second best.}
    \label{tab:tab1_full}
    \resizebox{\textwidth}{!}{%
    \begin{tabular}
    {
    >{\centering\arraybackslash}p{1.8cm}
    >{\centering\arraybackslash}p{2.0cm}
    *{11}{>{\centering\arraybackslash}p{1.1cm}}
    *{3}{>{\centering\arraybackslash}p{0.8cm}}
    }
        \hline
        \multirow{2}{*}{\textbf{Metric}} & \multirow{2}{*}{\textbf{Model}} & \multicolumn{11}{c}{\textbf{Univariate Datasets}} & \textbf{Total} & \textbf{Total} & \textbf{Avg} \\
        \cline{3-13}
        & & \textbf{IOPS} & \textbf{MGAB} & \textbf{NAB} & \textbf{NEK} & \textbf{Power} & \textbf{SED} & \textbf{Stock} & \textbf{TODS} & \textbf{UCR} & \textbf{WSD} & \textbf{YAHOO} & \textbf{1st} & \textbf{2nd} & \textbf{Rank} \\
        \hline
        \multicolumn{16}{c}{\textbf{Full-Shot Models}} \\
        \hline
        \multirow{6}{*}{Affiliation-F1} & VETime & \textcolor{red}{\textbf{90.53}} & 68.03 & 88.57 & 79.56 & \textcolor{red}{\textbf{78.78}} & \textcolor{red}{\textbf{97.31}} & 69.60 & \textcolor{red}{\textbf{85.85}} & \textcolor{red}{\textbf{85.06}} & \textcolor{red}{\textbf{94.31}} & \textcolor{red}{\textbf{97.15}} & \textcolor{red}{\textbf{07}} & 00 &\textcolor{red}{\textbf{1.91}}\\
        & TranAD & \textcolor{blue}{\textbf{83.19}} & 67.28 & 90.28 & \textcolor{blue}{\textbf{85.02}} & 71.56 & 61.03 & 57.94 & 52.76 & 73.31 & \textcolor{blue}{\textbf{84.34}} & \textcolor{blue}{\textbf{76.08}} & 00 & \textcolor{red}{\textbf{04}}&3.45 \\
        & USAD & 71.08 & 67.81 & \textcolor{blue}{\textbf{91.54}} & 71.13 & 76.48 & 55.60 & 35.92 & 47.90 & \textcolor{blue}{\textbf{76.00}} & 65.10 & 53.05 & 00 & \textcolor{blue}{\textbf{02}}&4.36 \\
        & OmniAnomaly & 80.32 & 67.35 & \textcolor{red}{\textbf{92.35}} & \textcolor{red}{\textbf{86.30}} & \textcolor{blue}{\textbf{78.16}} & 61.26 & \textcolor{red}{\textbf{75.24}} & 50.73 & 73.53 & 78.02 & 71.31 & \textcolor{blue}{\textbf{03}} & 01&\textcolor{blue}{\textbf{3.00}} \\
        & LOF & 81.06 & \textcolor{blue}{\textbf{68.44}} & 75.75 & 84.74 & 66.76 & \textcolor{blue}{\textbf{63.85}} & \textcolor{blue}{\textbf{69.74}} & \textcolor{blue}{\textbf{60.58}} & 73.53 & 81.29 & 75.63 & 00 & \textcolor{red}{\textbf{04}} &3.09\\
        & IForest & 52.81 & \textcolor{red}{\textbf{68.82}} & 39.84 & 71.15 & 00.00 & 70.09 & 0.06 & 44.17 & 50.56 & 41.24 & 33.30 & 01 & 00&5.09 \\
        \hline
        \multirow{6}{*}{F1\_T} & VETime & \textcolor{blue}{\textbf{46.15}} & \textcolor{blue}{\textbf{2.00}} & \textcolor{blue}{\textbf{44.31}} & 60.86 & 20.16 & \textcolor{red}{\textbf{74.35}} & 22.41 & \textcolor{red}{\textbf{68.33}} & \textcolor{red}{\textbf{36.25}} & \textcolor{red}{\textbf{50.42}} & \textcolor{red}{\textbf{91.54}} & \textcolor{red}{\textbf{05}} & \textcolor{blue}{\textbf{03}} &\textcolor{red}{\textbf{2.09}}\\
        & TranAD & 22.63 & 01.65 & 37.28 & 69.97 & 22.36 & 09.57 & 16.73 & 13.51 & 07.75 & 20.94 & 08.41 & 00 & 00 &4.18\\
        & USAD & 20.99 & \textcolor{red}{\textbf{04.07}} & \textcolor{red}{\textbf{61.46}} & \textcolor{blue}{\textbf{70.64}} & \textcolor{red}{\textbf{28.23}} & 09.54 & 16.86 & 20.85 & \textcolor{blue}{\textbf{14.63}} & 14.18 & 09.35 & \textcolor{blue}{\textbf{03}} & 02 &3.18\\
        & OmniAnomaly & \textcolor{red}{\textbf{51.17}} & 01.61 & 40.09 & \textcolor{red}{\textbf{82.20}} & \textcolor{blue}{\textbf{23.48}} & \textcolor{blue}{\textbf{09.68}} & \textcolor{blue}{\textbf{36.22}} & 14.33 & 08.47 & \textcolor{blue}{\textbf{34.79}} & 24.16 & 02 & \textcolor{red}{\textbf{04}}&\textcolor{blue}{\textbf{3.09}} \\
        & LOF & 27.97 & 01.15 & 35.76 & 63.57 & 19.80 & 09.60 & \textcolor{red}{\textbf{66.14}} & \textcolor{blue}{\textbf{31.63}} & 08.31 & 24.38 & \textcolor{blue}{\textbf{55.93}} & 01 & 02&3.45 \\
        & IForest & 07.64 & 00.84 & 21.44 & 65.56 & 00.00 & 09.54 & 1.10 & 11.06 & 06.36 & 4.28 & 04.90 & 00 & 00 &5.00\\
        \hline
        \multirow{6}{*}{Standard-F1} & VETime & 34.56 & \textcolor{blue}{\textbf{1.89}} & \textcolor{blue}{\textbf{37.87}} & 53.95 & 20.12 & \textcolor{red}{\textbf{75.02}} & 23.83 & \textcolor{red}{\textbf{69.13}} & \textcolor{red}{\textbf{30.61}} & \textcolor{red}{\textbf{46.55}} & \textcolor{red}{\textbf{88.93}} & \textcolor{red}{\textbf{05}} & \textcolor{blue}{\textbf{02}}&\textcolor{red}{\textbf{2.00}} \\
        & TranAD & \textcolor{blue}{\textbf{34.85}} & 01.46 & 27.33 & 60.36 & 22.36 & 02.63 & 16.23 & 11.94 & 04.40 & 20.23 & 05.70 & 00 & 01 &3.91\\
        & USAD & 30.66 & \textcolor{red}{\textbf{03.89}} & \textcolor{red}{\textbf{56.15}} & \textcolor{blue}{\textbf{62.91}} & \textcolor{red}{\textbf{28.24}} & 03.41 & 17.99 & 23.87 & \textcolor{blue}{\textbf{10.74}} & 13.20 & 07.21 & \textcolor{blue}{\textbf{03}} & \textcolor{blue}{\textbf{02}}&\textcolor{blue}{\textbf{2.73}} \\
        & OmniAnomaly & \textcolor{red}{\textbf{47.05}} & 01.44 & 28.81 & \textcolor{red}{\textbf{74.03}} & \textcolor{blue}{\textbf{23.50}} & 00.43 & \textcolor{blue}{\textbf{38.59}} & 12.65 & 05.11 & \textcolor{blue}{\textbf{29.57}} & 21.40 & 02 & \textcolor{red}{\textbf{03}} &3.36\\
        & LOF & 30.28 & 01.05 & 24.04 & 56.92 & 12.18 & \textcolor{blue}{\textbf{04.11}} & \textcolor{red}{\textbf{66.20}} & \textcolor{blue}{\textbf{25.77}} & 04.70 & 22.62 & \textcolor{blue}{\textbf{48.95}} & 01 & \textcolor{red}{\textbf{03}}&3.55\\
        & IForest & 08.37 & 00.73 & 29.41 & 58.10 & 19.77 & 03.81 & 16.91 & 13.35 & 04.09 & 2.07 & 03.20 & 00 & 00 &5.09\\
        \hline
        \multirow{6}{*}{VUS-PR} & VETime & \textcolor{red}{\textbf{30.79}} & \textcolor{blue}{\textbf{0.70}} & \textcolor{blue}{\textbf{32.98}} & 57.54 & 11.78 & \textcolor{red}{\textbf{89.05}} & 76.08 & \textcolor{red}{\textbf{94.20}} & \textcolor{red}{\textbf{24.99}} & \textcolor{red}{\textbf{33.85}} & \textcolor{red}{\textbf{88.61}} & \textcolor{red}{\textbf{06}} & \textcolor{blue}{\textbf{02}}&\textcolor{red}{\textbf{2.09}} \\
        & TranAD & 21.61 & 00.64 & 24.82 & \textcolor{blue}{\textbf{61.63}} & 13.04 & 05.75 & 78.08 & 47.33 & 02.25 & 12.20 & 25.78 & 00 & 01&3.73 \\
        & USAD & 16.58 & \textcolor{red}{\textbf{00.75}} & \textcolor{red}{\textbf{55.03}} & 58.53 & \textcolor{red}{\textbf{18.68}} & 04.37 & 74.53 & \textcolor{blue}{\textbf{56.36}} & \textcolor{blue}{\textbf{08.85}} & 10.00 & 14.15 & \textcolor{blue}{\textbf{03}}& \textcolor{blue}{\textbf{02}}&3.27 \\
        & OmniAnomaly & \textcolor{blue}{\textbf{25.35}} & 00.64 & 27.17 & \textcolor{red}{\textbf{74.51}} & \textcolor{blue}{\textbf{14.32}} & 06.20 & \textcolor{red}{\textbf{91.29}} & 45.55 & 02.40 & \textcolor{blue}{\textbf{16.37}} & 29.26 & 02 & \textcolor{red}{\textbf{03}}&\textcolor{blue}{\textbf{2.64}} \\
        & LOF & 19.43 & 00.57 & 21.18 & 58.52 & 09.31 & 06.81 & \textcolor{blue}{\textbf{83.07}} & 49.14 & 02.39 & 12.85 & \textcolor{blue}{\textbf{41.37}} & 00 & \textcolor{blue}{\textbf{02}} &4.18\\
        & IForest & 08.59 & 00.62 & 23.57 & 56.50 & 11.56 & \textcolor{blue}{\textbf{07.71}} & 70.99 & 46.62 & 02.88 & 2.06 & 10.47 & 00 & 01&5.00 \\
        \hline
        \multicolumn{13}{l}{\textbf{Grand Total (Full-Shot)}} & \textbf{23} & \textbf{07} & \textbf{2.02}\\
        \hline
    \end{tabular}
    }
\end{table*}

\begin{table*}[h!]
    \centering
    \renewcommand{\arraystretch}{0.85}
    \caption{Performance of VETime (zero-shot) against full-shot baselines on multivariate datasets. \textcolor{red}{Red}: best, \textcolor{blue}{Blue}: second best.}
    \label{tab:mainm_full}
    \resizebox{\textwidth}{!}{%
    \begin{tabular}{ll ccccccccccccccc cc}
        \hline
        \multirow{2}{*}{\textbf{Metric}} & \multirow{2}{*}{\textbf{Model}} & \multicolumn{5}{c}{\textbf{Multivariate Datasets}} & \multicolumn{10}{c}{\textbf{Additional Datasets}} & \textbf{Total} & \textbf{Total} \\
        \cmidrule(lr){3-7} \cmidrule(lr){8-17}
        & & \textbf{MSL} & \textbf{PSM} & \textbf{SMAP} & \textbf{SMD} & \textbf{SWaT} & \textbf{GHL} & \textbf{Daph.} & \textbf{Exath.} & \textbf{Gen.} & \textbf{GECCO} & \textbf{MIT} & \textbf{SVDB} & \textbf{LTDB} & \textbf{CATS} & \textbf{TAO} & \textbf{1st} & \textbf{2nd} \\
        \hline
        \multicolumn{19}{c}{\textbf{Full-Shot Models}} \\
        \hline
        \multirow{4}{*}{Affiliation-F1}
        & VETime & \textcolor{red}{\textbf{87.02}} & \textcolor{red}{\textbf{77.27}} & 87.06 & \textcolor{blue}{\textbf{88.94}} & \textcolor{red}{\textbf{75.45}} & \textcolor{red}{\textbf{74.93}} & \textcolor{red}{\textbf{84.61}} & \textcolor{red}{\textbf{99.90}} & \textcolor{red}{\textbf{75.41}} & \textcolor{red}{\textbf{87.63}} & \textcolor{red}{\textbf{75.16}} & \textcolor{red}{\textbf{73.03}} & \textcolor{red}{\textbf{78.28}} & 72.08 & \textcolor{red}{\textbf{75.41}} & 12 & 01 \\
        & TranAD & 79.91 & \textcolor{blue}{\textbf{73.83}} & \textcolor{blue}{\textbf{87.39}} & \textcolor{red}{\textbf{92.20}} & \textcolor{blue}{\textbf{75.37}} & 39.00 & \textcolor{blue}{\textbf{72.00}} & \textcolor{blue}{\textbf{94.00}} & \textcolor{blue}{\textbf{69.00}} & \textcolor{blue}{\textbf{26.00}} & \textcolor{blue}{\textbf{71.00}} & 71.00 & 64.00 & \textcolor{red}{\textbf{75.00}} & \textcolor{blue}{\textbf{41.00}} & 02 & 09 \\
        & USAD & 81.86 & 57.86 & 87.25 & 85.09 & 75.06 & 35.95 & 19.82 & 84.96 & 46.84 & 14.51 & 63.28 & 58.33 & 35.81 & 57.04 & 2.05 & 00 & 00 \\
        & OmniAnomaly & \textcolor{blue}{\textbf{83.15}} & 58.17 & \textcolor{red}{\textbf{91.38}} & 85.82 & 73.39 & \textcolor{blue}{\textbf{40.90}} & 67.74 & 93.89 & 68.63 & 18.28 & 70.95 & \textcolor{blue}{\textbf{71.96}} & \textcolor{blue}{\textbf{64.86}} & \textcolor{blue}{\textbf{74.32}} & 20.31 & 01 & 05 \\
        \hline
        \multirow{4}{*}{F1\_T}
        & VETime & 43.52 & \textcolor{red}{\textbf{39.06}} & \textcolor{blue}{\textbf{45.52}} & \textcolor{red}{\textbf{53.36}} & 39.94 & \textcolor{red}{\textbf{24.12}} & \textcolor{red}{\textbf{55.42}} & \textcolor{red}{\textbf{97.45}} & 3.06 & 20.25 & \textcolor{blue}{\textbf{17.80}} & \textcolor{blue}{\textbf{16.17}} & \textcolor{blue}{\textbf{32.10}} & 21.12 & \textcolor{blue}{\textbf{28.32}} & 05 & 05 \\
        & TranAD & 39.42 & 25.49 & 29.12 & 37.98 & \textcolor{blue}{\textbf{49.58}} & 4.00 & 17.00 & 95.00 & \textcolor{red}{\textbf{16.00}} & 40.00 & 16.00 & 15.00 & 30.00 & 24.00 & 18.00 & 01 & 01 \\
        & USAD & \textcolor{blue}{\textbf{48.71}} & 28.96 & 43.94 & 50.41 & \textcolor{red}{\textbf{50.41}} & \textcolor{blue}{\textbf{4.06}} & \textcolor{blue}{\textbf{23.29}} & \textcolor{blue}{\textbf{97.08}} & 2.82 & \textcolor{red}{\textbf{44.79}} & \textcolor{red}{\textbf{28.27}} & \textcolor{red}{\textbf{51.27}} & \textcolor{red}{\textbf{51.31}} & \textcolor{red}{\textbf{35.27}} & 17.31 & 06 & 04 \\
        & OmniAnomaly & \textcolor{red}{\textbf{49.36}} & \textcolor{blue}{\textbf{30.42}} & \textcolor{red}{\textbf{46.63}} & \textcolor{blue}{\textbf{51.84}} & 46.64 & 3.92 & 22.74 & 96.00 & \textcolor{blue}{\textbf{6.19}} & \textcolor{blue}{\textbf{41.59}} & 15.74 & 15.83 & 30.77 & \textcolor{blue}{\textbf{24.06}} & \textcolor{red}{\textbf{37.58}} & 03 & 05 \\
        \hline
        \multirow{4}{*}{Standard-F1}
        & VETime & 35.24 & 27.05 & 32.16 & 51.41 & 55.87 & \textcolor{red}{\textbf{11.72}} & \textcolor{red}{\textbf{41.06}} & \textcolor{red}{\textbf{97.43}} & 2.98 & 29.31 & \textcolor{blue}{\textbf{16.74}} & \textcolor{blue}{\textbf{15.63}} & \textcolor{red}{\textbf{32.02}} & \textcolor{blue}{\textbf{19.88}} & \textcolor{red}{\textbf{30.43}} & 05 & 03 \\
        & TranAD & 29.60 & 25.63 & 25.11 & 43.99 & \textcolor{blue}{\textbf{61.86}} & \textcolor{blue}{\textbf{0.00}} & 1.00 & 37.00 & \textcolor{blue}{\textbf{4.00}} & 33.00 & 7.00 & 7.00 & 5.00 & 14.00 & 5.00 & 00 & 03 \\
        & USAD & \textcolor{blue}{\textbf{38.71}} & \textcolor{blue}{\textbf{28.41}} & \textcolor{blue}{\textbf{38.66}} & \textcolor{blue}{\textbf{53.06}} & \textcolor{red}{\textbf{62.82}} & \textcolor{blue}{\textbf{0.00}} & 0.00 & \textcolor{blue}{\textbf{46.87}} & 0.00 & \textcolor{red}{\textbf{42.82}} & \textcolor{red}{\textbf{20.87}} & \textcolor{red}{\textbf{33.84}} & \textcolor{blue}{\textbf{11.84}} & \textcolor{red}{\textbf{25.87}} & 0.50 & 05 & 07 \\
        & OmniAnomaly & \textcolor{red}{\textbf{39.10}} & \textcolor{red}{\textbf{30.43}} & \textcolor{red}{\textbf{40.50}} & \textcolor{red}{\textbf{57.06}} & 55.93 & \textcolor{blue}{\textbf{0.00}} & \textcolor{blue}{\textbf{1.77}} & 27.65 & \textcolor{red}{\textbf{4.19}} & \textcolor{blue}{\textbf{35.48}} & 7.25 & 8.86 & 6.23 & 19.76 & \textcolor{blue}{\textbf{13.84}} & 05 & 04 \\
        \hline
        \multirow{4}{*}{VUS-PR}
        & VETime & \textcolor{blue}{\textbf{30.51}} & \textcolor{red}{\textbf{23.86}} & 23.70 & \textcolor{red}{\textbf{48.21}} & 42.93 & \textcolor{red}{\textbf{4.82}} & \textcolor{red}{\textbf{35.09}} & \textcolor{red}{\textbf{99.77}} & \textcolor{red}{\textbf{3.08}} & 15.37 & \textcolor{red}{\textbf{14.19}} & \textcolor{blue}{\textbf{11.87}} & \textcolor{blue}{\textbf{27.11}} & 12.00 & \textcolor{red}{\textbf{92.15}} & 08 & 03 \\
        & TranAD & 14.78 & 16.49 & 13.37 & 28.34 & \textcolor{red}{\textbf{47.37}} & 1.00 & 14.00 & 95.00 & \textcolor{blue}{\textbf{3.00}} & \textcolor{red}{\textbf{19.00}} & 6.00 & 9.00 & 21.00 & \textcolor{blue}{\textbf{17.00}} & 9.00 & 02 & 02 \\
        & USAD & 29.95 & 17.59 & \textcolor{blue}{\textbf{26.37}} & 34.53 & \textcolor{blue}{\textbf{44.73}} & 1.30 & 11.54 & \textcolor{blue}{\textbf{96.43}} & 1.17 & 8.37 & \textcolor{blue}{\textbf{12.18}} & \textcolor{red}{\textbf{43.75}} & \textcolor{red}{\textbf{48.84}} & \textcolor{red}{\textbf{20.93}} & 9.60 & 03 & 04 \\
        & OmniAnomaly & \textcolor{red}{\textbf{31.57}} & \textcolor{blue}{\textbf{18.58}} & \textcolor{red}{\textbf{28.07}} & \textcolor{blue}{\textbf{37.44}} & 42.97 & \textcolor{blue}{\textbf{1.31}} & \textcolor{blue}{\textbf{14.27}} & 96.19 & 1.63 & \textcolor{blue}{\textbf{17.43}} & 6.12 & 10.60 & 22.28 & 16.97 & \textcolor{blue}{\textbf{26.04}} & 02 & 06 \\
        \hline
        \multicolumn{17}{l}{\textbf{VETime Grand Total (Full-Shot)}} & \textbf{30} & \textbf{12} \\
        \hline
    \end{tabular}
    }
\end{table*}

\subsection{Statistical support}\label{appendix:statistical_support}
To provide robust statistical evidence for our evaluations, we conduct further analyses covering training stability across multiple random seeds and confidence interval estimations.

\paragraph{Training stability:}
To ensure a rigorous and controlled comparison, we retrained VETime alongside methods that support anomaly supervision (TimeRCD and DADA) on the same synthetic dataset across 5 independent random seeds. As shown in Table~\ref{tab:stat_e}, VETime achieves top-tier overall performance with remarkably low standard deviations (generally $<1\%$), confirming that its empirical gains stem directly from architectural advancements rather than data advantages.

\begin{table*}[h!]
\centering
\small
\setlength{\tabcolsep}{2pt}
\caption{Performance comparison of supervised baselines re-trained on the same synthetic dataset (mean $\pm$ standard deviation over 5 runs). Best result is in \textcolor{red}{red}, second-best is in \textcolor{blue}{blue}.}
\label{tab:stat_e}
\resizebox{\linewidth}{!}{%
\begin{tabular}{l l *{11}{c}}
\toprule
\textbf{Metric} & \textbf{Method} & \textbf{IOPS} & \textbf{MGAB} & \textbf{NAB} & \textbf{NEK} & \textbf{Power} & \textbf{SED} & \textbf{Stock} & \textbf{TODS} & \textbf{UCR} & \textbf{WSD} & \textbf{YAHOO} \\
\midrule
\multirow{3}{*}{Affiliation-F1}
& TimeRCD & \textcolor{blue}{\textbf{83.58}}$\pm$\textcolor{blue}{\textbf{0.62}} & \textcolor{red}{\textbf{70.19}}$\pm$\textcolor{red}{\textbf{1.39}} & 82.17$\pm$1.99 & \textcolor{blue}{\textbf{78.56}}$\pm$\textcolor{blue}{\textbf{2.20}} & \textcolor{red}{\textbf{85.25}}$\pm$\textcolor{red}{\textbf{1.71}} & \textcolor{blue}{\textbf{96.90}}$\pm$\textcolor{blue}{\textbf{0.42}} & \textcolor{blue}{\textbf{71.66}}$\pm$\textcolor{blue}{\textbf{1.05}} & \textcolor{red}{\textbf{86.66}}$\pm$\textcolor{red}{\textbf{2.60}} & \textcolor{blue}{\textbf{84.78}}$\pm$\textcolor{blue}{\textbf{0.40}} & \textcolor{blue}{\textbf{90.12}}$\pm$\textcolor{blue}{\textbf{0.23}} & \textcolor{blue}{\textbf{96.54}}$\pm$\textcolor{blue}{\textbf{0.37}} \\
& DADA & 73.00$\pm$0.92 & 66.28$\pm$1.66 & \textcolor{blue}{\textbf{84.51}}$\pm$\textcolor{blue}{\textbf{0.63}} & 78.20$\pm$1.81 & 68.05$\pm$0.87 & 68.20$\pm$1.58 & \textcolor{red}{\textbf{96.55}}$\pm$\textcolor{red}{\textbf{0.49}} & 66.31$\pm$0.32 & 66.36$\pm$0.23 & 86.10$\pm$0.73 & 90.46$\pm$0.64 \\
& VETime & \textcolor{red}{\textbf{90.22}}$\pm$\textcolor{red}{\textbf{0.66}} & \textcolor{blue}{\textbf{68.18}}$\pm$\textcolor{blue}{\textbf{0.18}} & \textcolor{red}{\textbf{88.19}}$\pm$\textcolor{red}{\textbf{0.43}} & \textcolor{red}{\textbf{79.95}}$\pm$\textcolor{red}{\textbf{0.97}} & \textcolor{blue}{\textbf{78.99}}$\pm$\textcolor{blue}{\textbf{0.33}} & \textcolor{red}{\textbf{97.95}}$\pm$\textcolor{red}{\textbf{0.98}} & 69.45$\pm$0.61 & \textcolor{blue}{\textbf{85.43}}$\pm$\textcolor{blue}{\textbf{0.54}} & \textcolor{red}{\textbf{85.06}}$\pm$\textcolor{red}{\textbf{0.24}} & \textcolor{red}{\textbf{94.28}}$\pm$\textcolor{red}{\textbf{0.06}} & \textcolor{red}{\textbf{97.12}}$\pm$\textcolor{red}{\textbf{0.16}} \\
\midrule
\multirow{3}{*}{VUS-PR}
& TimeRCD & 21.82$\pm$3.79 & \textcolor{red}{\textbf{1.03}}$\pm$\textcolor{red}{\textbf{0.06}} & 24.42$\pm$1.35 & 27.47$\pm$2.65 & \textcolor{red}{\textbf{21.17}}$\pm$\textcolor{red}{\textbf{0.15}} & \textcolor{blue}{\textbf{80.38}}$\pm$\textcolor{blue}{\textbf{0.90}} & \textcolor{blue}{\textbf{76.72}}$\pm$\textcolor{blue}{\textbf{1.45}} & \textcolor{blue}{\textbf{93.29}}$\pm$\textcolor{blue}{\textbf{1.69}} & \textcolor{blue}{\textbf{23.08}}$\pm$\textcolor{blue}{\textbf{0.07}} & 21.94$\pm$2.71 & \textcolor{blue}{\textbf{84.23}}$\pm$\textcolor{blue}{\textbf{0.75}} \\
& DADA & \textcolor{blue}{\textbf{24.77}}$\pm$\textcolor{blue}{\textbf{1.85}} & 0.62$\pm$0.04 & \textcolor{blue}{\textbf{24.46}}$\pm$\textcolor{blue}{\textbf{0.79}} & \textcolor{red}{\textbf{64.82}}$\pm$\textcolor{red}{\textbf{2.41}} & \textcolor{blue}{\textbf{12.42}}$\pm$\textcolor{blue}{\textbf{0.64}} & 6.43$\pm$0.26 & \textcolor{red}{\textbf{99.79}}$\pm$\textcolor{red}{\textbf{0.02}} & 57.47$\pm$0.11 & 2.12$\pm$0.65 & \textcolor{blue}{\textbf{30.38}}$\pm$\textcolor{blue}{\textbf{1.05}} & 73.05$\pm$0.47 \\
& VETime & \textcolor{red}{\textbf{30.84}}$\pm$\textcolor{red}{\textbf{1.62}} & \textcolor{blue}{\textbf{0.69}}$\pm$\textcolor{blue}{\textbf{0.02}} & \textcolor{red}{\textbf{32.84}}$\pm$\textcolor{red}{\textbf{0.21}} & \textcolor{blue}{\textbf{56.42}}$\pm$\textcolor{blue}{\textbf{2.71}} & 11.85$\pm$0.18 & \textcolor{red}{\textbf{88.37}}$\pm$\textcolor{red}{\textbf{2.24}} & 76.21$\pm$0.78 & \textcolor{red}{\textbf{93.82}}$\pm$\textcolor{red}{\textbf{0.53}} & \textcolor{red}{\textbf{24.67}}$\pm$\textcolor{red}{\textbf{0.78}} & \textcolor{red}{\textbf{33.16}}$\pm$\textcolor{red}{\textbf{0.65}} & \textcolor{red}{\textbf{87.92}}$\pm$\textcolor{red}{\textbf{1.06}} \\
\bottomrule
\end{tabular}%
}
\end{table*}

\paragraph{Confidence intervals:}
We calculated the 95\% confidence intervals (CIs) using the sequence-level Bootstrapping method, which resamples intact individual time series with replacement 1000 times per dataset. The detailed performance across zero-shot, full-shot, and vision-based baselines is summarized in Table~\ref{tab:stat_c} and Table~\ref{tab:stat_d}. Across the 11 univariate datasets, we performed Friedman tests followed by two-sided Wilcoxon signed-rank tests with Holm correction. A significant method effect was found for full-shot Affiliation-F1 ($p = 0.0076$), where VETime significantly outperforms TranAD ($p_{\text{Holm}} = 0.0488$) and USAD ($p_{\text{Holm}} = 0.0146$). For other metrics and panels, VETime consistently achieves the best average rank while demonstrating sustained advantages across diverse benchmarks.

\begin{table*}[h!]
\centering
\small
\setlength{\tabcolsep}{2pt}
\caption{Performance of VETime against zero-shot and full-shot time-series baselines across 11 datasets (Mean $\pm$ 95\% CI). Best result is in \textcolor{red}{red}, second-best is in \textcolor{blue}{blue}.}
\label{tab:stat_c}
\resizebox{\linewidth}{!}{%
\begin{tabular}{l l *{11}{c}}
\toprule
\textbf{Metric} & \textbf{Method} & \textbf{IOPS} & \textbf{MGAB} & \textbf{NAB} & \textbf{NEK} & \textbf{Power} & \textbf{SED} & \textbf{Stock} & \textbf{TODS} & \textbf{UCR} & \textbf{WSD} & \textbf{YAHOO} \\
\midrule
\multicolumn{13}{c}{\textbf{Zero-Shot Models}} \\
\midrule
\multirow{5}{*}{Affiliation-F1}
& VETime & \textcolor{red}{\textbf{90.39$\pm$3.10}} & \textcolor{blue}{\textbf{68.17$\pm$0.61}} & \textcolor{blue}{\textbf{88.36$\pm$3.38}} & 74.96$\pm$17.28 & \textcolor{blue}{\textbf{78.78$\pm$0.00}} & \textcolor{red}{\textbf{97.08$\pm$1.34}} & 69.63$\pm$0.54 & 85.84$\pm$5.26 & \textcolor{red}{\textbf{85.08$\pm$1.54}} & 93.25$\pm$1.48 & \textcolor{red}{\textbf{97.07$\pm$1.79}} \\
& TimeRCD & 83.37$\pm$3.67 & \textcolor{red}{\textbf{70.71$\pm$1.85}} & 83.07$\pm$9.54 & 82.39$\pm$12.56 & \textcolor{red}{\textbf{85.51$\pm$0.00}} & \textcolor{blue}{\textbf{96.48$\pm$1.79}} & 71.77$\pm$1.49 & \textcolor{blue}{\textbf{86.58$\pm$8.30}} & \textcolor{blue}{\textbf{84.68$\pm$1.86}} & 90.36$\pm$2.70 & \textcolor{blue}{\textbf{96.75$\pm$1.29}} \\
& DADA & \textcolor{blue}{\textbf{89.61$\pm$1.51}} & 67.70$\pm$0.40 & 87.51$\pm$6.79 & \textcolor{red}{\textbf{95.79$\pm$0.13}} & 69.79$\pm$0.00 & 65.18$\pm$2.03 & \textcolor{red}{\textbf{98.83$\pm$0.17}} & 76.91$\pm$12.69 & 72.20$\pm$0.83 & \textcolor{blue}{\textbf{93.81$\pm$2.72}} & 92.33$\pm$3.51 \\
& MOMENT & 87.60$\pm$2.02 & 67.08$\pm$3.42 & \textcolor{red}{\textbf{91.18$\pm$4.32}} & 91.53$\pm$5.67 & 75.97$\pm$0.00 & 58.19$\pm$4.52 & 45.28$\pm$0.91 & 59.76$\pm$13.08 & 75.61$\pm$2.55 & \textcolor{red}{\textbf{95.46$\pm$1.86}} & 79.81$\pm$3.79 \\
& TimesFM & 81.93$\pm$3.85 & 66.84$\pm$0.95 & 79.63$\pm$3.86 & \textcolor{blue}{\textbf{95.17$\pm$1.76}} & 69.88$\pm$0.00 & 66.18$\pm$3.40 & \textcolor{blue}{\textbf{97.51$\pm$0.42}} & \textcolor{red}{\textbf{88.91$\pm$4.93}} & 69.97$\pm$1.22 & 78.97$\pm$1.75 & 91.34$\pm$1.81 \\
\midrule
\multirow{5}{*}{$F1_T$}
& VETime & \textcolor{blue}{\textbf{44.20$\pm$8.13}} & \textcolor{red}{\textbf{2.01$\pm$0.51}} & \textcolor{blue}{\textbf{44.22$\pm$6.53}} & \textcolor{blue}{\textbf{62.28$\pm$12.04}} & \textcolor{blue}{\textbf{20.16$\pm$0.00}} & \textcolor{red}{\textbf{76.05$\pm$7.68}} & 22.45$\pm$1.35 & \textcolor{red}{\textbf{68.56$\pm$8.75}} & \textcolor{blue}{\textbf{36.42$\pm$3.95}} & \textcolor{red}{\textbf{50.28$\pm$3.88}} & \textcolor{red}{\textbf{91.54$\pm$1.65}} \\
& TimeRCD & 27.96$\pm$6.46 & 1.73$\pm$0.38 & 36.15$\pm$5.99 & 34.58$\pm$13.81 & \textcolor{red}{\textbf{28.47$\pm$0.00}} & \textcolor{blue}{\textbf{70.34$\pm$3.89}} & 31.17$\pm$2.37 & \textcolor{blue}{\textbf{64.97$\pm$9.65}} & 32.60$\pm$3.37 & 35.80$\pm$3.94 & \textcolor{blue}{\textbf{85.54$\pm$2.63}} \\
& DADA & 42.01$\pm$7.18 & \textcolor{blue}{\textbf{1.82$\pm$0.34}} & 37.79$\pm$5.35 & 48.88$\pm$15.55 & 19.80$\pm$0.00 & 9.46$\pm$1.67 & \textcolor{red}{\textbf{94.63$\pm$1.85}} & 34.39$\pm$4.91 & 6.88$\pm$1.15 & \textcolor{blue}{\textbf{48.28$\pm$3.89}} & 79.26$\pm$2.83 \\
& MOMENT & 31.35$\pm$9.57 & 0.77$\pm$0.10 & \textcolor{red}{\textbf{50.81$\pm$7.45}} & \textcolor{red}{\textbf{63.13$\pm$9.90}} & 19.91$\pm$0.00 & 8.62$\pm$1.03 & 19.73$\pm$2.91 & 16.06$\pm$2.55 & 13.75$\pm$2.67 & 41.21$\pm$3.70 & 10.88$\pm$1.50 \\
& TimesFM & \textcolor{red}{\textbf{45.48$\pm$8.65}} & 1.02$\pm$0.24 & 34.54$\pm$6.73 & 38.38$\pm$11.31 & 19.80$\pm$0.00 & 9.69$\pm$2.15 & \textcolor{blue}{\textbf{88.98$\pm$1.51}} & 52.17$\pm$9.34 & \textcolor{red}{\textbf{41.94$\pm$3.21}} & 10.10$\pm$2.07 & 83.89$\pm$2.59 \\
\midrule
\multirow{5}{*}{Standard-F1}
& VETime & \textcolor{blue}{\textbf{35.01$\pm$7.54}} & \textcolor{red}{\textbf{1.89$\pm$0.49}} & \textcolor{blue}{\textbf{38.06$\pm$5.95}} & \textcolor{blue}{\textbf{52.16$\pm$15.88}} & \textcolor{blue}{\textbf{20.12$\pm$0.00}} & \textcolor{red}{\textbf{76.71$\pm$7.70}} & 23.87$\pm$1.63 & \textcolor{red}{\textbf{69.39$\pm$8.49}} & \textcolor{red}{\textbf{30.71$\pm$3.92}} & \textcolor{red}{\textbf{46.55$\pm$4.08}} & \textcolor{red}{\textbf{88.85$\pm$2.52}} \\
& TimeRCD & 22.61$\pm$4.14 & \textcolor{blue}{\textbf{1.40$\pm$0.32}} & 27.20$\pm$4.71 & 32.03$\pm$11.17 & \textcolor{red}{\textbf{28.59$\pm$0.00}} & \textcolor{blue}{\textbf{69.04$\pm$2.48}} & 32.51$\pm$0.86 & \textcolor{blue}{\textbf{66.97$\pm$7.47}} & 27.78$\pm$3.28 & 29.48$\pm$3.32 & \textcolor{blue}{\textbf{87.48$\pm$3.38}} \\
& DADA & 32.70$\pm$1.22 & 0.96$\pm$0.32 & 28.73$\pm$3.74 & 47.22$\pm$6.94 & 15.99$\pm$0.00 & 2.56$\pm$0.23 & \textcolor{red}{\textbf{95.39$\pm$1.50}} & 28.35$\pm$2.35 & 3.90$\pm$0.56 & \textcolor{blue}{\textbf{45.21$\pm$4.16}} & 81.80$\pm$4.21 \\
& MOMENT & 29.90$\pm$7.38 & 0.66$\pm$0.05 & \textcolor{red}{\textbf{44.35$\pm$7.68}} & \textcolor{red}{\textbf{63.99$\pm$11.45}} & 16.39$\pm$0.00 & 3.34$\pm$0.15 & 19.88$\pm$0.49 & 14.18$\pm$3.37 & 9.59$\pm$2.29 & 41.05$\pm$3.34 & 10.52$\pm$0.76 \\
& TimesFM & \textcolor{red}{\textbf{35.95$\pm$6.34}} & 0.98$\pm$0.22 & 27.15$\pm$6.03 & 34.86$\pm$5.82 & 16.73$\pm$0.00 & 2.02$\pm$0.12 & \textcolor{blue}{\textbf{89.47$\pm$1.77}} & 37.20$\pm$10.14 & 8.27$\pm$2.74 & 37.05$\pm$3.25 & 84.76$\pm$3.02 \\
\midrule
\multirow{5}{*}{VUS-PR}
& VETime & \textcolor{blue}{\textbf{31.78$\pm$10.95}} & 0.71$\pm$0.15 & \textcolor{blue}{\textbf{33.13$\pm$7.59}} & \textcolor{blue}{\textbf{55.42$\pm$16.37}} & \textcolor{blue}{\textbf{11.78$\pm$0.00}} & \textcolor{red}{\textbf{89.98$\pm$4.68}} & 75.32$\pm$10.69 & \textcolor{red}{\textbf{93.89$\pm$2.76}} & \textcolor{red}{\textbf{25.20$\pm$4.03}} & 33.81$\pm$4.68 & \textcolor{red}{\textbf{88.60$\pm$2.82}} \\
& TimeRCD & 19.85$\pm$6.20 & \textcolor{red}{\textbf{1.03$\pm$0.28}} & 24.15$\pm$7.91 & 27.81$\pm$12.61 & \textcolor{red}{\textbf{21.25$\pm$0.00}} & \textcolor{blue}{\textbf{80.94$\pm$2.46}} & \textcolor{blue}{\textbf{77.61$\pm$11.69}} & \textcolor{blue}{\textbf{93.63$\pm$2.57}} & 23.02$\pm$3.83 & 21.52$\pm$3.82 & 84.46$\pm$3.26 \\
& DADA & 24.43$\pm$7.39 & 0.57$\pm$0.06 & 24.49$\pm$4.05 & 49.47$\pm$22.32 & 10.61$\pm$0.00 & 6.58$\pm$1.06 & \textcolor{red}{\textbf{99.52$\pm$0.14}} & 64.95$\pm$12.18 & 2.89$\pm$0.41 & 33.35$\pm$3.86 & 70.78$\pm$4.82 \\
& MOMENT & \textcolor{red}{\textbf{36.95$\pm$10.90}} & 0.55$\pm$0.05 & \textcolor{red}{\textbf{45.28$\pm$9.09}} & \textcolor{red}{\textbf{68.21$\pm$12.20}} & 10.50$\pm$0.00 & 4.27$\pm$0.46 & 77.25$\pm$10.59 & 57.00$\pm$11.36 & 6.12$\pm$1.72 & \textcolor{red}{\textbf{55.51$\pm$5.07}} & 30.72$\pm$3.30 \\
& TimesFM & 18.79$\pm$11.64 & 0.59$\pm$0.14 & 23.64$\pm$8.22 & 38.78$\pm$17.28 & 10.44$\pm$0.00 & 8.01$\pm$2.61 & \textcolor{blue}{\textbf{98.41$\pm$0.43}} & 73.01$\pm$4.13 & 5.90$\pm$4.01 & 21.43$\pm$4.95 & \textcolor{blue}{\textbf{86.77$\pm$3.37}} \\
\midrule
\multicolumn{13}{c}{\textbf{Full-Shot Models}} \\
\midrule
\multirow{4}{*}{Affiliation-F1}
& VETime & \textcolor{red}{\textbf{90.39$\pm$3.10}} & \textcolor{red}{\textbf{68.17$\pm$0.61}} & 88.36$\pm$3.38 & 74.96$\pm$17.28 & \textcolor{red}{\textbf{78.78$\pm$0.00}} & \textcolor{red}{\textbf{97.08$\pm$1.34}} & \textcolor{blue}{\textbf{69.63$\pm$0.54}} & \textcolor{red}{\textbf{85.84$\pm$5.26}} & \textcolor{red}{\textbf{85.08$\pm$1.54}} & \textcolor{red}{\textbf{93.25$\pm$1.48}} & \textcolor{red}{\textbf{97.07$\pm$1.79}} \\
& TranAD & \textcolor{blue}{\textbf{83.71$\pm$11.01}} & 67.35$\pm$0.44 & 90.52$\pm$6.41 & \textcolor{blue}{\textbf{85.77$\pm$6.84}} & 71.56$\pm$0.00 & 61.03$\pm$6.15 & 58.05$\pm$4.84 & \textcolor{blue}{\textbf{51.81$\pm$10.95}} & 73.29$\pm$1.32 & \textcolor{blue}{\textbf{84.16$\pm$3.10}} & \textcolor{blue}{\textbf{76.12$\pm$3.71}} \\
& USAD & 70.05$\pm$16.57 & \textcolor{blue}{\textbf{67.79$\pm$1.45}} & \textcolor{blue}{\textbf{92.32$\pm$3.36}} & 70.44$\pm$15.64 & 76.48$\pm$0.00 & 55.60$\pm$4.76 & 35.93$\pm$0.57 & 45.18$\pm$12.40 & \textcolor{blue}{\textbf{75.99$\pm$3.17}} & 65.33$\pm$6.61 & 52.57$\pm$6.88 \\
& OmniAnomaly & 80.95$\pm$11.65 & 67.40$\pm$0.51 & \textcolor{red}{\textbf{92.44$\pm$6.13}} & \textcolor{red}{\textbf{87.10$\pm$8.79}} & \textcolor{blue}{\textbf{78.16$\pm$0.00}} & \textcolor{blue}{\textbf{61.26$\pm$3.16}} & \textcolor{red}{\textbf{75.28$\pm$3.83}} & 50.74$\pm$5.92 & 73.46$\pm$1.38 & 77.93$\pm$5.24 & 71.18$\pm$3.55 \\
\midrule
\multirow{4}{*}{$F1_T$}
& VETime & \textcolor{blue}{\textbf{44.20$\pm$8.13}} & \textcolor{blue}{\textbf{2.01$\pm$0.51}} & \textcolor{blue}{\textbf{44.22$\pm$6.53}} & 62.28$\pm$12.04 & 20.16$\pm$0.00 & \textcolor{red}{\textbf{76.05$\pm$7.68}} & \textcolor{blue}{\textbf{22.45$\pm$1.35}} & \textcolor{red}{\textbf{68.56$\pm$8.75}} & \textcolor{red}{\textbf{36.42$\pm$3.95}} & \textcolor{red}{\textbf{50.28$\pm$3.88}} & \textcolor{red}{\textbf{91.54$\pm$1.65}} \\
& TranAD & 26.77$\pm$11.70 & 1.47$\pm$0.30 & 35.04$\pm$3.76 & 63.45$\pm$6.34 & 22.36$\pm$0.00 & 8.13$\pm$1.03 & 17.66$\pm$3.22 & 12.68$\pm$1.47 & 7.03$\pm$0.87 & 21.66$\pm$3.79 & 8.79$\pm$1.43 \\
& USAD & 21.61$\pm$12.58 & \textcolor{red}{\textbf{4.28$\pm$0.70}} & \textcolor{red}{\textbf{61.90$\pm$7.73}} & \textcolor{blue}{\textbf{72.68$\pm$16.15}} & \textcolor{red}{\textbf{28.23$\pm$0.00}} & \textcolor{blue}{\textbf{9.12$\pm$1.03}} & 16.67$\pm$3.10 & \textcolor{blue}{\textbf{23.89$\pm$6.86}} & \textcolor{blue}{\textbf{13.47$\pm$2.85}} & 14.32$\pm$3.55 & 9.27$\pm$1.43 \\
& OmniAnomaly & \textcolor{red}{\textbf{48.57$\pm$12.92}} & 1.33$\pm$0.27 & 36.36$\pm$3.43 & \textcolor{red}{\textbf{81.01$\pm$9.07}} & \textcolor{blue}{\textbf{23.48$\pm$0.00}} & 8.73$\pm$1.03 & \textcolor{red}{\textbf{37.52$\pm$1.54}} & 13.28$\pm$1.36 & 8.64$\pm$1.06 & \textcolor{blue}{\textbf{32.90$\pm$6.02}} & \textcolor{blue}{\textbf{19.65$\pm$3.53}} \\
\midrule
\multirow{4}{*}{Standard-F1}
& VETime & \textcolor{red}{\textbf{35.01$\pm$7.54}} & \textcolor{blue}{\textbf{1.89$\pm$0.49}} & \textcolor{blue}{\textbf{38.06$\pm$5.95}} & 52.16$\pm$15.88 & 20.12$\pm$0.00 & \textcolor{red}{\textbf{76.71$\pm$7.70}} & \textcolor{blue}{\textbf{23.87$\pm$1.63}} & \textcolor{red}{\textbf{69.39$\pm$8.49}} & \textcolor{red}{\textbf{30.71$\pm$3.92}} & \textcolor{red}{\textbf{46.55$\pm$4.08}} & \textcolor{red}{\textbf{88.85$\pm$2.52}} \\
& TranAD & \textcolor{blue}{\textbf{31.97$\pm$9.90}} & 1.43$\pm$0.27 & 27.85$\pm$3.36 & 60.02$\pm$7.47 & 22.36$\pm$0.00 & 2.34$\pm$0.30 & 16.12$\pm$0.94 & 10.52$\pm$1.33 & 4.43$\pm$0.52 & 19.93$\pm$3.19 & 5.42$\pm$0.76 \\
& USAD & 22.68$\pm$12.58 & \textcolor{red}{\textbf{2.27$\pm$1.18}} & \textcolor{red}{\textbf{52.12$\pm$8.68}} & \textcolor{blue}{\textbf{64.55$\pm$12.33}} & \textcolor{red}{\textbf{28.24$\pm$0.00}} & \textcolor{blue}{\textbf{3.44$\pm$0.52}} & 17.40$\pm$0.71 & \textcolor{blue}{\textbf{24.39$\pm$5.09}} & \textcolor{blue}{\textbf{9.77$\pm$2.42}} & 12.92$\pm$3.05 & 7.38$\pm$0.88 \\
& OmniAnomaly & 32.39$\pm$14.05 & 1.22$\pm$0.25 & 26.03$\pm$4.12 & \textcolor{red}{\textbf{72.42$\pm$5.26}} & \textcolor{blue}{\textbf{23.50$\pm$0.00}} & 0.43$\pm$0.13 & \textcolor{red}{\textbf{38.03$\pm$2.10}} & 12.97$\pm$0.44 & 5.58$\pm$0.59 & \textcolor{blue}{\textbf{28.22$\pm$5.16}} & \textcolor{blue}{\textbf{21.52$\pm$2.62}} \\
\midrule
\multirow{4}{*}{VUS-PR}
& VETime & \textcolor{red}{\textbf{31.78$\pm$10.95}} & \textcolor{blue}{\textbf{0.71$\pm$0.15}} & \textcolor{blue}{\textbf{33.13$\pm$7.59}} & 55.42$\pm$16.37 & 11.78$\pm$0.00 & \textcolor{red}{\textbf{89.98$\pm$4.68}} & 75.32$\pm$10.69 & \textcolor{red}{\textbf{93.89$\pm$2.76}} & \textcolor{red}{\textbf{25.20$\pm$4.03}} & \textcolor{red}{\textbf{33.81$\pm$4.68}} & \textcolor{red}{\textbf{88.60$\pm$2.82}} \\
& TranAD & 21.44$\pm$6.34 & 0.64$\pm$0.05 & 24.67$\pm$4.08 & \textcolor{blue}{\textbf{61.72$\pm$16.02}} & 13.04$\pm$0.00 & 5.72$\pm$0.29 & \textcolor{blue}{\textbf{78.49$\pm$11.04}} & \textcolor{blue}{\textbf{47.89$\pm$12.51}} & 2.24$\pm$0.35 & 12.04$\pm$2.99 & \textcolor{blue}{\textbf{25.63$\pm$2.40}} \\
& USAD & 16.11$\pm$4.62 & \textcolor{red}{\textbf{0.72$\pm$0.14}} & \textcolor{red}{\textbf{54.65$\pm$9.38}} & 59.23$\pm$17.84 & \textcolor{red}{\textbf{18.68$\pm$0.00}} & \textcolor{blue}{\textbf{4.30$\pm$0.46}} & 74.74$\pm$11.12 & 57.26$\pm$11.32 & \textcolor{blue}{\textbf{8.71$\pm$2.43}} & 10.00$\pm$2.71 & 14.11$\pm$1.56 \\
& OmniAnomaly & \textcolor{blue}{\textbf{24.64$\pm$7.11}} & 0.63$\pm$0.05 & 27.04$\pm$4.13 & \textcolor{red}{\textbf{74.63$\pm$9.01}} & \textcolor{blue}{\textbf{14.32$\pm$0.00}} & 6.23$\pm$0.54 & \textcolor{red}{\textbf{91.52$\pm$4.90}} & 46.19$\pm$12.01 & 2.37$\pm$0.43 & \textcolor{blue}{\textbf{16.19$\pm$3.70}} & \textcolor{red}{\textbf{29.30$\pm$4.10}} \\
\bottomrule
\end{tabular}
}
\end{table*}

\begin{table*}[h!]
\centering
\small
\setlength{\tabcolsep}{2pt}
\caption{Performance of VETime against vision-based baselines across 4 datasets (Mean $\pm$ 95\% CI). Best result is in \textcolor{red}{red}, second-best is in \textcolor{blue}{blue}.}
\label{tab:stat_d}
\resizebox{0.85\linewidth}{!}{%
\begin{tabular}{l l c c c c}
\toprule
\textbf{Metric} & \textbf{Method} & \textbf{NAB} & \textbf{YAHOO} & \textbf{SMAP} & \textbf{MSL} \\
\midrule
\multirow{5}{*}{Affiliation-F1}
& VETime & \textcolor{red}{\textbf{88.36$\pm$3.38}} & \textcolor{red}{\textbf{97.07$\pm$1.79}} & \textcolor{red}{\textbf{95.13$\pm$3.72}} & \textcolor{red}{\textbf{90.66$\pm$2.65}} \\
& VIT4TS & \textcolor{blue}{\textbf{64.64$\pm$11.91}} & \textcolor{blue}{\textbf{65.48$\pm$3.67}} & 93.55$\pm$4.20 & \textcolor{blue}{\textbf{69.90$\pm$22.83}} \\
& VLM4TS & 60.50$\pm$13.28 & 64.27$\pm$3.66 & \textcolor{blue}{\textbf{94.48$\pm$3.42}} & 64.63$\pm$25.15 \\
& VisualTimeAnomaly & 62.12$\pm$4.25 & 60.80$\pm$3.10 & 90.89$\pm$3.76 & 86.23$\pm$3.33 \\
& AnomLLM & 58.23$\pm$10.61 & 36.36$\pm$4.24 & 42.03$\pm$20.36 & 46.34$\pm$20.64 \\
\midrule
\multirow{5}{*}{$F1_T$}
& VETime & \textcolor{red}{\textbf{44.22$\pm$6.53}} & \textcolor{red}{\textbf{91.54$\pm$1.65}} & \textcolor{red}{\textbf{66.04$\pm$4.28}} & \textcolor{red}{\textbf{49.07$\pm$6.93}} \\
& VIT4TS & 30.07$\pm$3.03 & 8.05$\pm$1.77 & 51.13$\pm$2.58 & 23.48$\pm$8.77 \\
& VLM4TS & 28.24$\pm$2.62 & \textcolor{blue}{\textbf{9.44$\pm$1.73}} & \textcolor{blue}{\textbf{58.59$\pm$3.29}} & \textcolor{blue}{\textbf{38.81$\pm$2.54}} \\
& VisualTimeAnomaly & \textcolor{blue}{\textbf{40.34$\pm$5.73}} & 2.58$\pm$0.47 & 13.32$\pm$2.11 & 14.88$\pm$2.87 \\
& AnomLLM & 16.53$\pm$4.56 & 1.34$\pm$0.80 & 4.86$\pm$1.31 & 3.12$\pm$1.08 \\
\midrule
\multirow{5}{*}{Standard-F1}
& VETime & \textcolor{red}{\textbf{38.06$\pm$5.95}} & \textcolor{red}{\textbf{88.85$\pm$2.52}} & \textcolor{red}{\textbf{62.23$\pm$4.98}} & \textcolor{red}{\textbf{53.04$\pm$5.06}} \\
& VIT4TS & \textcolor{blue}{\textbf{35.75$\pm$7.73}} & 7.20$\pm$2.11 & 52.56$\pm$2.33 & 27.54$\pm$5.97 \\
& VLM4TS & 33.81$\pm$7.20 & \textcolor{blue}{\textbf{9.45$\pm$0.94}} & \textcolor{blue}{\textbf{58.90$\pm$3.33}} & \textcolor{blue}{\textbf{45.38$\pm$7.80}} \\
& VisualTimeAnomaly & 32.72$\pm$5.69 & 3.86$\pm$0.57 & 16.53$\pm$0.88 & 15.39$\pm$3.52 \\
& AnomLLM & 22.39$\pm$3.81 & 2.29$\pm$0.17 & 6.73$\pm$1.64 & 10.21$\pm$1.74 \\
\midrule
\multirow{5}{*}{VUS-PR}
& VETime & \textcolor{red}{\textbf{33.13$\pm$7.59}} & \textcolor{red}{\textbf{88.60$\pm$2.82}} & \textcolor{red}{\textbf{60.63$\pm$11.56}} & \textcolor{red}{\textbf{39.34$\pm$16.01}} \\
& VIT4TS & \textcolor{blue}{\textbf{27.20$\pm$5.43}} & \textcolor{blue}{\textbf{34.63$\pm$3.55}} & 51.83$\pm$9.51 & 27.88$\pm$14.78 \\
& VLM4TS & 25.71$\pm$5.33 & 33.30$\pm$3.44 & \textcolor{blue}{\textbf{52.92$\pm$8.87}} & \textcolor{blue}{\textbf{29.32$\pm$15.32}} \\
& VisualTimeAnomaly & \textcolor{blue}{\textbf{27.20$\pm$9.34}} & 13.15$\pm$1.78 & 12.70$\pm$7.33 & 20.73$\pm$15.02 \\
& AnomLLM & 19.58$\pm$5.20 & 10.49$\pm$1.45 & 5.26$\pm$2.74 & 10.12$\pm$3.94 \\
\midrule
\multirow{5}{*}{Per-series runtime (ms)}
& VETime & \textcolor{red}{\textbf{36.54$\pm$6.89}} & \textcolor{red}{\textbf{24.64$\pm$0.03}} & \textcolor{red}{\textbf{47.06$\pm$3.19}} & \textcolor{red}{\textbf{74.11$\pm$48.72}} \\
& VIT4TS & \textcolor{blue}{\textbf{1733.8$\pm$230.5}} & 2764.9$\pm$402.3 & \textcolor{blue}{\textbf{2440.1$\pm$885.0}} & 6553.2$\pm$442.3 \\
& VLM4TS & 6730.6$\pm$596.5 & 7416.8$\pm$389.8 & 7148.8$\pm$996.6 & 9958.2$\pm$402.1 \\
& VisualTimeAnomaly & 1980.6$\pm$129.3 & \textcolor{blue}{\textbf{2075.1$\pm$600.0}} & 3388.6$\pm$172.3 & \textcolor{blue}{\textbf{1832.0$\pm$232.6}} \\
& AnomLLM & 3593.9$\pm$256.4 & 3957.5$\pm$140.1 & 3501.7$\pm$539.3 & 3928.4$\pm$311.5 \\
\bottomrule
\end{tabular}
}
\end{table*}

\subsection{Comparison under different settings}\label{appendix:C2}

\begin{figure}
    \centering
    \includegraphics[width=0.8\linewidth]{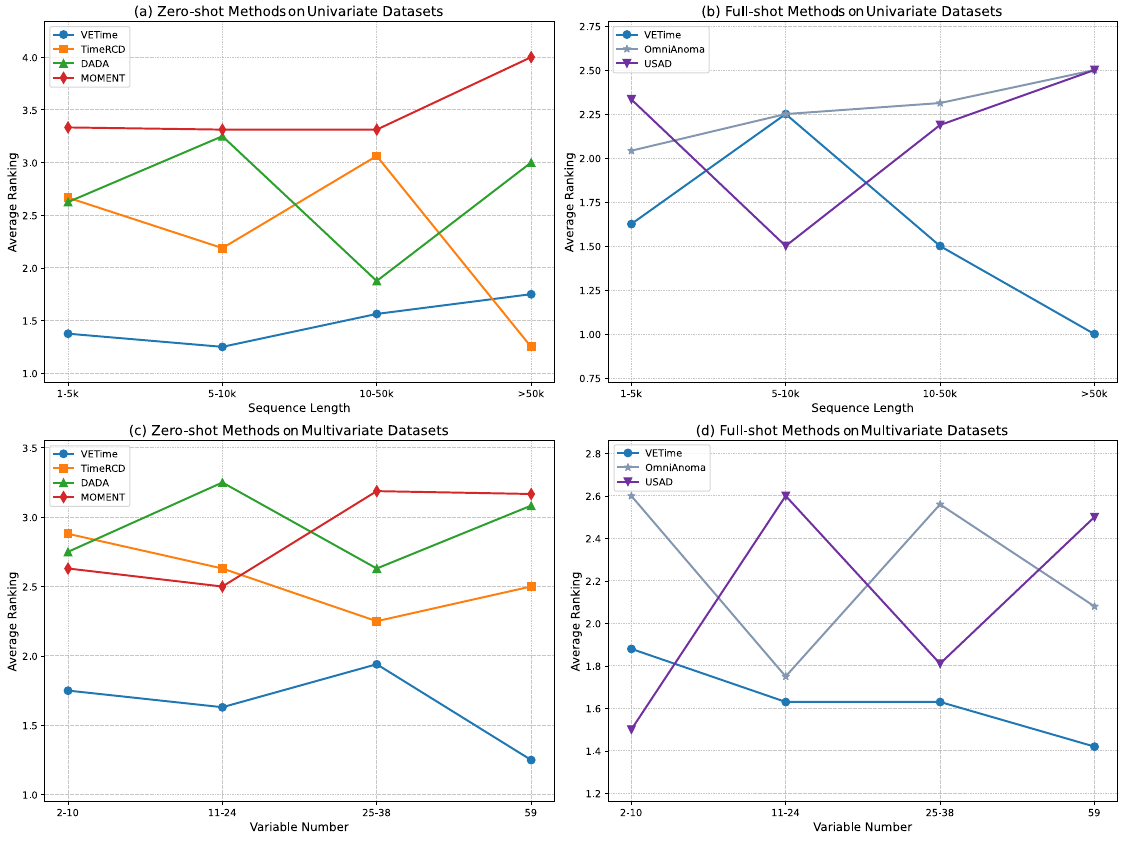}
    \caption{Performance of VETime (strictly zero-shot) against zero-shot and full-shot models. Metrics: Average Rank (lower is better) across 4 metrics on different datasets.}
    \label{fig:B}
\end{figure}

\paragraph{Comparison under different scalability.}
Figure~\ref{fig:B} evaluates the scalability of VETime with respect to varying sequence lengths and variable dimensions. For extremely long sequences, rather than naively compressing the entire series into a single visual representation, VETime adopts a sliding-window mechanism (with a window size of 5,000) inherited from its temporal backbone. This design choice ensures stable performance and prevents resolution degradation as sequence length increases.

Furthermore, when handling high-dimensional multivariate data, VETime maintains a highly competitive average rank as the number of variables grows. However, mapping multivariate data onto a fixed-size visual canvas ($224 \times 224$) over a constant temporal window (5000 steps) introduces a theoretical risk of information bottlenecks. A practical physical upper limit of approximately 10 variables ($224 \times 224 / 5000 \approx 10$) ensures that each data point occupies at least one distinct pixel without severe spatial overlap (aliasing). Exceeding this limit severely compresses the per-timestamp pixel representation, leading to visual information sparsity and a loss of fine-grained point-level details. As shown in Figure~\ref{fig:B}, empirical results indicate that performance experiences some degradation in the mid-to-high dimensional range (25-38 variables) due to this bottleneck.

Interestingly, VETime remains highly competitive on the 59-variable SMAP dataset, demonstrating that our framework remains robust and effective even in ultra-high-dimensional settings. This indicates that the actual failure mode is tied to \textbf{information redundancy} rather than just variable count. Datasets with highly correlated variables (like SMAP) suffer less from the visual compression bottleneck.

\subsection{Normal-anomaly feature separability}\label{appendix:C4}
\begin{figure}[t]
    \centering
    \includegraphics[width=0.8\linewidth]{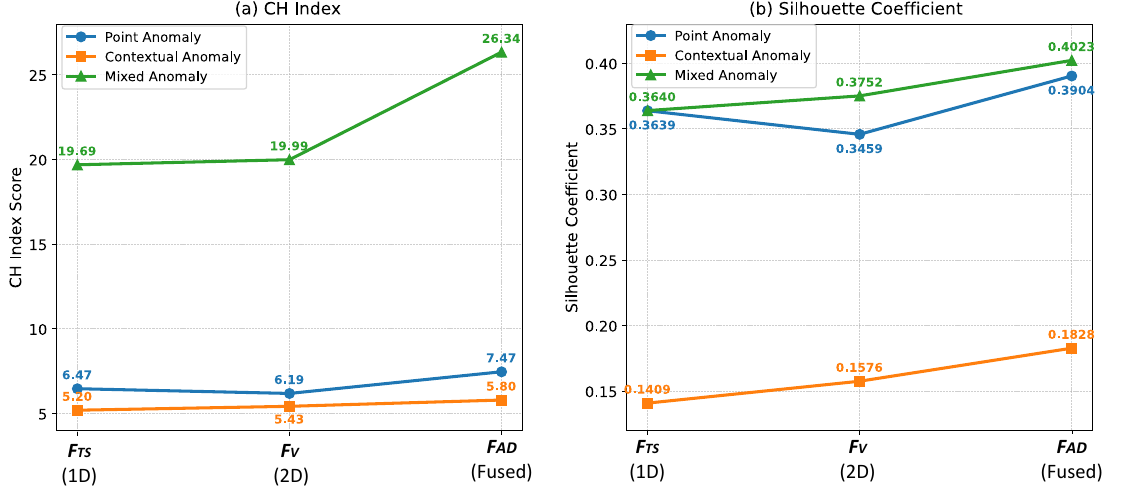}
    \caption{Quantitative validation in feature separability between normal and anomalous representations. The clustering metrics clearly illustrate that the 2D visual transformation introduces global awareness, while the final fusion $F_{AD}$ maximizes the overall normal/anomalous separability.}
    \label{fig:E}
\end{figure}

To further assess the discriminative power of the learned representations, we evaluated normal-anomaly feature separability of 1D temporal ($F_{TS}$), 2D visual ($F_V$), and fused ($F_{AD}$) outputs using the Calinski-Harabasz (CH) Index and Silhouette Coefficient. As shown in Figure~\ref{fig:E}, $F_{TS}$ excels at capturing local spikes, while $F_V$ trades local precision for stronger global awareness. Crucially, the fused representation $F_{AD}$ successfully synergizes both, achieving the highest separability across all scenarios.

The unique advantages of the 2D visual branch stem from three key mechanisms:

\begin{itemize}
    \item \textbf{Cross-Periodic Dependencies via folding:} Following the 2D temporal modeling principles established in TimesNet \citep{wu2023timesnet}, the folding mechanism converts these long-range temporal distances into vertical spatial adjacencies. This enables the 2D backbone to efficiently gain cross-periodic patterns without requiring inefficiently large 1D receptive fields.
    \item \textbf{Cross-Component Interaction via Channel Mapping:} By mapping the raw signal, Trend, and Remainder components into distinct RGB channels, ViT integrates the complex interplay between trend and residual information into unified token embeddings, naturally achieving early cross-component fusion.
    \item \textbf{Rich Contextual Patching:} For a length-5000 sequence, unlike 1D temporal patching yielding 313 localized, low-level numerical tokens (patch size 16), our 2D transformation constructs 196 structured tokens (patch size 16x16) derived from a $224\times224$ representation. Rather than diluting information, each visual patch intrinsically aggregates aforementioned cross-periodic dependencies and multi-component interactions, significantly amplifying the semantic density compared to 1D numerical tokens.
\end{itemize}

Moreover, pre-trained ViTs function as robust local edge detectors and global texture matchers \citep{chen2024visionts}. By transforming cross-periodic dynamics into spatial adjacencies, anomalies naturally manifest as structural edge breaks or localized texture distortions, aligning perfectly with the inductive biases of natural image perception.

\subsection{Efficiency \& performance.}\label{appendix:C5}
Table \ref{tab:F} compares the inference costs and parameters of VETime against a temporal-only baseline (TS Encoder + Classifier) and a size-matched variant (Dual-TS) that replaces the visual encoder with a temporal Transformer. While VETime incurs higher overhead than the baseline, it yields substantial gains. Under comparable costs, VETime significantly outperforms Dual-TS. This confirms our gains stem from the visual-temporal dual-branch design rather than mere parameter expansion, making the complexity a worthwhile trade-off.
\begin{table}[tbp]
\centering
\small
\setlength{\tabcolsep}{4pt}
\caption{Efficiency and performance comparison. Metrics are reported as Affiliation-F1 / VUS-PR (\%). Best result is in red, second-best is in blue.}
\label{tab:F}
\begin{tabular}{
    >{\raggedright\arraybackslash}p{1.1cm}
    >{\centering\arraybackslash}p{1.7cm}
    >{\centering\arraybackslash}p{1.7cm}
    >{\centering\arraybackslash}p{2.1cm}
    *{3}{>{\centering\arraybackslash}p{1.8cm}}
}
\toprule
\textbf{Model} & \textbf{Params. (M)} & \textbf{Mem. (MB)} & \textbf{Avg. Time (ms)} & \textbf{NAB} & \textbf{WSD} & \textbf{YAHOO} \\
\midrule
Base    & 37.06  & 223.73 & 24.49 & \textbf{\textcolor{blue}{82.48}} / 24.32 & \textbf{\textcolor{blue}{90.33}} / 21.77 & 96.65 / 84.41 \\
Dual-TS & 160.49 & 928.52 & 48.19 & 80.84 / \textbf{\textcolor{blue}{30.28}} & 88.87 / \textbf{\textcolor{blue}{27.24}} & \textbf{\textcolor{blue}{96.81}} / \textbf{\textcolor{blue}{87.25}} \\
VETime  & 157.96 & 787.41 & 51.02 & \textbf{\textcolor{red}{88.57}} / \textbf{\textcolor{red}{32.98}} & \textbf{\textcolor{red}{94.31}} / \textbf{\textcolor{red}{33.85}} & \textbf{\textcolor{red}{97.15}} / \textbf{\textcolor{red}{88.61}} \\
\bottomrule
\end{tabular}
\end{table}

\section{Ablation studies and in-depth analysis}\label{appendix:D}

\begin{table}[bp!]
\centering
\small
\setlength{\tabcolsep}{6pt}
\caption{Ablation study and comparison of vision encoders (Affiliation-F1 / VUS-PR (\%)). Best result is in red, second-best is in blue.}
\label{tab:vision_encoder_unified}
\begin{tabular}{
    >{\raggedright\arraybackslash}p{2.1cm}
    >{\centering\arraybackslash}p{1.7cm}
    *{3}{>{\centering\arraybackslash}p{2.8cm}}
}
\toprule
\textbf{Model} & \textbf{Params. (M)} & \textbf{NAB} & \textbf{WSD} & \textbf{YAHOO} \\
\midrule
Resnet          & 25.63  & 78.41 / 22.96 & 75.17 / 13.57 & 70.85 / 15.75 \\
ConvNeXt-Base   & 88.58  & 78.51 / 23.36 & 82.15 / 25.38 & 79.73 / 38.90 \\
CLIP-Base       & 85.64  & 88.42 / \textbf{\textcolor{blue}{32.72}} & 93.01 / 30.90 & 95.87 / 86.12 \\
MAE-Base        & 85.64  & \textbf{\textcolor{blue}{88.57}} / \textbf{\textcolor{red}{32.98}} & \textbf{\textcolor{red}{94.31}} / \textbf{\textcolor{blue}{33.85}} & \textbf{\textcolor{red}{97.15}} / \textbf{\textcolor{red}{88.61}} \\
MAE-Large       & 303.10 & \textbf{\textcolor{red}{88.67}} / 31.54 & \textbf{\textcolor{blue}{93.94}} / \textbf{\textcolor{red}{34.10}} & \textbf{\textcolor{blue}{97.03}} / \textbf{\textcolor{blue}{87.20}} \\
\bottomrule
\end{tabular}
\end{table}

\subsection{Ablation study on vision backbones}\label{appendix:D2}

We evaluate VETime across different vision architectures, pre-training paradigms, and scales, with the comprehensive results summarized in Table \ref{tab:vision_encoder_unified}.

\begin{itemize}
    \item \textbf{Architecture:} VETime implemented with a patch-tokenized ViT backbone significantly outperforms representative CNNs (ResNet, ConvNeXt-Base). This superiority arises because ViT's patch-based tokenization inherently aligns with our patch-level temporal alignment design, whereas CNN's spatial downsampling inherently disrupts the strict temporal alignment required for precise anomaly localization.
    \item \textbf{Pre-training Paradigm:} Among ViT architectures, MAE-based encoders consistently perform better than CLIP-based ones. This is likely because reconstruction-style pre-training better preserves the fine-grained structural cues that are critical for identifying anomalous patterns.
    \item \textbf{Scaling:} MAE-Base delivers the optimal balance of efficiency and performance. Scaling up the vision encoder to MAE-Large yields only marginal and inconsistent gains, which do not justify the $3.5\times$ increase in parameters.
\end{itemize}

\begin{figure}[t!]
    \centering
    \includegraphics[width=0.99\linewidth]{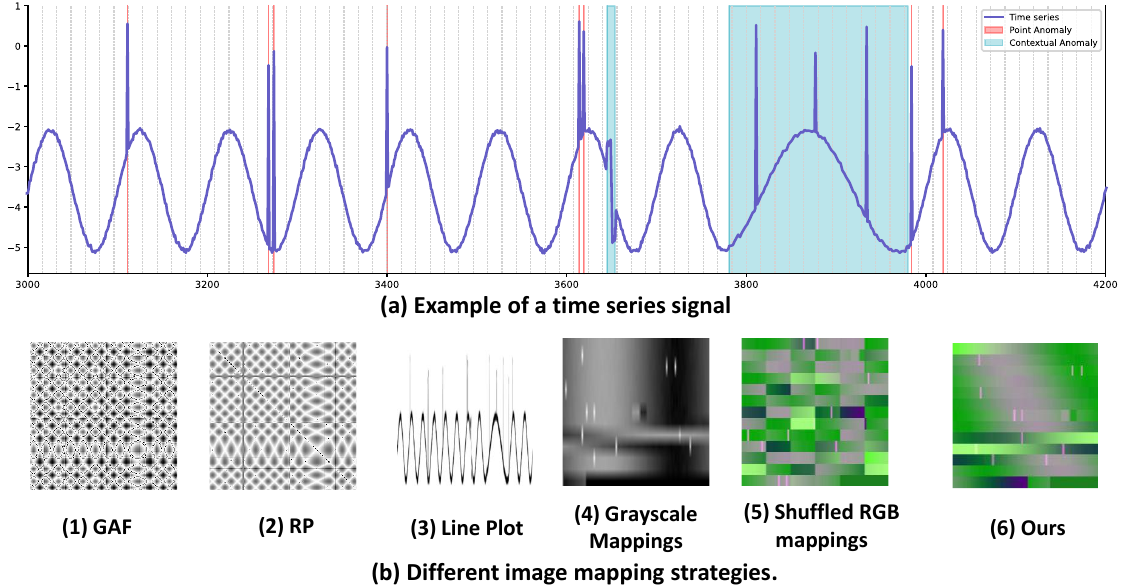}
    \caption{Qualitative comparison of different imaging strategies.}
    \label{fig:C2}
\end{figure}

\subsection{Comparison of the TS-to-image strategy}\label{appendix:C3}
To quantitatively validate the effectiveness of visual representations, the time series from 11 univariate datasets are categorized as: Point (short-duration spikes), Context (long-duration structural shifts), and Mixed.
Empirical visualizations (Figure~\ref{fig:C2}) and the comparison (Table \ref{tab:E2}) confirm our method outperforms others in identifying contextual anomalies. Unlike unaligned transformations (GAF/RP) that disrupt temporal semantics or sparse line plots (ViT4TS) causing massive spatial redundancy, VETime constructs a dense, patch-alignable 2D representation.

Motivated by the success of folded 2D spaces in capturing intra- and inter-period variations, we utilize dense intensity mappings combined with adaptive folding to preserve strict temporal alignment. Furthermore, mapping the raw series, trend, and remainder components into distinct RGB channels disentangles complex temporal signals. This design enables the Vision Transformer (ViT) to effectively aggregate periodicity and amplitude variations, thereby exposing long-range disruptions.

\begin{table}[htbp]
\centering
\small
\setlength{\tabcolsep}{6pt}
\caption{Performance comparison under different imaging strategies (Affiliation-F1 / VUS-PR (\%)). The time series from 11 univariate datasets are categorized as: \textit{Point (short-duration spikes), Contextual (long-duration structural shifts), and Mixed}. \textit{Base}: TS Encoder + Classifier. \textit{Shuffled RGB mappings}: Temporal order shuffled before mapping and restored before alignment. Best result is in \textbf{\textcolor{red}{bold red}}, second-best is in \textbf{\textcolor{blue}{bold blue}}.}
\label{tab:E2}
\begin{tabular}{
    >{\raggedright\arraybackslash}p{3.6cm}
    *{3}{>{\centering\arraybackslash}p{2.6cm}}
}
\toprule
\textbf{Imaging Strategies} & \textbf{Point} & \textbf{Contextual} & \textbf{Mixed} \\
\midrule
Base                      & 85.65 / 67.31 & 77.18 / 23.89 & 83.90 / 39.35 \\
(1) GAF                   & 86.12 / 72.88 & 77.01 / 23.57 & 84.63 / 42.05 \\
(2) RP                    & 85.79 / 73.91 & 77.07 / 23.87 & 84.88 / 42.65 \\
(3) Line plot             & 86.14 / 74.82 & 78.20 / 24.92 & 84.77 / 45.31 \\
(4) Gray mappings         & 87.97 / 77.74 & \textbf{\textcolor{blue}{79.50}} / \textbf{\textcolor{blue}{26.65}} & \textbf{\textcolor{blue}{86.07}} / \textbf{\textcolor{blue}{46.48}} \\
(5) Shuffled RGB mappings & \textbf{\textcolor{blue}{88.75}} / \textbf{\textcolor{blue}{78.42}} & 78.94 / 25.51 & 84.43 / 46.36 \\
(6) Ours                  & \textbf{\textcolor{red}{89.15}} / \textbf{\textcolor{red}{79.78}} & \textbf{\textcolor{red}{80.54}} / \textbf{\textcolor{red}{28.35}} & \textbf{\textcolor{red}{86.71}} / \textbf{\textcolor{red}{47.99}} \\
\bottomrule
\end{tabular}
\end{table}

\paragraph{Effectiveness of Traceable Image Conversion (TIC).}
While basic operations such as seasonal-trend decomposition or RGB mapping are individually established, the technical novelty of Traceable Image Conversion (TIC) lies in their synergistic integration into an informative, artifact-free 1D-to-2D interface tailored for TSAD. Specifically, TIC operates through four core mechanisms:
\begin{enumerate}
    \item \textbf{Semantic Channel Mapping:} Allocating the raw series, trend, and residual components to separate RGB channels jointly exposes low-frequency structural deviations and high-frequency residual spikes to the visual backbone.
    \item \textbf{High-Density Context Aggregation:} Condensing a 5000-step time series into 196 visual tokens aggregates long-range context while exploiting pre-trained vision models, accelerating inference $\sim 100\times$ compared to plot-rendering baselines (e.g., 0.03s/iter vs. ViT4TS at 3.14s/iter).
    \item \textbf{Dimension-Aware Scaling:} Decoupling scaling along the two axes respects their distinct physical semantics: temporal interpolation ensures continuity along the time axis, while copy-padding along the period axis prevents visual boundary artifacts.
    \item \textbf{Traceable Patch Alignment:} Constraining the folding width exactly to the ViT patch size (16) guarantees that each visual token encapsulates strictly continuous 1D timestamps, enabling precise 2D-to-1D feature alignment.
\end{enumerate}

\subsection{Comparison of multimodal fusion strategies}\label{appendix:late_fusion}
To justify feature-level temporal alignment, we evaluated late-fusion baselines as well as direct feature fusion strategies, with detailed performance comparison reported in Table \ref{tab:late_fusion}. Our Patch-level Alignment outperforms them by resolving two critical limitations:
\begin{enumerate}
    \item \textbf{Limitations of Late Fusion:} Late fusion restricts the model to local optima and cannot leverage early cross-modal interaction. Consequently, visual global context cannot guide temporal features, and precise temporal features cannot rectify visual localization errors, inevitably leading to error accumulation.
    \item \textbf{Limitations of Direct Feature Fusion:} Visual patches inherently lack explicit temporal indices. Naively concatenating multimodal features introduces severe spatio-temporal noise. Our alignment anchors visual tokens to corresponding temporal tokens, which is crucial for point-wise TSAD localization.
\end{enumerate}
Furthermore, our superiority stems from structurally indispensable cross-modal interactions that resolve late-fusion's architectural bottleneck.

\begin{table*}[t!]
\centering
\small
\setlength{\tabcolsep}{2.5pt}
\renewcommand{\arraystretch}{0.95}
\caption{Performance comparison of different multimodal fusion strategies (Affiliation-F1 / VUS-PR (\%)). Best result is in \textcolor{red}{red}, second-best is in \textcolor{blue}{blue}.}
\label{tab:late_fusion}
\resizebox{\linewidth}{!}{%
\begin{tabular}{l *{11}{c}}
\toprule
\textbf{Strategy} & \textbf{IOPS} & \textbf{MGAB} & \textbf{NAB} & \textbf{NEK} & \textbf{Power} & \textbf{SED} & \textbf{Stock} & \textbf{TODS} & \textbf{UCR} & \textbf{WSD} & \textbf{YAHOO} \\
\midrule
Direct feature-level fusion & \textcolor{blue}{\textbf{87.40}}/\textcolor{blue}{\textbf{22.67}} & \textcolor{blue}{\textbf{67.51}}/\textcolor{blue}{\textbf{0.73}} & \textcolor{red}{\textbf{83.31}}/26.99 & 79.07/50.03 & 83.62/21.34 & 95.58/\textcolor{blue}{\textbf{79.05}} & 68.18/\textcolor{blue}{\textbf{74.38}} & \textcolor{blue}{\textbf{86.30}}/\textcolor{blue}{\textbf{93.14}} & 84.06/\textcolor{blue}{\textbf{22.48}} & \textcolor{blue}{\textbf{89.53}}/\textcolor{blue}{\textbf{22.76}} & \textcolor{blue}{\textbf{94.97}}/80.33 \\
Score-level fusion & 86.78/20.35 & 67.09/0.71 & 81.91/\textcolor{blue}{\textbf{27.04}} & \textcolor{red}{\textbf{81.09}}/\textcolor{red}{\textbf{60.98}} & \textcolor{blue}{\textbf{83.96}}/\textcolor{blue}{\textbf{21.36}} & 93.87/77.53 & \textcolor{blue}{\textbf{68.33}}/74.33 & 86.04/92.04 & \textcolor{blue}{\textbf{84.33}}/22.32 & 89.51/22.43 & 94.22/\textcolor{blue}{\textbf{82.34}} \\
Decision-level fusion & 86.24/16.07 & \textcolor{red}{\textbf{67.93}}/0.70 & 80.89/25.50 & 79.50/48.66 & 83.04/21.16 & \textcolor{blue}{\textbf{95.79}}/77.32 & 66.93/71.92 & 86.12/92.57 & 82.27/18.38 & 89.44/18.20 & 93.47/79.43 \\
Alignment feature-level fusion & \textcolor{red}{\textbf{88.40}}/\textcolor{red}{\textbf{25.75}} & 67.36/\textcolor{red}{\textbf{0.75}} & \textcolor{blue}{\textbf{82.93}}/\textcolor{red}{\textbf{27.75}} & \textcolor{blue}{\textbf{79.54}}/\textcolor{blue}{\textbf{50.63}} & \textcolor{red}{\textbf{85.63}}/\textcolor{red}{\textbf{21.41}} & \textcolor{red}{\textbf{96.36}}/\textcolor{red}{\textbf{84.84}} & \textcolor{red}{\textbf{68.65}}/\textcolor{red}{\textbf{75.84}} & \textcolor{red}{\textbf{86.53}}/\textcolor{red}{\textbf{93.55}} & \textcolor{red}{\textbf{84.88}}/\textcolor{red}{\textbf{23.37}} & \textcolor{red}{\textbf{90.56}}/\textcolor{red}{\textbf{24.70}} & \textcolor{red}{\textbf{95.19}}/\textcolor{red}{\textbf{83.82}} \\
\bottomrule
\end{tabular}
}
\end{table*}

\subsection{Ablation study on training objectives}\label{appendix:D4}
To analyze how different multi-task optimization objectives interact during joint training, we systematically evaluate combinations of Binary Cross-Entropy ($\mathcal{L}_{BCE}$), Mean Squared Error ($\mathcal{L}_{MSE}$), Anomaly Window Contrastive Loss ($\mathcal{L}_{aw}$), and Routing Entropy Regularization ($\mathcal{L}_{e}$). The ablation study in Table \ref{tab:B} illustrates these objective interactions. $\mathcal{L}_{aw}$ (anomaly contrastive) generally works cooperatively with $\mathcal{L}_{BCE}$ (anomaly supervision) increasing separation between normal and anomalous features. Conversely, the temporal fidelity regularized by $\mathcal{L}_{MSE}$ conflicts with the sharp anomaly boundaries required by $\mathcal{L}_{BCE}$. Crucially, $\mathcal{L}_{e}$ helps reconcile these objectives via Task-Adaptive Routing to achieve peak performance.

\begin{table}[bp]
\centering
\small
\setlength{\tabcolsep}{2pt}
\renewcommand{\arraystretch}{0.9}
\caption{Ablation study on training objectives (Affiliation-F1 / VUS-PR (\%)). Base: $\mathcal{L}_{BCE}$. Best result is in red, second-best is in blue.}
\label{tab:B}
\begin{tabular}{
    >{\raggedright\arraybackslash}p{2.0cm}
    *{3}{>{\centering\arraybackslash}p{1.0cm}}
    *{3}{>{\centering\arraybackslash}p{2.6cm}}
}
\toprule
\textbf{Variant} & $\mathcal{L}_{MSE}$ & $\mathcal{L}_{aw}$ & $\mathcal{L}_{e}$ & \textbf{NAB} & \textbf{WSD} & \textbf{YAHOO} \\
\midrule
Base & & & & 82.93 / 27.75 & 90.56 / 24.70 & 95.19 / 83.82 \\
Only $\mathcal{L}_{MSE}$  & $\checkmark$ & & & 82.05 / 27.64 & 90.10 / 24.86 & 95.20 / 83.29 \\
Only $\mathcal{L}_{aw}$  & & $\checkmark$ & & 83.99 / 29.52 & 90.11 / 26.30 & 96.86 / 85.31 \\
w/o $\mathcal{L}_{e}$  & $\checkmark$ & $\checkmark$ & & 84.62 / 28.81 & 90.72 / 27.47 & 96.64 / 85.56 \\
w/o $\mathcal{L}_{MSE}$  & & $\checkmark$ & $\checkmark$ & \textbf{\textcolor{blue}{86.96}} / 28.44 & 90.87 / 28.63 & 96.88 / 86.58 \\
w/o $\mathcal{L}_{aw}$  & $\checkmark$ & & $\checkmark$ & 86.84 / \textbf{\textcolor{blue}{31.23}} & \textbf{\textcolor{blue}{91.63}} / \textbf{\textcolor{blue}{29.00}} & \textbf{\textcolor{blue}{97.11}} / \textbf{\textcolor{blue}{87.26}} \\
Full Model & $\checkmark$ & $\checkmark$ & $\checkmark$ & \textbf{\textcolor{red}{88.57}} / \textbf{\textcolor{red}{32.98}} & \textbf{\textcolor{red}{94.31}} / \textbf{\textcolor{red}{33.85}} & \textbf{\textcolor{red}{97.15}} / \textbf{\textcolor{red}{88.61}} \\
\bottomrule
\end{tabular}
\end{table}

\begin{table*}[t!]
\centering
\small
\setlength{\tabcolsep}{2pt}
\renewcommand{\arraystretch}{0.9}
\caption{Ablation on all 11 datasets (Affiliation-F1 / VUS-PR (\%)). Best results are in red, and second-best are in blue. Base: TS Encoder + Classifier, TIC: Traceable Image Conversion, PTA: Patch-level Temporal Alignment, AWCL: Anomaly Window Contrastive Learning, TMF: Task-adaptive Multi-modal Fusion.}
\label{tab:ablation_all_11}
\resizebox{\linewidth}{!}{%
\begin{tabular}{l c c c c *{11}{c}}
\toprule
\textbf{Variant} & \textbf{TIC} & \textbf{PTA} & \textbf{AWCL} & \textbf{TMF} & \textbf{IOPS} & \textbf{MGAB} & \textbf{NAB} & \textbf{NEK} & \textbf{Power} & \textbf{SED} & \textbf{Stock} & \textbf{TODS} & \textbf{UCR} & \textbf{WSD} & \textbf{YAHOO} \\
\midrule
Base & & & & & 83.28/20.23 & \textcolor{red}{\textbf{70.69}}/\textcolor{red}{\textbf{1.05}} & 82.48/24.32 & \textcolor{blue}{\textbf{79.73}}/27.88 & 85.51/21.25 & 96.87/80.75 & \textcolor{red}{\textbf{71.84}}/\textcolor{red}{\textbf{77.28}} & 86.37/93.46 & 84.63/23.09 & 90.33/21.77 & 96.65/84.41 \\
V1 & \checkmark & & & & 87.40/22.67 & 67.51/0.73 & 83.31/26.99 & 79.07/50.03 & 83.62/21.34 & 95.58/79.05 & 68.18/74.38 & 86.30/93.14 & 84.06/22.48 & 89.53/22.76 & 94.97/80.33 \\
V2 & \checkmark & \checkmark & & & 88.40/25.75 & 67.36/0.75 & 82.93/27.75 & 79.54/50.63 & \textcolor{blue}{\textbf{85.63}}/\textcolor{blue}{\textbf{21.41}} & 96.36/84.84 & 68.65/75.84 & \textcolor{blue}{\textbf{86.53}}/93.55 & \textcolor{blue}{\textbf{84.88}}/23.37 & 90.56/24.70 & 95.19/83.82 \\
V3 & \checkmark & \checkmark & \checkmark & & 90.53/26.15 & 68.23/0.77 & 83.99/29.52 & \textcolor{red}{\textbf{79.74}}/53.10 & \textcolor{red}{\textbf{85.78}}/\textcolor{red}{\textbf{21.48}} & \textcolor{blue}{\textbf{97.16}}/86.94 & \textcolor{blue}{\textbf{70.89}}/\textcolor{blue}{\textbf{76.58}} & \textcolor{red}{\textbf{86.73}}/93.76 & \textcolor{red}{\textbf{85.06}}/23.99 & 90.11/26.30 & 96.86/85.31 \\
V4 & \checkmark & \checkmark & & \checkmark & \textcolor{blue}{\textbf{90.84}}/\textcolor{blue}{\textbf{28.15}} & \textcolor{blue}{\textbf{68.73}}/\textcolor{blue}{\textbf{0.78}} & \textcolor{blue}{\textbf{86.84}}/\textcolor{blue}{\textbf{31.23}} & 79.32/\textcolor{blue}{\textbf{55.38}} & 79.08/11.54 & 96.79/\textcolor{blue}{\textbf{87.39}} & 69.07/75.68 & 85.60/\textcolor{blue}{\textbf{93.90}} & 84.46/\textcolor{blue}{\textbf{24.83}} & \textcolor{blue}{\textbf{91.63}}/\textcolor{blue}{\textbf{29.00}} & \textcolor{blue}{\textbf{97.11}}/\textcolor{blue}{\textbf{87.26}} \\
V5 & \checkmark & & & \checkmark & 90.56/26.26 & 67.22/0.75 & 83.23/24.81 & 78.85/48.04 & 78.63/11.33 & 95.78/85.37 & 69.60/75.60 & 86.05/93.20 & 84.75/24.40 & 90.54/24.91 & 96.06/82.62 \\
V6 & & \checkmark & & \checkmark & 90.29/24.56 & 67.20/0.73 & 82.89/25.64 & 77.50/46.10 & 78.62/11.18 & 94.68/82.29 & 68.85/74.36 & 84.96/93.24 & 84.26/22.59 & 88.63/22.32 & 94.84/81.95 \\
V7 & & & \checkmark & \checkmark & 90.53/24.86 & 66.24/0.70 & 81.84/24.86 & 75.07/45.05 & 77.32/10.91 & 94.13/77.84 & 67.51/74.22 & 85.48/93.54 & 83.16/22.33 & 88.34/20.20 & 94.12/77.68 \\
\textbf{Full} & \checkmark & \checkmark & \checkmark & \checkmark & \textcolor{red}{\textbf{90.53}}/\textcolor{red}{\textbf{30.79}} & 68.03/0.70 & \textcolor{red}{\textbf{88.57}}/\textcolor{red}{\textbf{32.98}} & 79.56/\textcolor{red}{\textbf{57.54}} & 78.78/11.78 & \textcolor{red}{\textbf{97.31}}/\textcolor{red}{\textbf{89.05}} & 69.60/76.08 & 85.85/\textcolor{red}{\textbf{94.20}} & \textcolor{red}{\textbf{85.06}}/\textcolor{red}{\textbf{24.99}} & \textcolor{red}{\textbf{94.31}}/\textcolor{red}{\textbf{33.85}} & \textcolor{red}{\textbf{97.15}}/\textcolor{red}{\textbf{88.61}} \\
\bottomrule
\end{tabular}
}
\end{table*}

\subsection{Detailed Ablation Study}\label{appendix:D5}

To assess cross-domain robustness without dataset-specific tuning and evaluate the individual necessity of each core module, we extend the incremental ablation study (variants V1-V7) across all 11 benchmark datasets. Each variant is pretrained on the synthetic corpus and evaluated without
target-domain retraining or fine-tuning. The quantitative results are summarized in Table \ref{tab:ablation_all_11}. As observed, the removal of any key module leads to a discernible performance drop, confirming that all components are integral to achieving optimal detection capabilities.

\textbf{Traceable Image Conversion (TIC):} Removing TIC (e.g., variants V6 and V7) causes the most severe performance collapse, with VUS-PR plummeting from 33.85\% to 20.20\% (V7) on WSD and from 57.54\% to 45.05\% (V7) on NEK. This result supports the conclusion that the proposed image conversion lays a solid foundation for subsequent alignment and fusion. Furthermore, without this visual-temporal alignment foundation, multimodal interaction fails, which in most cases degrades performance below the unimodal baseline (e.g., on datasets like YAHOO and UCR).

\textbf{Patch-level Temporal Alignment (PTA):} Omitting PTA (e.g., variant V5) restricts fusion to coarse semantic interactions. The resulting misalignment and loss of fine-grained local correspondences lead to a noticeable decline in detection capability, evidenced by the VUS-PR dropping to 26.26\% on the IOPS dataset compared to the full model's 30.79\%.

\textbf{Anomaly Window Contrastive Learning (AWCL):} Although constructing effective contrastive pairs is challenging on densely anomalous datasets, AWCL helps the model distinguish normal from anomalous patterns more sharply. Specifically, AWCL is essential for capturing both fine-grained local variations (e.g., pushing NAB VUS-PR from 31.23\% in V4 to 32.98\% in the full model) and global contextual disparities in structural anomalies (e.g., elevating SED from 87.39\% to 89.05\%). This multi-scale discriminability confirms the highly complementary nature of our intra- and inter-window contrastive losses.

\textbf{Task-adaptive Multi-modal Fusion (TMF):} Eliminating TMF (e.g., variant V3, which relies on simple concatenation) causes consistent performance regression across benchmarks, such as a drop in VUS-PR from 94.20\% to 93.76\% on TODS and from 88.61\% to 85.31\% on YAHOO. This validates that the task-aware gating mechanism dynamically regulates feature integration, ensuring robust performance by balancing the conflicting demands of reconstruction and classification.

Building on this component-wise validation, we also observe performance degradations on specific datasets (MGAB, Stock, Power). Analyzing the variants reveals failure modes tied to intrinsic dataset properties:

\begin{itemize}
    \item \textbf{Non-Periodic Signals (MGAB, Stock):} These datasets feature deterministic chaos (MGAB) or stochastic random walks (Stock) without meaningful periodicity. Forcing them into a 2D grid generates visual noise. Consequently, performance drops immediately upon introducing Traceable Image Conversion (\textbf{V1}), and downstream fusion cannot recover it.
    \item \textbf{Periodic Spike Anomalies (Power):} In the Power dataset, normal sequences lack spikes, whereas anomalies manifest as periodic spike noise. Because our visual conversion (TIC) explicitly constructs 2D representations based on periodicity, it inadvertently encodes these periodic anomalies as normal background structures. While the temporal modality correctly identifies them (making \textbf{V3} the best performer), the zero-shot fusion router (\textbf{V4}) incorrectly assigns high weight to this deceived visual representation, triggering massive negative transfer.
\end{itemize}

\begin{table}[t]
\centering
\small
\caption{Ablation study grouped by model components: PTA (Patch-level Temporal Alignment), AWCL (Anomaly Window Contrastive Learning), and TMF (Task-adaptive Multi-modal Fusion). Metrics are reported as Affiliation-F1 / VUS-PR (\%). Best result is in red, second-best is in blue.}
\label{tab:ablation_all}
\begin{tabular}{
    >{\centering\arraybackslash}p{1.0cm}
    >{\raggedright\arraybackslash}p{3.4cm}
    *{3}{>{\centering\arraybackslash}p{2.4cm}}
}
\toprule
\textbf{Model} & \textbf{Component} & \textbf{NAB} & \textbf{WSD} & \textbf{YAHOO} \\
\midrule
\multirow{3}{*}{PTA}
& w/o adaptive folding       & \textbf{\textcolor{blue}{87.81}} / 29.99 & 91.49 / 30.41 & 97.10 / 86.70 \\
& w/o positional encoding    & \textbf{\textcolor{red}{88.16}} / 30.27 & 92.29 / \textbf{\textcolor{red}{31.18}} & \textbf{\textcolor{blue}{97.25}} / 86.81 \\
& w/o attention              & 85.69 / 29.88 & 91.57 / 29.46 & 96.92 / 85.53 \\
\midrule
\multirow{2}{*}{AWCL}
& w/o inter-window contrast  & 86.69 / \textbf{\textcolor{red}{31.17}} & \textbf{\textcolor{blue}{92.85}} / 30.54 & 97.15 / \textbf{\textcolor{blue}{87.29}} \\
& w/o intra-window contrast  & 86.15 / \textbf{\textcolor{blue}{30.86}} & \textbf{\textcolor{red}{92.93}} / \textbf{\textcolor{blue}{30.57}} & \textbf{\textcolor{red}{97.77}} / \textbf{\textcolor{red}{88.29}} \\
\midrule
\multirow{3}{*}{TMF}
& Concatenation              & 85.07 / 28.54 & 89.80 / 26.39 & 95.69 / 85.72 \\
& Addition                   & 83.99 / 29.52 & 90.11 / 26.30 & 96.86 / 85.31 \\
& w/o reconstruction head    & 86.96 / 28.44 & 90.87 / 28.63 & 96.88 / 86.58 \\
\bottomrule
\end{tabular}
\end{table}

\subsection{Ablation study on each component}\label{appendix:D3}
Beyond high-level modular ablations, we conduct a micro-level sub-component ablation to isolate the granular impact of internal mechanisms within PTA (adaptive folding, positional encoding, attention), AWCL (intra- vs. inter-window contrast), and TMF (fusion operations and auxiliary reconstruction head) on the NAB, WSD, and YAHOO datasets. Table \ref{tab:ablation_all} summarizes these component-level empirical findings.

\paragraph{Effectiveness of patch-level temporal alignment (PTA).}
We investigated the contributions of adaptive folding, positional encoding, and the attention mechanism within the PTA module. As observed, removing the \textbf{attention mechanism} resulted in the most significant performance degradation across all datasets (e.g., Affiliation-F1 dropped to 85.69\% on NAB). This indicates that the attention mechanism is pivotal for capturing long-range dependencies and accurately aligning visual patches with temporal contexts. Furthermore, the exclusion of \textbf{adaptive folding} also led to a notable decline, validating its role in preserving intrinsic temporal structures during feature transformation. These results confirm that the complete PTA design is essential for ensuring fine-grained semantic correspondence.

\paragraph{Effectiveness of anomaly window contrastive learning (AWCL).}
The ablation results for AWCL demonstrate the necessity of combined contrastive strategies, though their individual impacts vary by dataset. The removal of \textbf{intra-window contrast} causes a severe drop in detection accuracy on datasets with fine-grained local anomalies like NAB. Conversely, the absence of \textbf{inter-window contrast} is more detrimental on datasets like WSD, highlighting its importance in distinguishing broader contextual anomalies globally. Interestingly, on densely anomalous datasets like YAHOO, removing intra-window contrast can even marginally improve performance. Overall, the synergy of both strategies is crucial for maximizing the discriminability of diverse anomaly patterns across different scenarios.

\paragraph{Effectiveness of task-adaptive multi-modal fusion (TMF).}
In the TMF module, we compared our proposed adaptive fusion against naive strategies (Concatenation and Addition) and examined the role of the reconstruction head. The results reveal that simple \textbf{Concatenation} and \textbf{Addition} yielded the poorest performance, proving that simple linear combinations fail to capture complex cross-modal interactions. Additionally, removing the \textbf{reconstruction head} led to a consistent performance drop. This empirical evidence supports the conclusion that sequence reconstruction serves as a critical auxiliary constraint, promoting deep feature interaction and enhancing the robustness of fused representations for the primary anomaly classification task.

\begin{figure}[b]
    \centering
    \includegraphics[width=0.8\linewidth]{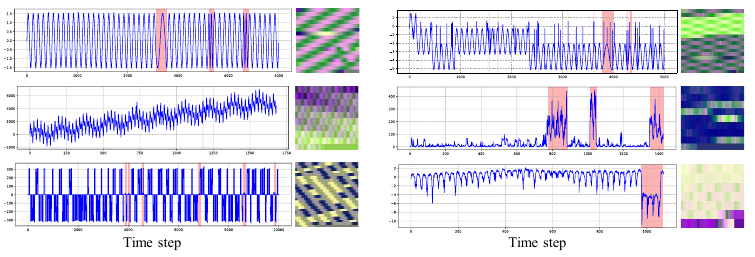}
    \caption{Initial time series and mapping images.}
    \label{fig:img9}
\end{figure}

\begin{figure}[t]
    \centering
    \includegraphics[width=0.8\linewidth]{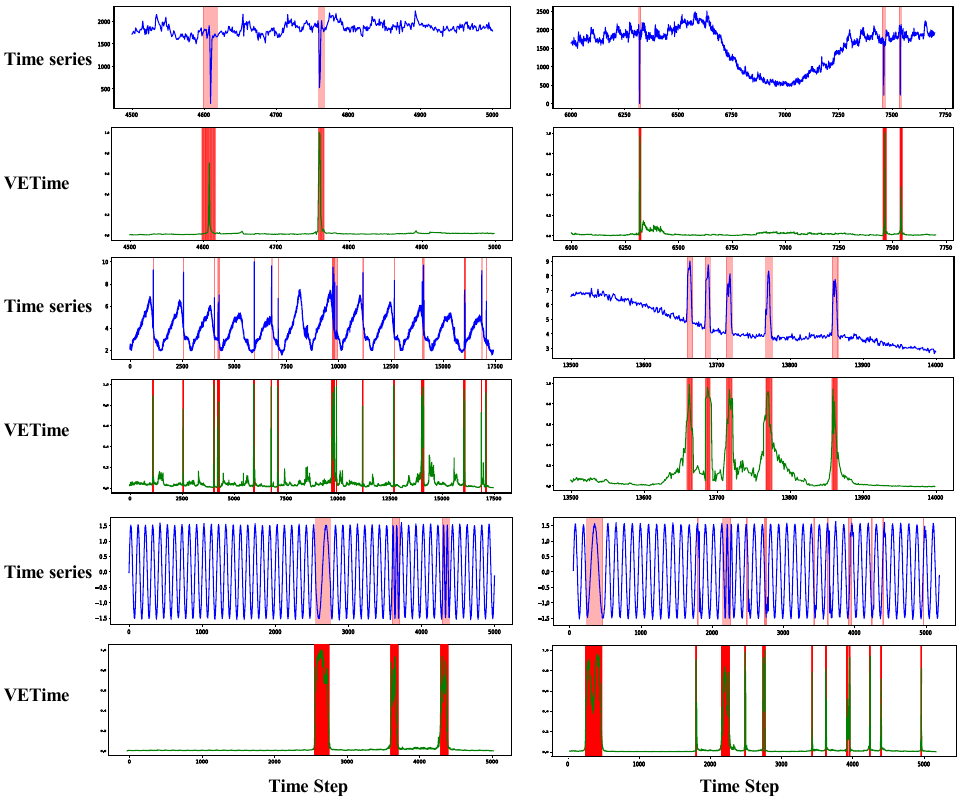}
    \caption{Qualitative visualization of anomaly detection results. Blue curve: original time series; red markers: ground-truth anomaly labels; green curve: predicted anomaly scores.}
    \label{fig:img10}
\end{figure}

\section{Visualized results}\label{appendix:E}
\subsection{Visualized mappings}\label{appendix:E1}
Figure~\ref{fig:img9} visualizes the representations generated by our Traceable Image Conversion module across diverse data samples. The left panels display the raw 1D time-series sequences with ground-truth anomalies highlighted in red, while the right panels present their corresponding 2D visual transformations. As observed, the generated images effectively encode key temporal characteristics, such as periodicity, trends, and noise, into distinct visual textures and patterns. This visualization confirms that our conversion method successfully preserves information-dense global contexts, providing a discriminative basis for the subsequent multi-modal fusion.

\subsection{Visualized anomaly scores}\label{appendix:E2}
Figure~\ref{fig:img10} presents qualitative anomaly detection results across heterogeneous time series patterns, validating the method's robustness to diverse anomaly patterns. For point anomalies—such as abrupt spikes or isolated dips (visible in Rows 1–2), the green anomaly score curve exhibits sharp, well-localized peaks that precisely align with annotated labels. For context anomalies characterized by extended deviations like sustained trend shifts or contextual drifts (Row 3), the anomaly scores maintain elevated levels throughout the entire anomalous interval, maintaining continuous and coherent alignment with the red-labeled events.


\newpage
\section*{NeurIPS Paper Checklist}

\begin{enumerate}

\item {\bf Claims}
    \item[] Question: Do the main claims made in the abstract and introduction accurately reflect the paper's contributions and scope?
    \item[] Answer: \answerYes{} 
    \item[] Justification: The claims made in the abstract and introduction, including the proposed VETime framework and its superior zero-shot performance, are well supported by the theoretical explanations in Section 3 and the extensive experimental results in Section 4.
    \item[] Guidelines:
    \begin{itemize}
        \item The answer \answerNA{} means that the abstract and introduction do not include the claims made in the paper.
        \item The abstract and/or introduction should clearly state the claims made, including the contributions made in the paper and important assumptions and limitations. A \answerNo{} or \answerNA{} answer to this question will not be perceived well by the reviewers. 
        \item The claims made should match theoretical and experimental results, and reflect how much the results can be expected to generalize to other settings. 
        \item It is fine to include aspirational goals as motivation as long as it is clear that these goals are not attained by the paper. 
    \end{itemize}

\item {\bf Limitations}
    \item[] Question: Does the paper discuss the limitations of the work performed by the authors?
    \item[] Answer: \answerYes{} 
    \item[] Justification: We explicitly discuss the limitations of our work in the Limitations.
    \item[] Guidelines:
    \begin{itemize}
        \item The answer \answerNA{} means that the paper has no limitation while the answer \answerNo{} means that the paper has limitations, but those are not discussed in the paper. 
        \item The authors are encouraged to create a separate ``Limitations'' section in their paper.
        \item The paper should point out any strong assumptions and how robust the results are to violations of these assumptions (e.g., independence assumptions, noiseless settings, model well-specification, asymptotic approximations only holding locally). The authors should reflect on how these assumptions might be violated in practice and what the implications would be.
        \item The authors should reflect on the scope of the claims made, e.g., if the approach was only tested on a few datasets or with a few runs. In general, empirical results often depend on implicit assumptions, which should be articulated.
        \item The authors should reflect on the factors that influence the performance of the approach. For example, a facial recognition algorithm may perform poorly when image resolution is low or images are taken in low lighting. Or a speech-to-text system might not be used reliably to provide closed captions for online lectures because it fails to handle technical jargon.
        \item The authors should discuss the computational efficiency of the proposed algorithms and how they scale with dataset size.
        \item If applicable, the authors should discuss possible limitations of their approach to address problems of privacy and fairness.
        \item While the authors might fear that complete honesty about limitations might be used by reviewers as grounds for rejection, a worse outcome might be that reviewers discover limitations that aren't acknowledged in the paper. The authors should use their best judgment and recognize that individual actions in favor of transparency play an important role in developing norms that preserve the integrity of the community. Reviewers will be specifically instructed to not penalize honesty concerning limitations.
    \end{itemize}

\item {\bf Theory assumptions and proofs}
    \item[] Question: For each theoretical result, does the paper provide the full set of assumptions and a complete (and correct) proof?
    \item[] Answer: \answerNA{} 
    \item[] Justification: This paper focuses on proposing a novel deep learning architecture and empirical evaluations, and does not include formal theoretical proofs or theorems.
    \item[] Guidelines:
    \begin{itemize}
        \item The answer \answerNA{} means that the paper does not include theoretical results. 
        \item All the theorems, formulas, and proofs in the paper should be numbered and cross-referenced.
        \item All assumptions should be clearly stated or referenced in the statement of any theorems.
        \item The proofs can either appear in the main paper or the supplemental material, but if they appear in the supplemental material, the authors are encouraged to provide a short proof sketch to provide intuition. 
        \item Inversely, any informal proof provided in the core of the paper should be complemented by formal proofs provided in appendix or supplemental material.
        \item Theorems and Lemmas that the proof relies upon should be properly referenced. 
    \end{itemize}

    \item {\bf Experimental result reproducibility}
    \item[] Question: Does the paper fully disclose all the information needed to reproduce the main experimental results of the paper to the extent that it affects the main claims and/or conclusions of the paper (regardless of whether the code and data are provided or not)?
    \item[] Answer: \answerYes{} 
    \item[] Justification: We provide comprehensive details regarding datasets, baselines, evaluation metrics, and implementation hyperparameters (e.g., learning rate, batch size, LoRA settings) in Appendix \ref{appendix:B} and Appendix \ref{appendix:C} to ensure reproducibility.
    \item[] Guidelines:
    \begin{itemize}
        \item The answer \answerNA{} means that the paper does not include experiments.
        \item If the paper includes experiments, a \answerNo{} answer to this question will not be perceived well by the reviewers: Making the paper reproducible is important, regardless of whether the code and data are provided or not.
        \item If the contribution is a dataset and\slash or model, the authors should describe the steps taken to make their results reproducible or verifiable. 
        \item Depending on the contribution, reproducibility can be accomplished in various ways. For example, if the contribution is a novel architecture, describing the architecture fully might suffice, or if the contribution is a specific model and empirical evaluation, it may be necessary to either make it possible for others to replicate the model with the same dataset, or provide access to the model. In general. releasing code and data is often one good way to accomplish this, but reproducibility can also be provided via detailed instructions for how to replicate the results, access to a hosted model (e.g., in the case of a large language model), releasing of a model checkpoint, or other means that are appropriate to the research performed.
        \item While NeurIPS does not require releasing code, the conference does require all submissions to provide some reasonable avenue for reproducibility, which may depend on the nature of the contribution. For example
        \begin{enumerate}
            \item If the contribution is primarily a new algorithm, the paper should make it clear how to reproduce that algorithm.
            \item If the contribution is primarily a new model architecture, the paper should describe the architecture clearly and fully.
            \item If the contribution is a new model (e.g., a large language model), then there should either be a way to access this model for reproducing the results or a way to reproduce the model (e.g., with an open-source dataset or instructions for how to construct the dataset).
            \item We recognize that reproducibility may be tricky in some cases, in which case authors are welcome to describe the particular way they provide for reproducibility. In the case of closed-source models, it may be that access to the model is limited in some way (e.g., to registered users), but it should be possible for other researchers to have some path to reproducing or verifying the results.
        \end{enumerate}
    \end{itemize}

\item {\bf Open access to data and code}
    \item[] Question: Does the paper provide open access to the data and code, with sufficient instructions to faithfully reproduce the main experimental results, as described in supplemental material?
    \item[] Answer: \answerYes{} 
    \item[] Justification: The TSB-AD datasets are publicly available, and the VETime implementation, trained model, and reproduction instructions will be made publicly available with the final release.
    \item[] Guidelines:
    \begin{itemize}
        \item The answer \answerNA{} means that paper does not include experiments requiring code.
        \item Please see the NeurIPS code and data submission guidelines (\url{https://neurips.cc/public/guides/CodeSubmissionPolicy}) for more details.
        \item While we encourage the release of code and data, we understand that this might not be possible, so \answerNo{} is an acceptable answer. Papers cannot be rejected simply for not including code, unless this is central to the contribution (e.g., for a new open-source benchmark).
        \item The instructions should contain the exact command and environment needed to run to reproduce the results. See the NeurIPS code and data submission guidelines (\url{https://neurips.cc/public/guides/CodeSubmissionPolicy}) for more details.
        \item The authors should provide instructions on data access and preparation, including how to access the raw data, preprocessed data, intermediate data, and generated data, etc.
        \item The authors should provide scripts to reproduce all experimental results for the new proposed method and baselines. If only a subset of experiments are reproducible, they should state which ones are omitted from the script and why.
        \item At submission time, to preserve anonymity, the authors should release anonymized versions (if applicable).
        \item Providing as much information as possible in supplemental material (appended to the paper) is recommended, but including URLs to data and code is permitted.
    \end{itemize}

\item {\bf Experimental setting/details}
    \item[] Question: Does the paper specify all the training and test details (e.g., data splits, hyperparameters, how they were chosen, type of optimizer) necessary to understand the results?
    \item[] Answer: \answerYes{} 
    \item[] Justification: Detailed experimental settings, including the optimizer, learning rate, batch size, loss coefficients, and network architecture specifics, are well documented in Appendix \ref{appendix:B}.
    \item[] Guidelines:
    \begin{itemize}
        \item The answer \answerNA{} means that the paper does not include experiments.
        \item The experimental setting should be presented in the core of the paper to a level of detail that is necessary to appreciate the results and make sense of them.
        \item The full details can be provided either with the code, in appendix, or as supplemental material.
    \end{itemize}

\item {\bf Experiment statistical significance}
    \item[] Question: Does the paper report error bars suitably and correctly defined or other appropriate information about the statistical significance of the experiments?
    \item[] Answer: \answerYes{} 
    \item[] Justification: Appendix \ref{appendix:statistical_support} reports mean $\pm$ standard deviation over five independent runs and sequence-level bootstrap 95\% confidence intervals, together with Friedman tests and Holm-corrected Wilcoxon signed-rank tests.
    \item[] Guidelines:
    \begin{itemize}
        \item The answer \answerNA{} means that the paper does not include experiments.
        \item The authors should answer \answerYes{} if the results are accompanied by error bars, confidence intervals, or statistical significance tests, at least for the experiments that support the main claims of the paper.
        \item The factors of variability that the error bars are capturing should be clearly stated (for example, train/test split, initialization, random drawing of some parameter, or overall run with given experimental conditions).
        \item The method for calculating the error bars should be explained (closed form formula, call to a library function, bootstrap, etc.)
        \item The assumptions made should be given (e.g., Normally distributed errors).
        \item It should be clear whether the error bar is the standard deviation or the standard error of the mean.
        \item It is OK to report 1-sigma error bars, but one should state it. The authors should preferably report a 2-sigma error bar than state that they have a 96\% CI, if the hypothesis of Normality of errors is not verified.
        \item For asymmetric distributions, the authors should be careful not to show in tables or figures symmetric error bars that would yield results that are out of range (e.g., negative error rates).
        \item If error bars are reported in tables or plots, the authors should explain in the text how they were calculated and reference the corresponding figures or tables in the text.
    \end{itemize}

\item {\bf Experiments compute resources}
    \item[] Question: For each experiment, does the paper provide sufficient information on the computer resources (type of compute workers, memory, time of execution) needed to reproduce the experiments?
    \item[] Answer: \answerYes{} 
    \item[] Justification: Appendix \ref{appendix:B} reports the core training hardware (one NVIDIA H20 24 GB GPU), model memory consumption, and inference runtime comparisons needed to reproduce and assess the computational requirements of the experiments.
    \item[] Guidelines:
    \begin{itemize}
        \item The answer \answerNA{} means that the paper does not include experiments.
        \item The paper should indicate the type of compute workers CPU or GPU, internal cluster, or cloud provider, including relevant memory and storage.
        \item The paper should provide the amount of compute required for each of the individual experimental runs as well as estimate the total compute. 
        \item The paper should disclose whether the full research project required more compute than the experiments reported in the paper (e.g., preliminary or failed experiments that didn't make it into the paper). 
    \end{itemize}
    
\item {\bf Code of ethics}
    \item[] Question: Does the research conducted in the paper conform, in every respect, with the NeurIPS Code of Ethics \url{https://neurips.cc/public/EthicsGuidelines}?
    \item[] Answer: \answerYes{} 
    \item[] Justification: Our research focuses on general time-series anomaly detection algorithms and conforms to the NeurIPS Code of Ethics without violating any ethical guidelines.
    \item[] Guidelines:
    \begin{itemize}
        \item The answer \answerNA{} means that the authors have not reviewed the NeurIPS Code of Ethics.
        \item If the authors answer \answerNo, they should explain the special circumstances that require a deviation from the Code of Ethics.
        \item The authors should make sure to preserve anonymity (e.g., if there is a special consideration due to laws or regulations in their jurisdiction).
    \end{itemize}

\item {\bf Broader impacts}
    \item[] Question: Does the paper discuss both potential positive societal impacts and negative societal impacts of the work performed?
    \item[] Answer: \answerYes{} 
    \item[] Justification: VETime may improve fault detection in industrial operations and medical monitoring. However, false positives or missed anomalies caused by distribution shift or training-data bias could trigger unnecessary intervention or delayed responses, so high-stakes deployments should retain human oversight and domain-specific validation.
    \item[] Guidelines:
    \begin{itemize}
        \item The answer \answerNA{} means that there is no societal impact of the work performed.
        \item If the authors answer \answerNA{} or \answerNo, they should explain why their work has no societal impact or why the paper does not address societal impact.
        \item Examples of negative societal impacts include potential malicious or unintended uses (e.g., disinformation, generating fake profiles, surveillance), fairness considerations (e.g., deployment of technologies that could make decisions that unfairly impact specific groups), privacy considerations, and security considerations.
        \item The conference expects that many papers will be foundational research and not tied to particular applications, let alone deployments. However, if there is a direct path to any negative applications, the authors should point it out. For example, it is legitimate to point out that an improvement in the quality of generative models could be used to generate Deepfakes for disinformation. On the other hand, it is not needed to point out that a generic algorithm for optimizing neural networks could enable people to train models that generate Deepfakes faster.
        \item The authors should consider possible harms that could arise when the technology is being used as intended and functioning correctly, harms that could arise when the technology is being used as intended but gives incorrect results, and harms following from (intentional or unintentional) misuse of the technology.
        \item If there are negative societal impacts, the authors could also discuss possible mitigation strategies (e.g., gated release of models, providing defenses in addition to attacks, mechanisms for monitoring misuse, mechanisms to monitor how a system learns from feedback over time, improving the efficiency and accessibility of ML).
    \end{itemize}
    
\item {\bf Safeguards}
    \item[] Question: Does the paper describe safeguards that have been put in place for responsible release of data or models that have a high risk for misuse (e.g., pre-trained language models, image generators, or scraped datasets)?
    \item[] Answer: \answerNA{} 
    \item[] Justification: The proposed time-series anomaly detection model and the benchmarks used do not pose a high risk for misuse (unlike generative LLMs or deepfakes), so no special safeguards are necessary.
    \item[] Guidelines:
    \begin{itemize}
        \item The answer \answerNA{} means that the paper poses no such risks.
        \item Released models that have a high risk for misuse or dual-use should be released with necessary safeguards to allow for controlled use of the model, for example by requiring that users adhere to usage guidelines or restrictions to access the model or implementing safety filters. 
        \item Datasets that have been scraped from the Internet could pose safety risks. The authors should describe how they avoided releasing unsafe images.
        \item We recognize that providing effective safeguards is challenging, and many papers do not require this, but we encourage authors to take this into account and make a best faith effort.
    \end{itemize}

\item {\bf Licenses for existing assets}
    \item[] Question: Are the creators or original owners of assets (e.g., code, data, models), used in the paper, properly credited and are the license and terms of use explicitly mentioned and properly respected?
    \item[] Answer: \answerYes{} 
    \item[] Justification: We cite the creators of TSB-AD and the pre-trained models used in our experiments, respect their applicable licenses and terms of use, and will document the corresponding license information in the public code repository.
    \item[] Guidelines:
    \begin{itemize}
        \item The answer \answerNA{} means that the paper does not use existing assets.
        \item The authors should cite the original paper that produced the code package or dataset.
        \item The authors should state which version of the asset is used and, if possible, include a URL.
        \item The name of the license (e.g., CC-BY 4.0) should be included for each asset.
        \item For scraped data from a particular source (e.g., website), the copyright and terms of service of that source should be provided.
        \item If assets are released, the license, copyright information, and terms of use in the package should be provided. For popular datasets, \url{paperswithcode.com/datasets} has curated licenses for some datasets. Their licensing guide can help determine the license of a dataset.
        \item For existing datasets that are re-packaged, both the original license and the license of the derived asset (if it has changed) should be provided.
        \item If this information is not available online, the authors are encouraged to reach out to the asset's creators.
    \end{itemize}

\item {\bf New assets}
    \item[] Question: Are new assets introduced in the paper well documented and is the documentation provided alongside the assets?
    \item[] Answer: \answerNA{} 
    \item[] Justification: The paper proposes a new method but does not release a new dataset, trained-model checkpoint, or standalone asset package; experiments rely on existing public benchmarks.
    \item[] Guidelines:
    \begin{itemize}
        \item The answer \answerNA{} means that the paper does not release new assets.
        \item Researchers should communicate the details of the dataset\slash code\slash model as part of their submissions via structured templates. This includes details about training, license, limitations, etc. 
        \item The paper should discuss whether and how consent was obtained from people whose asset is used.
        \item At submission time, remember to anonymize your assets (if applicable). You can either create an anonymized URL or include an anonymized zip file.
    \end{itemize}

\item {\bf Crowdsourcing and research with human subjects}
    \item[] Question: For crowdsourcing experiments and research with human subjects, does the paper include the full text of instructions given to participants and screenshots, if applicable, as well as details about compensation (if any)? 
    \item[] Answer: \answerNA{} 
    \item[] Justification: This algorithmic research does not involve any crowdsourcing or research with human subjects.
    \item[] Guidelines:
    \begin{itemize}
        \item The answer \answerNA{} means that the paper does not involve crowdsourcing nor research with human subjects.
        \item Including this information in the supplemental material is fine, but if the main contribution of the paper involves human subjects, then as much detail as possible should be included in the main paper. 
        \item According to the NeurIPS Code of Ethics, workers involved in data collection, curation, or other labor should be paid at least the minimum wage in the country of the data collector. 
    \end{itemize}

\item {\bf Institutional review board (IRB) approvals or equivalent for research with human subjects}
    \item[] Question: Does the paper describe potential risks incurred by study participants, whether such risks were disclosed to the subjects, and whether Institutional Review Board (IRB) approvals (or an equivalent approval/review based on the requirements of your country or institution) were obtained?
    \item[] Answer: \answerNA{} 
    \item[] Justification: Our research does not involve human subjects, making IRB approval inapplicable.
    \item[] Guidelines:
    \begin{itemize}
        \item The answer \answerNA{} means that the paper does not involve crowdsourcing nor research with human subjects.
        \item Depending on the country in which research is conducted, IRB approval (or equivalent) may be required for any human subjects research. If you obtained IRB approval, you should clearly state this in the paper. 
        \item We recognize that the procedures for this may vary significantly between institutions and locations, and we expect authors to adhere to the NeurIPS Code of Ethics and the guidelines for their institution. 
        \item For initial submissions, do not include any information that would break anonymity (if applicable), such as the institution conducting the review.
    \end{itemize}

\item {\bf Declaration of LLM usage}
    \item[] Question: Does the paper describe the usage of LLMs if it is an important, original, or non-standard component of the core methods in this research? Note that if the LLM is used only for writing, editing, or formatting purposes and does \emph{not} impact the core methodology, scientific rigor, or originality of the research, declaration is not required.
    \item[] Answer: \answerNA{} 
    \item[] Justification: Large Language Models (LLMs) are not utilized as a component in our core methodology.
    \item[] Guidelines:
    \begin{itemize}
        \item The answer \answerNA{} means that the core method development in this research does not involve LLMs as any important, original, or non-standard components.
        \item Please refer to our LLM policy in the NeurIPS handbook for what should or should not be described.
    \end{itemize}

\end{enumerate}

\end{document}